\documentclass{article}

\PassOptionsToPackage{table}{xcolor}
\usepackage[table]{xcolor}
\usepackage{colortbl}

\usepackage[eandd, final]{neurips_2026}

\definecolor{rankone}{RGB}{140,180,255} 
\definecolor{ranktwo}{RGB}{190,215,255} 
\definecolor{rankthree}{RGB}{230,240,255} 

\usepackage[utf8]{inputenc} 
\usepackage[T1]{fontenc}    
\usepackage{hyperref}       
\usepackage{url}            
\usepackage{booktabs}       
\usepackage{amsfonts}       
\usepackage{amsmath}
\usepackage{nicefrac}       
\usepackage{microtype}      
\usepackage{xcolor}         
\usepackage{enumitem}
\usepackage{graphicx}
\usepackage{makecell}
\usepackage{arydshln}
\usepackage{adjustbox}
\usepackage{tcolorbox}
\tcbuselibrary{breakable, skins}
\usepackage{longtable}
\usepackage{multirow}
\usepackage{caption}

\newcommand{\projectname}{Spatial-IQ}

\title{\projectname: Deconstructing Spatial Intelligence via Hierarchical Capability Tests}

\author{%
  Patrick Rim$^{1,2}$\quad Tom Long$^{1}$\quad Ekta Prashnani$^{1}$\quad Ruth Rosenholtz$^{1}$\quad Ben Boudaoud$^{1}$ \\
  \textbf{Peter Xenopoulos$^{1}$\quad Alex Wong$^{2}$\quad Joohwan Kim$^{1}$\quad Jae-Hyun Jung$^{1}$} \\
  $^{1}$NVIDIA Research\quad $^{2}$Yale University
}

\begin{document}

\maketitle

\begin{abstract}
Multimodal large language models (MLLMs) excel at visual interpretation but fail on spatial reasoning tasks that humans solve reliably. Existing benchmarks evaluate these models as black boxes, limiting their ability to identify the underlying causes of lower performance: when a model fails a spatial reasoning task, it remains difficult to ascertain whether the hurdle is perceptual, such as recognizing object boundaries, or cognitive, such as reasoning about occlusion to infer hidden geometry. We introduce \projectname, a hierarchical diagnostic framework that decomposes object counting in stacked 3D structures into 9 perceptual and cognitive sub-tasks organized by the developmental stages of human spatial cognition, with mental rotation as an additional target probe. Using NVIDIA Isaac Sim, we procedurally generated a diverse dataset of roughly 80{,}000 scenes of stacked 3D structures with per-task ground truth. We evaluate models across three output formats (free-response text, multiple-choice images, and image editing) alongside a human baseline. The \projectname{} framework shows that top-performing models often succeed at the target task (object counting) without succeeding on the lower-level sub-tasks intended to support it, and that models differ in how much of these hierarchical chains they preserve. Raw target-task accuracy alone would obscure how often they reach the target in ways that diverge from the human decomposition. Finally, we find that chain-of-thought (CoT) supervision over our sub-tasks, combined with reinforcement learning with verifiable rewards, jointly improves target-task accuracy and sub-task competence. We therefore use the decomposition as a diagnostic tool and as a training signal rather than as a proven causal prerequisite.
\end{abstract}

\section{Introduction}
Spatial intelligence is a fundamental prerequisite for physical AI and embodied agents to reliably navigate, manipulate, and interact with the real physical world~\citep{goyal2025vla}. Consider a robot stacking boxes in a warehouse: to place a new box safely, it must individuate the boxes on the pallet, infer non-visible boxes that provide support, and judge whether the resulting structure will hold under gravity. 

Current foundation models appear to possess spatial intelligence. They can identify which of two objects is on top of the other, generate a 3D-rendered view of a room, or provide a plausible description of given images. And yet, the same models routinely produce videos in which objects shrink, expand, or pass through one another~\citep{li2025worldmodelbench, meng2024towards}, fall well short of human performance on spatial reasoning over real scenes~\citep{yang2025thinking}, and falter on counting a stack of identical boxes under occlusion~\citep{zhan20263viewsense, Gong2025SpaCE10AC, pothiraj2025capture}, a task children solve with relative ease~\citep{kaufman1983kaufman, drozdick2018kaufman}. 

We ask to what extent the spatial intelligence of humans and AI models conforms to systematic and hierarchical reasoning. The aforementioned model failures suggest that they may act as surface-level simulators, relying on statistical correlations between text and pixels rather than a grounded causal model of the 3D world~\citep{chen2026babyvision}. Humans themselves often lack ground-truth metric knowledge of scenes~\cite{koenderink2011vision}, and some human understanding of physics may remain at the intuitive level~\cite{battaglia2013simulation, fischer2016functional, smith2013sources}.

\begin{figure}[t]
\centering
\includegraphics[width=0.92\textwidth]{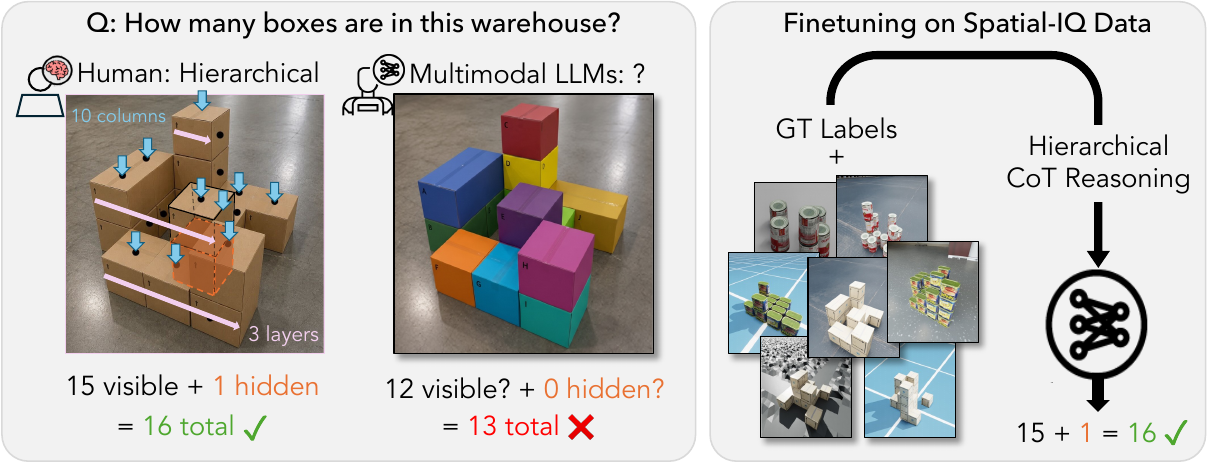}
\vspace{-1mm}
\caption{We hypothesize that humans count stacked 3D objects by decomposing the structure into columns, layers, and visible versus hidden blocks (left). Current MLLMs miscount, and fine-tuning on \projectname{} with chain-of-thought supervision over our proposed hierarchy improves their counts (right).}
\vspace{-4mm}
\label{fig:teaser}
\end{figure}

Current benchmarks typically evaluate models on complex reasoning tasks that yield a binary success metric~\citep{yang2025thinking} or performance on multiple-choice questions~\citep{xu2025spatialbench, xiao2026spatialtree}, limiting the capacity for diagnostic evaluation. For example, a failure could stem from a low-level perceptual breakdown, such as misidentifying object boundaries, or from a high-level cognitive breakdown, such as failing to infer hidden geometry for gravitational support.

\textbf{A hierarchical diagnostic benchmark.} To close this gap, we propose a benchmark that treats spatial intelligence as a hierarchy of dependencies, inspired by classical human developmental psychology~\citep{piaget1956child} and spatial cognition~\citep{piaget1969psychology}. We present \projectname{}, featuring a framework deconstructing a target task (e.g., 3D object counting) into a hierarchical set of sub-tasks, a dataset of visual stimuli with ground truths, and tools for analyzing the results. This framework allows us to not only judge model performance on the target task, but also to diagnose \textit{why} the model failed on the target task.

We design this framework around a target task drawn directly from standardized cognitive assessment: 3D object counting. The task is to count the number of identical objects in a stacked 3D structure, including those occluded from view. This task appears as the Block Counting subtest of the Kaufman Assessment Battery for Children~\citep{kaufman1983kaufman, drozdick2018kaufman}, which measures visual processing and fluid reasoning. The task was designed to diagnose 3D structural understanding: The constituent objects are visually identical across scenes, while the structures they form differ substantially. To perform the task correctly, a responder typically reconstructs a 3D arrangement of identical objects from partial observations, infers occlusion and support constraints, and enumerates exactly~\citep{dumery2025counting}. The novelty of our work as an AI benchmark lies in a per-scene diagnostic within one target task: we test whether a responder that reaches the correct count also answers correctly the sub-tasks the count presupposes, rather than reporting only whether the final count is correct.

\textbf{Evaluation and human studies.} We evaluate a wide range of state-of-the-art models spanning closed-source MLLMs, open-weight MLLMs, and image-editing models, and we additionally conduct human studies to anchor our analysis. We find a consistent pattern: Even the top-performing models often succeed on the target task while failing on the sub-tasks that the target task ostensibly requires; humans, by contrast, exhibit the strong hierarchical structure that the developmental psychology literature predicts. This gap points to a path forward: training models with the hierarchical structure to recover the intermediate competencies rather than only the target task answer.

\textbf{A two-phase training paradigm.} To this end, we extend the \projectname{} framework from evaluation to training. We first apply supervised fine-tuning to teach models an explicit chain-of-thought format derived from the sub-task dependencies. We then apply reinforcement learning with verifiable rewards~\citep{yu2025dapo}, rewarding the final integer answer and the required answer format. Our experiments show that the decomposition serves as a training signal that jointly improves target and sub-task competence relative to standard end-task training, rather than as a proven causal prerequisite.

\section{Related Work}

\paragraph{Human Spatial Intelligence.}
The human developmental psychology literature has developed and validated a battery of spatial intelligence tasks over more than half a century, beginning with Piaget and Inhelder's work on the development of spatial reasoning in children~\citep{piaget1956child, piaget1969psychology}. They describe spatial cognition as proceeding through topological, projective, and Euclidean stages. We adopt two diagnostic tasks from this literature as candidate targets: counting in stacked structures~\citep{kaufman1983kaufman, drozdick2018kaufman}, which integrates object individuation, structural decomposition, and physical reasoning about hidden support~\citep{baillargeon1987object}, and mental rotation~\citep{shepard1971mental,tarr1989mental}, which probes the ability to imagine objects from unseen viewpoints. A stacked-structure count combines visible objects with occluded objects that \emph{must} exist to support them under gravity, and end-task accuracy alone cannot show whether a responder's answers are consistent with these components \citep{baillargeon1987object,spelke1992origins}. Classic instruments typically rely on schematic or abstract diagrams. In contrast, our benchmark features realistic 3D scenes where occlusion, support, and viewpoint must be inferred directly from the image, providing a more ecologically aligned complement to existing tools~\citep{chen2026babyvision}.

\paragraph{Evaluation of Spatial Intelligence in Large Models.}
A growing body of work has documented spatial reasoning failures in MLLMs. VSI-Bench~\citep{yang2025thinking}, BabyVision~\citep{chen2026babyvision}, SAT~\citep{ray2024sat}, MMSI-Bench~\citep{yang2025mmsi}, CAPTURe \citep{pothiraj2025capture}, and EgoExoBench~\citep{he2025egoexobench} respectively evaluate egocentric video reasoning, infant-developmental milestones, dynamic camera and object motion, multi-image settings, occluded object counting, and viewpoint association across paired first- and third-person video. Each documents systematic gaps between MLLM and human performance, but each evaluates models as black boxes, reporting end-task scores without diagnosing whether the underlying failure is perceptual, structural, or compositional. We additionally evaluate VLA-0~\citep{goyal2025vla}, a vision-language-action model built on the Qwen2.5-VL backbone, allowing direct comparison between a VLM and the VLA derived from it.

Most directly related to our framework, SpaCE-10~\citep{Gong2025SpaCE10AC} formalizes spatial intelligence as ten atomic capabilities combined into eight compositional tasks, and it reports atomic-level analyses alongside compositional results. The two levels are reported separately, so SpaCE-10 does not show whether a compositional answer succeeds on an item while its linked atomic capabilities fail on that same item. SpatialBench~\citep{xu2025spatialbench} and SpatialTree~\citep{xiao2026spatialtree} propose taxonomic hierarchies of spatial competence (five and four levels, respectively, spanning observation through high-level planning), and SpatialLadder~\citep{li2026spatialladder} organizes a progressive training curriculum across seven independent dimensions. These levels and dimensions index separate tasks, whereas our hierarchy traces dependencies inside one task, which lets us condition on prerequisite correctness. CAPTURe~\citep{pothiraj2025capture} and 3ViewSense~\citep{zhan20263viewsense} are also close in domain. CAPTURe counts objects behind an occluder by pattern continuation rather than gravity-governed support and scores end-task accuracy, with no target-linked component decomposition. 3ViewSense is a training method over orthographic views that does not elicit and score components per scene. Capability hierarchies and verifiable-reward RL both appear in prior work, so neither is our contribution. We contribute a per-scene test, within a single target task, of whether a correct answer coexists with failed prerequisites.

A parallel line of work asks whether video generation models simulate physical reality or produce plausible-looking pixels without grounded physical understanding~\citep{li2025worldmodelbench, meng2024towards, duan2025worldscore, qin2024worldsimbench}. Like the MLLM benchmarks above, these evaluate end-task generation quality rather than diagnosing where the underlying reasoning breaks down. \projectname{} is extensible to video and world-model evaluation (Appendix~\ref{app:future_work}).

\section{\projectname{} Benchmark}
\paragraph{Target Tasks and Sub-Task Suite}

\begin{figure}[t]
\centering
\includegraphics[width=\textwidth]{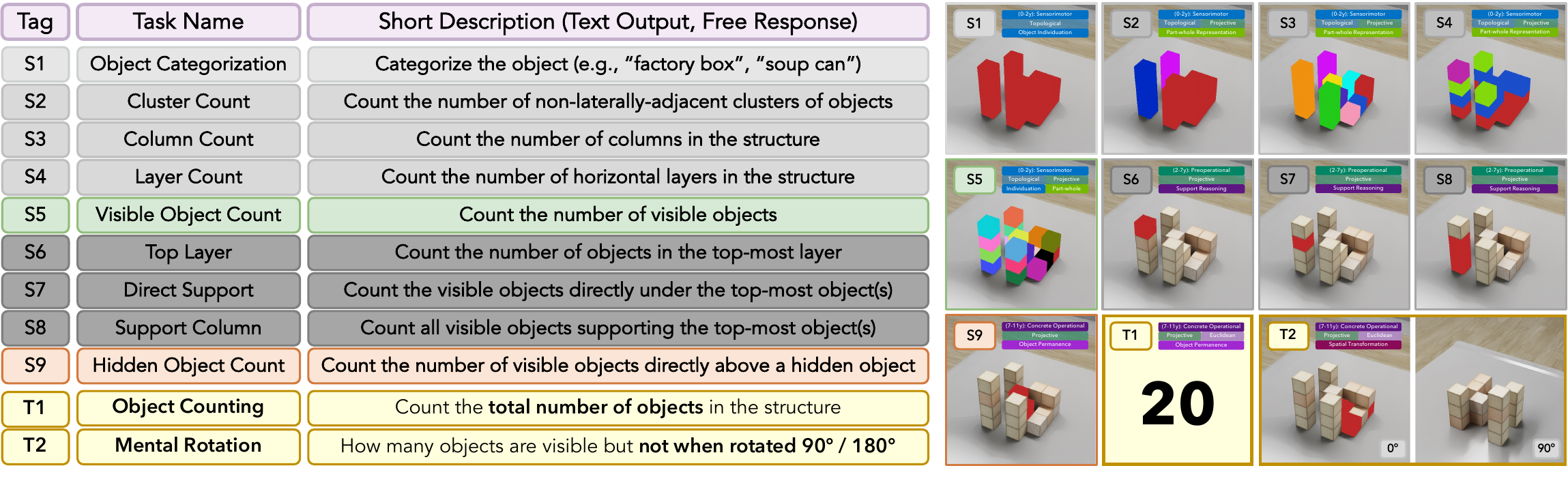}
\caption{[Left] The nine sub-tasks (S1--S9) and two target tasks (T1, T2) of \projectname{}. [Right] An example scene with the per-task ground-truth masks used to evaluate image-editing models overlaid. Each task panel is tagged with its stage(s) on three developmental frameworks laid out in Figure~\ref{fig:development_stages}.}
\label{fig:task_overview}
\end{figure}

We hypothesize that performing a target task recruits a hierarchy of lower-level perceptual-cognitive competencies, although a responder may also reach the answer without them. To make this structure explicit, we include and organize sub-tasks in a Piaget-Inhelder-inspired progression~\cite{piaget1969psychology} from simpler spatial representations to more integrated structural reasoning. We do not claim to operationalize Piagetian stage theory; rather, we use that tradition as a principled framework, motivating our choice of sub-tasks, their ordering and dependencies, and empirical tests.

We propose target tasks aligned with the concrete-operational level. \textit{Object Counting} (T1) is the primary target task: given an image of a stacked 3D structure, the model reports the total number of objects, including hidden objects required to support the visible structure under gravity. The additional target task is \textit{Mental Rotation} (T2): given two views of the same structure, the model reports which objects are visible in one view but not the other under $90^\circ$ and $180^\circ$ viewpoint rotation. We develop the sub-task hierarchy under \textit{Object Counting}, whose support and occlusion structure induces a richer set of internal dependencies, and treat \textit{Mental Rotation} as a complementary cross-view probe.

We propose nine sub-tasks to analyze the spatial reasoning structure of humans and models (Figure~\ref{fig:task_overview}). S1--S4 examine primitive perceptual grouping such as object individuation (S1), clustering (S2), and column and layer counting (S3, S4). S5 is to count all the visible objects. S6--S8 test structural reasoning that involves referencing. S6 is to identify object(s) at the top layer. S7 and S8 ask about the supporting structures for the top block(s): S7 finds directly supporting block(s), and S8 finds supporting column(s). S9 requires inferring hidden support: it counts the visible objects that rest directly on a hidden object.

For further analysis (Section~\ref{sec:analysis}), we define four hierarchical relations among the sub-tasks:

\begin{figure}[h]
\centering
\includegraphics[width=\textwidth]{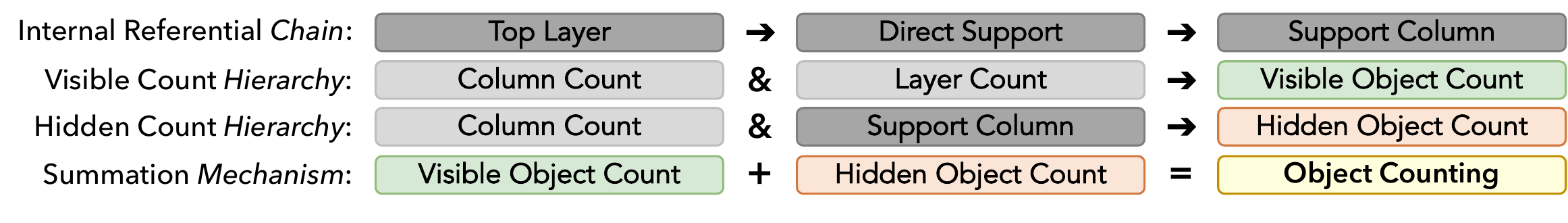}
\caption{We pre-specify these four sub-task relations for hierarchy dependency analysis in Section~\ref{sec:analysis}. The Visible and Hidden Count Hierarchies are diagnostic hypotheses.}
\label{fig:chains}
\end{figure}

The Summation Mechanism is an arithmetic composition, and the Internal Referential Chain is built into the task wording. The Visible and Hidden Count Hierarchies are pre-specified diagnostic hypotheses about how counting typically proceeds, not steps every counting strategy must take. We ground the visible branch in work showing that enumeration requires objects to be represented as distinct units~\citep{spelke1992origins} and that grouping cues can facilitate exact enumeration~\citep{starkey2014emergence}. We ground the hidden branch in object permanence~\citep{baillargeon1987object} and support-based physical reasoning~\citep{battaglia2013simulation}. Because sub-tasks are elicited independently, the benchmark tests whether a responder's answers are consistent with the decomposition, not whether the responder computed the count through it. S9 counts visible objects above hidden ones, so it falls short of the number of hidden objects when hidden objects stack on one another (17.9\% of benchmark views). We treat it as a probe of hidden-object inference and score the Summation Mechanism on errors rather than raw values (Section~\ref{sec:analysis}).

\paragraph{Modality-Specific Questions.}
We pose the tasks in three response formats: text free-response, multiple-choice (MCQ) with image options, and image editing. Outside the chain-of-thought checkpoints of Section~\ref{sec:training}, all tasks are queried independently, so each evaluates a competence in isolation rather than chained on prior answers. In the MCQ format, most wrong answers are constructed to represent specific named failure modes rather than as random perturbations. For sub-tasks, wrong-choice images are visually well-formed renderings of specific counting confusions (e.g., merging two columns into one, hallucinating a phantom block above the top layer). For the \textit{Object Counting} target task, wrong-answer integers are sampled around an offset pivot rather than centered on the correct answer, which reduces but does not eliminate a ``the answer is the median option'' heuristic. We additionally report MCQ results at four-choice and three-choice answer-space variants. In the image-editing format, the model directly edits the reference image to color the objects specified by the sub-task (e.g., highlighting the top layer or the visible objects), and writes the total \textit{Object Counting} as a natural number on the image for the \textit{Object Counting} target task. This format probes whether models can localize the queried structure in pixel space, not merely describe it. Full prompts, concept definitions, the per-format question wording, the wrong-answer taxonomy, and the prompt-validation protocol are in Appendix~\ref{app:prompts}.

\paragraph{Spatial-IQ Dataset.}
\label{sec:dataset_generation}
We procedurally generate scenes in NVIDIA Isaac Sim 5.1~\cite{nvidia_isaacsim_5_1}, each consisting of a foreground structure of multiple objects of a single type (textured cubes, factory boxes, soup cans, or potted meat cans) constrained to a $4 \times 4 \times 4$ voxel grid. Object positions are sampled to produce a target distribution of structure sizes that discriminates between models rather than concentrating on saturated easy or floor hard scenes, then consolidated under gravity into ground-supported columns. Hidden objects are identified through a pixel-level depth-buffer occlusion test that retains an object only if it is camera-visible or required to support a retained object above it, guaranteeing that every hidden object in the scene is logically required by physical support and never merely concealed behind a visible neighbor at an unrecoverable depth. Because we control the scene generation, we have voxel-level access to the underlying structure and can analytically derive ground truth for each task; image-editing ground truth is generated by re-rendering with task-specific colorings using the Replicator semantic-segmentation annotator~\cite{nvidia_isaac_replicator_420}. The full dataset comprises approximately 80{,}000 scenes. Our primary benchmark uses a 3{,}000-sample evaluation set, and training experiments (Section~\ref{sec:training}) use disjoint structures. Full sampling distributions, camera parameters, and a worked occlusion-test example are in Appendix~\ref{app:dataset_details}. The released benchmark view images and integer ground-truth annotations are available at \url{https://huggingface.co/datasets/patrickqrim/spatial-iq}, and code and scored model responses at \url{https://github.com/NVIDIA/Spatial-IQ}.

\section{Validating the \projectname{} Benchmark}
\label{sec:validation}

We first establish that \projectname{} serves as a meaningful benchmark. We evaluate models spanning three modalities. For text-output (text free-response and MCQ), we include the proprietary frontier models Gemini 3 Pro~\cite{deepmind2026gemini31pro}, GPT 5.4~\cite{singh2025openai}, and Claude Opus 4.6~\cite{anthropic2026claude}, and the open-weights frontier models Qwen3.5-27B~\cite{qwen35} (``Qwen''), Kimi K2.5~\cite{team2026kimi}, and GLM-4.6V-Flash~\cite{vteam2025glm45v}. We additionally evaluate Qwen2.5-VL-3B~\cite{bai2025qwen25vl} (``Qwen3B'') as a small-model anchor with VLA-0~\cite{goyal2025vla}. For image-editing, we evaluate Gemini 3.1 Flash Image~\cite{deepmind2026geminiflashimage}, Qwen-Image-Edit~\cite{wu2025qwen}, and HunyuanImage-Instruct~\cite{cao2025hunyuanimage}. In this section, we validate two key properties of our benchmark in the text free-response modality, with MCQ and image-editing analyses in Appendix~\ref{app:validation} and detailed MCQ wrong-answer analyses in Appendix~\ref{app:task_accuracy_modalities}.

\begin{figure}[t]
\centering
\includegraphics[width=\textwidth]{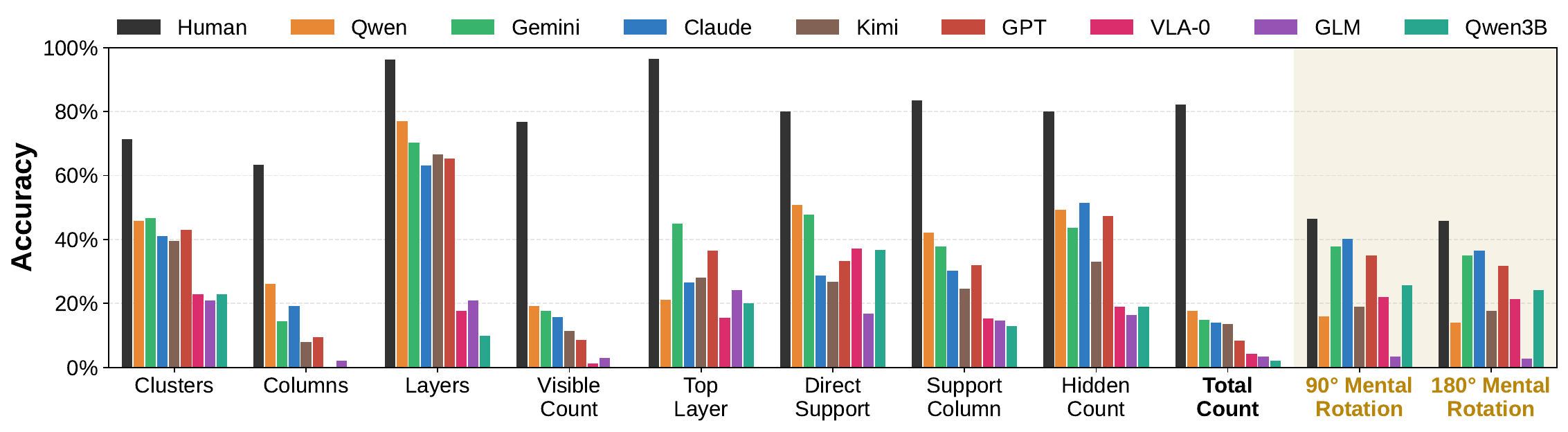}
\caption{Per-task accuracy, text free-response, ordered according to our task taxonomy (Figure~\ref{fig:task_overview}). Humans (black) perform well above all current models on \textit{Object Counting}, while model performance varies by an order of magnitude. Models are shown in order of performance on \textit{Object Counting}.}
\label{fig:task_accuracy_text}
\end{figure}

\paragraph{The target task is solvable by humans and meaningfully ranks models.}
\label{sec:validation_spread}
Figure~\ref{fig:task_accuracy_text} plots per-task accuracy for humans and text-output models. Humans reach 82.1\% on \textit{Object Counting}, demonstrating that the task is solvable, yet not at ceiling. Humans perform well above the best model (Qwen, 17.7\%), which in turn outperforms the worst model by an order of magnitude (Qwen2.5-VL-3B, 2.1\%). Paired McNemar tests with multiple-comparisons correction separate most adjacent models, although the near-tied groups Gemini, Claude and Kimi, and VLA-0 and GLM, are not statistically distinguishable (Table~\ref{tab:text_task_accuracy_full}). 

\begin{figure}[t]
\centering
\includegraphics[width=\textwidth]{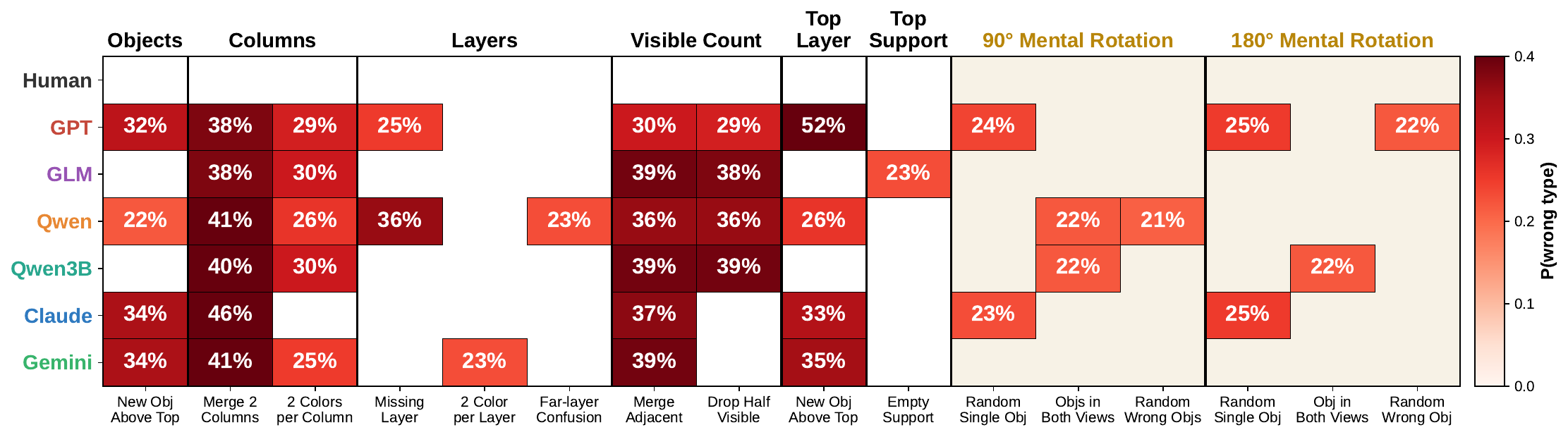}
\caption{Wrong-answer preferences in the five-choice MCQ condition. Each column denotes a wrong-answer type; each cell reports the raw percentage of responses assigned to that type. Only cells significant under a one-sided binomial test against the 20\% chance baseline are shown. Models exhibit task-dependent wrong-answer preferences, whereas humans show no significant preference.}
\vspace{-3mm}
\label{fig:mcq_wrong_answers}
\end{figure}

The MCQ responses provide valuable additional diagnostic information. With five choices (5CQ), every model performs at chance on \textit{Object Counting} (between 19.5\% and 20.7\% correct), while humans maintain good performance (86.2\%). However, models do not necessarily lack spatial competence: sub-task accuracy varies substantially (e.g., \textit{Visible Object Count} ranges from 20.3\% for GLM to 52.6\% for Claude). Rather, models make consistent errors (Figure~\ref{fig:mcq_wrong_answers}). \textit{Top Layer} and \textit{Object Categorization} are dominated by \texttt{phantom-block-above-top errors}, \textit{Column Count} by \texttt{merge-two-columns} errors, and \textit{Visible Object Count} by \texttt{merge-adjacent} and \texttt{drop-half-visible-blocks} errors. Reducing the options to three or four choices improves \textit{Object Counting} to above chance for some models, but the preference for certain kinds of wrong answers persists across answer-space conditions (see Figure~\ref{fig:mcq_m1_345} in Appendix~\ref{app:task_accuracy_modalities}).

\begin{figure}[t]
\centering
\includegraphics[width=\textwidth]{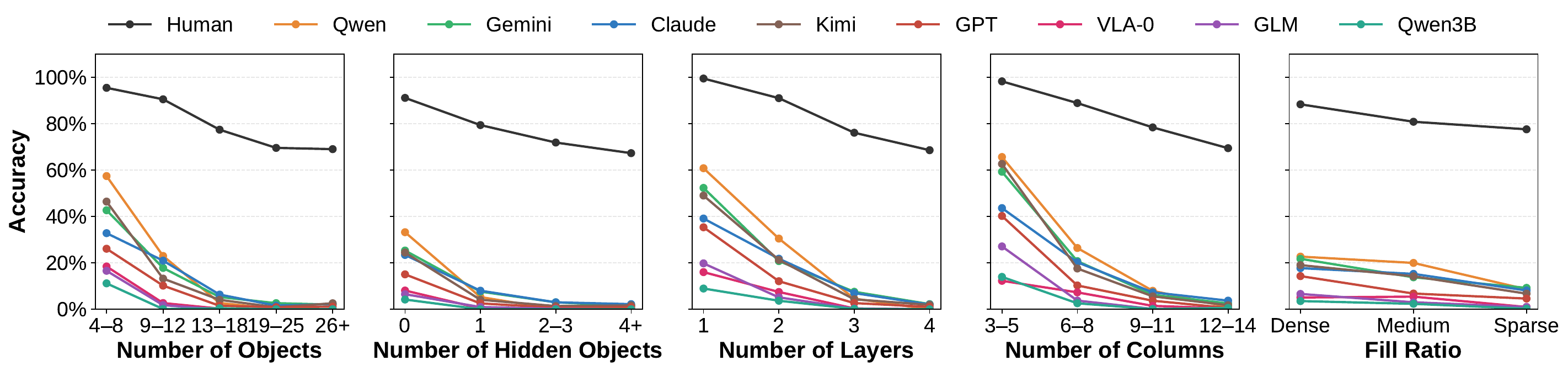}
\caption{\textit{Object Counting} accuracy as a function of five difficulty controls in the text modality. Total objects, hidden objects, layers, columns, and fill ratio all affect difficulty in the direction expected (more is harder, except fill ratio where denser is easier), for both humans and models.}
\vspace{-4mm}
\label{fig:difficulty_text}
\end{figure}

\paragraph{Success of difficulty manipulations.}
\label{sec:validation_difficulty}
Figure~\ref{fig:difficulty_text} plots \textit{Object Counting} accuracy along five dimensions: total number of objects, number of hidden objects, number of layers, number of columns, and the degree to which the structure fills the voxel grid (the \textit{fill ratio}, defined as $\mathrm{total\_blocks} / (\mathrm{num\_columns} \times \mathrm{num\_layers})$) for the text response modality.

These difficulty factors behave coherently for both humans and models. For humans, \textit{Object Counting} accuracy falls from 95.5\% for the fewest objects (4--8) to 69.1\% with the most ($\geq 26$), and from 91.2\% with no hidden objects to 67.3\% with four or more. Models show the same decline more steeply: pooled \textit{Object Counting} accuracy falls from 31.5\% on the easiest bin (4--8 objects) to 1.0\% on the hardest ($\geq 26$ objects). Denser scenes (high \textit{fill ratio}), on the other hand, lead to better performance than less dense scenes, as expected; as denser scenes approach a filled rectangular prism, one can derive the large number of objects by instead counting the small number of \textit{empty} voxels. 

The MCQ and image-generation modalities exhibit the same qualitative directional trends along these difficulty factors, albeit at different absolute performance levels.
We also included stimuli to test for robustness to camera perspective, azimuth offset, and object identity (Appendices~\ref{app:difficulty_modalities} and~\ref{app:object_results}). Stronger perspective lowers accuracy on \textit{Column Count} and \textit{Visible Object Count} while raising it on \textit{Top Layer}, a sign reversal that partly separates perception from structural reasoning. Object identity moves human accuracy by at most 8.4 points, far less than total object count does, and no zero-shot model exceeds 30\% on any object. As a further perception control, we re-rendered a subset of benchmark views with distinctly colored objects, which removes most of the individuation difficulty. On the same views, \textit{Object Counting} rises only from 14.3--16.5\% to 21.5--24.3\% for Qwen, Gemini and Claude, far below the human 82.1\% in the standard condition (Appendix~\ref{app:distinct_color}).

\section{Decomposing Model Performance on the \projectname{} Framework}
\label{sec:analysis}

\paragraph{Metrics for decomposition and hierarchical chain structures.} The analysis in this section is organized around the four analyses pre-specified in Figure~\ref{fig:chains}. This design follows the attribute-hierarchy convention in cognitive diagnosis, where pre-requisite structures are specified from theory before evaluation~\citep{tatsuoka1983rule, leighton2004attribute, templin2014hierarchical}. The two Count Hierarchies are diagnostic hypotheses, so the analyses below test consistency with them, not necessity. Three columns of Table~\ref{tab:text_summary} report raw accuracies (\texttt{Obj.\ Count}, \texttt{Visible}, \texttt{Hidden}), and the remaining four capture the pre-specified relations. The Internal Referential Chain (\texttt{Int.\ Ref.\ Chain}, along $\textit{Top Layer} \rightarrow \textit{Direct Support} \rightarrow \textit{Support Column}$) and the Visible and Hidden Count Hierarchies (\texttt{Vis.\ Hier.} and \texttt{Hid.\ Hier.}) report relation integrity $I(P \rightarrow C) = P(P \text{ correct} \mid C \text{ correct})$: given that the dependent (higher-level) sub-task is answered correctly, how often its prerequisite is also correct. The chain score pools violations over its two edges, and each hierarchy score is the mean of its two edge integrities. A relation is undefined (n/a) when its dependent sub-task is never answered correctly. The Summation Mechanism (\texttt{Sum.\ Mech.}) is, for the text modality, the share of scenes on which the error of the reported total equals the sum of the errors of the reported visible and hidden counts. For MCQ and image editing, which return no component counts, it is the conditional $P(\textit{Object Counting}{=}\texttt{true} \mid \textit{Visible Object Count}{=}\texttt{true},\ \textit{Hidden Object Count}{=}\texttt{true})$. Appendix~\ref{app:hierarchy_dependency} adds two scene-controlled dependence measures, the within-scene lift and the bundle-to-target coupling $r_{ws}$, which hold the scene fixed but not the responder. A two-way estimate with scene and responder fixed effects stays positive for both Count Hierarchies, at +20.2 pp (95\% view-clustered bootstrap CI [18.3, 22.3]) for Visible and +15.7 pp ([13.4, 17.8]) for Hidden.

\subsection{Text Modality}
\label{sec:analysis_text}

\begin{table}[t]
\centering
\small
\caption{Text modality evaluative summary (\% accuracy). See Section~\ref{sec:analysis} (Metrics) for definitions.
Per-column model rankings (humans excluded): \colorbox{rankone}{first}, \colorbox{ranktwo}{second}, \colorbox{rankthree}{third}.
}
\label{tab:text_summary}
\begin{adjustbox}{max width=\textwidth}
\begin{tabular}{l c c c c c c c}
\toprule
Responder & Obj.\ Count & Int.\ Ref.\ Chain & Visible & Vis.\ Hier. & Hidden & Hid.\ Hier. & Sum.\ Mech. \\
\midrule
Human   & 82.1 & 90.1 & 76.8 & 84.5 & 79.9 & 78.5 & 69.1 \\
\midrule
Qwen    & \cellcolor{rankone}17.7 & 50.7 & \cellcolor{rankone}19.2 & \cellcolor{rankone}78.2 & \cellcolor{ranktwo}49.2 & \cellcolor{rankone}47.9 & \cellcolor{rankone}64.2 \\
Gemini  & \cellcolor{ranktwo}14.8 & \cellcolor{rankone}64.5 & \cellcolor{ranktwo}17.7 & \cellcolor{ranktwo}56.7 & 43.6 & \cellcolor{ranktwo}34.6 & 18.6 \\
Claude  & \cellcolor{rankthree}14.1 & \cellcolor{rankthree}52.4 & \cellcolor{rankthree}15.7 & 48.5 & \cellcolor{rankone}51.4 & \cellcolor{rankthree}32.5 & \cellcolor{ranktwo}43.7 \\
Kimi    & 13.6 & 47.2 & 11.5 & 52.9 & 33.1 & 24.5 & 25.0 \\
GPT     & 8.3 & \cellcolor{ranktwo}54.3 & 8.7 & \cellcolor{rankthree}56.2 & \cellcolor{rankthree}47.4 & 28.4 & \cellcolor{rankthree}32.8 \\
VLA-0   & 4.2 & 16.7 & 1.2 & 0.0 & 19.0 & 7.4 & 19.9 \\
GLM     & 3.5 & 39.9 & 2.9 & 32.8 & 16.4 & 9.5 & 8.2 \\
Qwen3B  & 2.1 & 17.1 & 0.2 & 0.0 & 19.0 & 6.4 & 24.7 \\
\bottomrule
\end{tabular}
\end{adjustbox}
\end{table}

\begin{figure}[t]
\centering
\includegraphics[width=\textwidth]{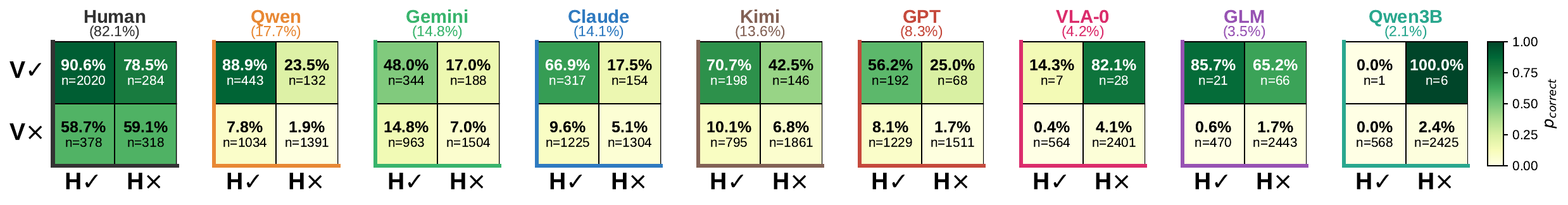}
\caption{\textit{Object Counting} accuracy, conditioned on correct performance on the \textit{Visible Object Count} (V) and \textit{Hidden Object Count} (H) tasks (text free-response modality). Rows and columns mark correct ($\checkmark$) or incorrect ($\times$) performance on the corresponding sub-task, and cell color and label indicate the conditional \textit{Object Counting} performance in \% over \textit{n} scenes. Full details in Appendix~\ref{app:text_mechanism}.}
\label{fig:text_mechanism}
\end{figure}

Figure~\ref{fig:text_mechanism} reports the \textit{Summation Mechanism} matrix cell structure for each text model, and Table~\ref{tab:text_summary} reports the model-level evaluative summary. Three observations emerge.

First, the human baseline shows the empirical signature of the \textit{Summation Mechanism} (Figure \ref{fig:text_mechanism}): \textit{Object Counting} accuracy is concentrated in the both \textit{V} and \textit{H} correct cell (90.6\%) and drops sharply once the visible count is missed (to about 59\% whether or not the hidden count is correct), with a \textit{Summation Mechanism} exact-match rate of 69.1\%. This suggests that the \textit{Summation Mechanism} captures a route humans often use.

Second, model recovery of the \textit{Summation Mechanism} is uneven and partially independent of headline accuracy. Qwen most closely approximates the human signature, with the largest both \textit{V} and \textit{H} correct (88.9\%) among all models and the highest exact-match rate (64.2\%). Gemini reaches comparable headline accuracy through a markedly weaker mechanism: only 48.0\% in the both \textit{V} and \textit{H} correct and 18.6\% exact match. Failures can also be located on the reasoning side: among trials with a correct visible count, the hidden answer is still wrong on 23--42\% of trials for Qwen, GPT, Claude, Gemini and Kimi, against 12.3\% for humans.

Third, the pre-specified hierarchy is largely preserved in the human baseline (\textit{Internal Referential Chain} 0.901, \textit{Visible Count Hierarchy} 0.845, \textit{Hidden Count Hierarchy} 0.785), and the model rows largely preserve this ordering: the \textit{Internal Referential Chain} is cleaner than the \textit{Hidden Count Hierarchy} for every model and than the \textit{Visible Count Hierarchy} for most (exceptions: Qwen, Kimi and GPT). The VLA-0 and Qwen3B rows additionally show that VLA-0 outperforms its Qwen2.5-VL-3B backbone on \textit{Object Counting} (4.2\% vs 2.1\%) and on the \textit{Hidden Count Hierarchy} despite being trained only on robot trajectory data, so action-grounded training may transfer positively to spatial reasoning. Full text modality analysis is reported in Appendix~\ref{app:text_full}.

\subsection{Multiple-Choice Modality}
\label{sec:analysis_mcq}

Table~\ref{tab:mcq_summary} reports the MCQ summary on a chance-adjusted scale, $\mathrm{adj\_acc} = (\mathrm{acc} - 1/k)/(1 - 1/k)$, so that three- (3CQ), four- (4CQ), and 5CQ columns ($k=3,4,5$) are directly comparable. Two patterns dominate. First, under 5CQ, humans appear to add correct visible and hidden blocks to get a correct total (both \textit{V} and \textit{H} correct at 83.6\%) while every model row collapses to chance level on \textit{Object Counting}, with the conditional correspondingly decreased. Reducing the number of options to four and three choices partially restores the expected \textit{Summation Mechanism}, but the recovery is uneven: GPT shows the full expected ordering (both correct highest, both wrong lowest) at both three and four choices, Claude and Gemini at one of the two sizes each, and Qwen improves on \textit{Object Counting} without recovering the component-conditioned pattern. Second, the chain-level pattern largely mirrors the text modality: the \textit{Internal Referential Chain} is cleaner than the \textit{Hidden Count Hierarchy} in every condition and than the \textit{Visible Count Hierarchy} except for Gemini. Full MCQ modality analysis is reported in Appendix~\ref{app:mcq_full}.

\begin{table}[t]
\centering
\small
\caption{Multiple-choice (MCQ) modality evaluative summary (chance-adjusted \% accuracy). Model results for 3CQ/4CQ/5CQ are shown side-by-side. Human performance was only measured for 5CQ.
}
\label{tab:mcq_summary}
\begin{adjustbox}{max width=\textwidth}
\begin{tabular}{l ccc ccc ccc ccc ccc ccc ccc}
\toprule
& \multicolumn{3}{c}{Obj.\ Count} & \multicolumn{3}{c}{Int.\ Ref.\ Chain} & \multicolumn{3}{c}{Visible} & \multicolumn{3}{c}{Vis.\ Hier.} & \multicolumn{3}{c}{Hidden} & \multicolumn{3}{c}{Hid.\ Hier.} & \multicolumn{3}{c}{Sum. Mech.} \\
\cmidrule(lr){2-4} \cmidrule(lr){5-7} \cmidrule(lr){8-10} \cmidrule(lr){11-13} \cmidrule(lr){14-16} \cmidrule(lr){17-19} \cmidrule(lr){20-22}
Responder & 3CQ & 4CQ & 5CQ & 3CQ & 4CQ & 5CQ & 3CQ & 4CQ & 5CQ & 3CQ & 4CQ & 5CQ & 3CQ & 4CQ & 5CQ & 3CQ & 4CQ & 5CQ & 3CQ & 4CQ & 5CQ \\
\midrule
Human  & --- & --- & 82.7 & --- & --- & 98.2 & --- & --- & 99.3 & --- & --- & 95.6 & --- & --- & 89.3 & --- & --- & 94.8 & --- & --- & 83.6 \\
\midrule
Claude & \cellcolor{rankone}23.7 & \cellcolor{rankone}17.8 & 0.0 & \cellcolor{rankone}30.4 & \cellcolor{rankone}28.0 & \cellcolor{rankone}22.8 & \cellcolor{ranktwo}44.6 & \cellcolor{rankone}43.6 & \cellcolor{rankone}40.7 & \cellcolor{ranktwo}16.2 & \cellcolor{ranktwo}19.2 & \cellcolor{ranktwo}18.1 & 8.0 & \cellcolor{ranktwo}10.7 & 11.7 & \cellcolor{ranktwo}20.3 & \cellcolor{rankone}23.8 & \cellcolor{rankone}20.4 & \cellcolor{rankone}26.8 & \cellcolor{rankone}26.1 & -2.2 \\
Gemini & 14.0 & 12.8 & -0.6 & \cellcolor{ranktwo}29.1 & \cellcolor{ranktwo}21.0 & 19.8 & \cellcolor{rankone}45.9 & \cellcolor{ranktwo}35.6 & \cellcolor{ranktwo}32.6 & \cellcolor{rankone}32.3 & \cellcolor{rankone}21.8 & \cellcolor{rankone}26.2 & \cellcolor{rankone}21.9 & \cellcolor{rankone}17.7 & \cellcolor{rankone}14.1 & \cellcolor{rankone}26.3 & 16.8 & 16.1 & \cellcolor{ranktwo}21.4 & \cellcolor{ranktwo}20.0 & -2.9 \\
GPT    & 7.9 & 5.9 & \cellcolor{rankone}0.9 & 17.3 & 19.8 & \cellcolor{ranktwo}21.1 & 42.1 & 24.3 & 24.0 & 12.3 & 7.7 & 9.2 & \cellcolor{ranktwo}15.0 & 10.4 & \cellcolor{ranktwo}12.6 & 16.6 & \cellcolor{ranktwo}19.0 & \cellcolor{ranktwo}17.5 & 12.9 & 16.9 & \cellcolor{rankone}2.2 \\
Qwen   & \cellcolor{ranktwo}20.5 & \cellcolor{ranktwo}14.8 & \cellcolor{ranktwo}0.2 & 23.7 & 16.4 & 11.1 & 36.5 & 17.3 & 8.0 & 1.5 & -2.5 & -1.5 & -20.9 & -12.6 & -2.0 & 14.8 & 8.7 & 6.4 & 7.4 & 5.7 & \cellcolor{ranktwo}1.0 \\
\bottomrule
\end{tabular}
\end{adjustbox}
\end{table}

\subsection{Image Modality}
\label{sec:analysis_image}

The image-editing modality probes the target task and sub-tasks, but asks each model to answer by coloring the reference image (see Figure~\ref{fig:task_overview}).
Table~\ref{tab:image_summary} reports the evaluative summary.

The qualitative pattern from text and MCQ replicates: the \textit{Internal Referential Chain} remains the most format-stable backbone of the proposed hierarchy, and target-task success and local hierarchy adherence are partially dissociable. Gemini Flash Image leads on \textit{Object Counting} and on the \textit{Summation Mechanism} conditional. Qwen-Image-Edit gives the clearest example of the local-versus-target dissociation, with the strongest \textit{Internal Referential Chain} integrity in the modality (0.676) but raw \textit{Visible Object Count} accuracy of only 3.3\%. Absolute target-task coupling is weaker than in text or MCQ; full analysis of image-editing models is reported in Appendix~\ref{app:image_full}.

\begin{table}[t]
\centering
\small
\caption{Image-editing modality evaluative summary (\% accuracy). Columns parallel Tables~\ref{tab:text_summary} and~\ref{tab:mcq_summary}.
}
\label{tab:image_summary}
\begin{adjustbox}{max width=\textwidth}
\begin{tabular}{l c c c c c c c}
\toprule
Responder & Obj.\ Count & Int.\ Ref.\ Chain & Visible & Vis.\ Hier. & Hidden & Hid.\ Hier. & Sum.\ Mech. \\
\midrule
Gemini Flash Image    & \cellcolor{rankone}14.7 & \cellcolor{ranktwo}53.7 & \cellcolor{rankone}36.1 & \cellcolor{ranktwo}37.0 & \cellcolor{rankone}46.5 & \cellcolor{rankone}36.5 & \cellcolor{rankone}29.5 \\
Qwen-Image-Edit       & \cellcolor{ranktwo}12.0 & \cellcolor{rankone}67.6 & 3.3 & \cellcolor{rankone}58.7 & \cellcolor{ranktwo}38.7 & 11.2 & \cellcolor{ranktwo}23.8 \\
HunyuanImage-Instruct & 5.9 & 50.1 & \cellcolor{ranktwo}10.8 & 32.2 & 30.5 & \cellcolor{ranktwo}19.0 & 9.4 \\
\bottomrule
\end{tabular}
\end{adjustbox}
\end{table}

\subsection{Mental Rotation as a Complementary Probe}
\label{sec:mental_rotation_complement}

Our second target task addresses a distinct facet of spatial cognition: viewpoint transformation \citep{tarr1989mental,shepard1971mental}. Because the sub-tasks are designed for visible-object aggregation, hidden-object inference, and their Summation Mechanism for the total \textit{Object Count}, the hierarchy is not expected to transfer wholesale to mental rotation tasks. Empirically, we observe this separation: the proposed hierarchy aligns much more strongly with \textit{Object Counting} than with \textit{Mental Rotation}, while the two rotation tasks align primarily with each other for humans and half of the models. Humans perform worse on \textit{Mental Rotation} ($46.4\%$ at $90^\circ$ and $45.8\%$ at $180^\circ$) than on \textit{Object Counting}, which limits the headroom available for distinguishing between models. Cross-view dependency analyses, bundle-to-target coupling, and per-model \textit{Mental Rotation} accuracies are in Appendix~\ref{app:mental_rotation}.

\section{Does Training on \projectname{} Sub-Tasks Improve Spatial Reasoning?}
\label{sec:training}

The structure of the \projectname{} benchmark suggests a natural training signal: a model trained to walk through the sub-task decomposition before committing to a final count should learn the prerequisite competencies that the additive decomposition relies on, and should generalize better than a model trained on the target task alone. We test this hypothesis on Qwen2.5-VL-7B-Instruct and Qwen2.5-VL-32B-Instruct~\citep{bai2025qwen25vl} under three training conditions, in addition to a zero-shot baseline. \textbf{SFT-plain} is supervised fine-tuning on the integer total alone, with no decomposition. \textbf{SFT-CoT} is supervised fine-tuning on a chain-of-thought target~\citep{lightman2023let} that lists the sub-task answers in the order required to compute the total additively, terminating with the integer wrapped in answer tags. \textbf{DAPO-tight}, in the spirit of DeepSeek-R1~\citep{guo2025deepseek}, initializes from the SFT-CoT checkpoint and continues with reinforcement learning with verifiable rewards~\citep{lambert2024tulu} (RLVR) on a 50\% subset of the training data, optimizing GRPO~\citep{shao2024deepseekmath} under the DAPO recipe~\citep{yu2025dapo}. All training is conducted in the text-output modality on textured cubes only, with the train/benchmark split enforced at the structure level, and evaluation covers all four object categories (Appendix~\ref{app:training_results}). The full chain-of-thought trace specification and hyperparameters are in Appendix~\ref{app:training}.

\begin{table}[t]
\centering
\small
\caption{Text modality evaluative summary (\% accuracy) for two trained Qwen2.5-VL checkpoints. Values in parentheses indicate absolute point gains relative to the corresponding zero-shot backbone. n/a: undefined, because the relation's dependent sub-task is never answered correctly. SFT-CoT and DAPO-tight sub-task columns are read from one chain-of-thought response (Appendix~\ref{app:training_setup}).
}
\label{tab:training_summary}
\begin{adjustbox}{max width=\textwidth}
\begin{tabular}{l l l l l l l l}
\toprule
Responder & Obj.\ Count & Int.\ Ref.\ Chain & Visible & Vis.\ Hier. & Hidden & Hid.\ Hier. & Sum.\ Mech. \\
\midrule
Human                       & 82.1 & 90.1 & 76.8 & 84.5 & 79.9 & 78.5 & 69.1 \\
\midrule
DAPO-tight 7B               & \cellcolor{rankone}50.7 (+46.4) & \cellcolor{rankone}98.8 (+50.7) & \cellcolor{rankone}51.7 (+48.3) & \cellcolor{rankone}87.5 (+79.2) & \cellcolor{ranktwo}76.0 (+30.7) & \cellcolor{rankone}74.9 (+55.5) & \cellcolor{rankone}90.0 (+45.3) \\
SFT-CoT 7B                  & \cellcolor{ranktwo}40.8 (+36.5) & \cellcolor{ranktwo}97.8 (+49.7) & \cellcolor{ranktwo}41.9 (+38.5) & \cellcolor{ranktwo}85.0 (+76.7) & \cellcolor{rankone}76.7 (+31.4) & \cellcolor{ranktwo}68.3 (+48.9) & \cellcolor{ranktwo}89.6 (+44.9) \\
SFT-plain 7B                & 39.7 (+35.4) & n/a & 29.2 (+25.8) & \phantom{0}7.9 (-0.4) & \phantom{0}0.0  (-45.3) & n/a & \phantom{0}0.0  (-44.7) \\
Qwen2.5-VL-7B (zero-shot)   & \phantom{0}4.3           & 48.1 & \phantom{0}3.4           & \phantom{0}8.3           & 45.3 & 19.4 & 44.7 \\
\midrule
DAPO-tight 32B              & \cellcolor{rankone}62.6 (+59.7) & \cellcolor{rankone}99.3 (+60.5) & \cellcolor{rankone}64.5 (+62.6) & \cellcolor{ranktwo}88.4 (+67.7) & \cellcolor{ranktwo}78.2 (+27.8) & \cellcolor{rankone}82.1 (+61.5) & \cellcolor{rankone}93.3 (+42.3) \\
SFT-CoT 32B                 & \cellcolor{ranktwo}46.7 (+43.8) & \cellcolor{ranktwo}99.2 (+60.4) & \cellcolor{ranktwo}48.9 (+47.0) & \cellcolor{rankone}89.7 (+69.0) & \cellcolor{rankone}79.2 (+28.8) & \cellcolor{ranktwo}77.2 (+56.6) & \cellcolor{ranktwo}92.9 (+41.9) \\
SFT-plain 32B               & 40.1 (+37.2) & n/a & 30.9 (+29.0) & \phantom{0}9.8 (-10.9) & \phantom{0}0.0  (-50.4) & n/a & \phantom{0}0.0  (-51.0) \\
Qwen2.5-VL-32B (zero-shot)  & \phantom{0}2.9           & 38.8 & \phantom{0}1.9           & 20.7          & 50.4 & 20.6 & 51.0 \\
\bottomrule
\end{tabular}
\end{adjustbox}
\end{table}

The decomposition-aware checkpoints (SFT-CoT and DAPO-tight) narrow the gap to the human baseline on \textit{Object Counting} and also raise the sub-task columns.
We highlight the main observations below and report further ones in Appendix~\ref{app:training_results}, including that RLVR concentrates the policy rather than expanding the candidate set, and that the gains extend to unseen object categories.

\paragraph{Supervising the decomposition helps, and the gain compounds with scale.}
At 7B, SFT-CoT and SFT-plain are essentially tied on \textit{Object Counting} accuracy (1.1-point gap), but at 32B the decomposition produces a 6.6-point gap in its favor. More importantly, the two diverge sharply on the hierarchy columns. SFT-plain never answers \textit{Layer Count}, \textit{Direct Support}, \textit{Support Column} or \textit{Hidden Object Count} correctly, so its \textit{Internal Referential Chain} and \textit{Hidden Count Hierarchy} are undefined. Its \textit{Visible Count Hierarchy} stays at or below the zero-shot backbone (7.9 vs 8.3 at 7B, 9.8 vs 20.7 at 32B), and its \textit{Summation Mechanism} is 0. It reaches correct totals without the component answers they presuppose, repeating the total on 88\% (32B) and 93\% (7B) of sub-task queries. Because SFT-plain is trained only to emit a total, these scores partly reflect how the sub-tasks are elicited, so we read them as a response-level dissociation and not as evidence about internal computation. SFT-CoT preserves all four hierarchy relations, with its sub-task scores read from one chain-of-thought response. Together, these results describe the decomposition as a training signal that jointly improves target and sub-task competence, not as a proven causal prerequisite.

\paragraph{A controlled ablation separates labels from algorithm.}
SFT-CoT, SFT-plain and DAPO-tight differ in algorithm as well as in labels, so we also varied one factor at a time on Qwen2.5-VL-7B, holding the DAPO-tight recipe and the validation split fixed. CoT cold starts beat plain cold starts after identical RL at every split of cold-start and RL data (10/90: 29.0\% vs 19.6\%, 50/50: 39.6\% vs 15.4\%, 100/50: 37.2\% vs 32.8\%), and RL adds 8.0 to 9.0 points to CoT cold starts against 3.0 to 4.9 points to plain ones (Appendix~\ref{app:cold_start}).

\paragraph{Reinforcement learning with verifiable rewards adds substantial gain beyond SFT-CoT.}
DAPO-tight beats SFT-CoT by 9.9 absolute points at 7B and 15.9 at 32B, with the 32B DAPO-tight model reaching 62.6\% \textit{Object Counting} accuracy where the zero-shot backbone achieves 2.9\%. The reward covers only the integer total and the answer format, so the gains over SFT-CoT come from concentrating the policy onto chain-of-thought trajectories that lead to correct totals, not from new sub-task supervision. A replication on Gemma-4-31B~\citep{google2026gemma4} follows the same ordering on the validation split (Appendix~\ref{app:training_results}).

\paragraph{Two further tests of dependence on the decomposition.}
Supplying the correct intermediate values at inference raises \textit{Object Counting} from 14--18\% to 77--81\% across three models. In answer-masked training, we supervise Qwen2.5-VL-7B on the sub-task chain with the final total masked, and the total implied by the model's own reasoning is correct on 25.6\% of validation views, against 29.2\% for supervised SFT-CoT and 5.3\% for the backbone (Appendix~\ref{app:dependence}). The total is defined as the sum of its visible and hidden components, so both tests show sufficiency at most, and neither validates the two Count Hierarchies.

\section{Discussion and Future Work}
\label{sec:discussion}

\projectname{} reframes spatial-intelligence evaluation as a hierarchical decomposition and uses the same hierarchy as a diagnostic instrument and as a training signal. Top models often succeed on \textit{Object Counting} while failing the underlying sub-tasks. Chain-of-thought supervision over the hierarchy combined with verifiable-reward reinforcement learning jointly improves target and sub-task competence and narrows the gap to humans, although we do not establish the decomposition as a causal prerequisite. As an initial sim-to-real probe, Qwen2.5-VL-7B checkpoints trained only on synthetic renders reach 38.0\% on 50 photographed box structures after DAPO-tight, against 4.0\% at zero shot. Across five VLAs, \projectname{} accuracy shows a suggestive rank agreement with published LIBERO~\citep{liu2023libero} success rates (Spearman $\rho = 0.80$, $n = 5$, not significant). Both results are preliminary (Appendix~\ref{app:external}).

One limitation is that cross-modality comparisons are approximate, since response format changes the answer-selection demands (Appendix~\ref{app:cross_modality}). The benchmark uses synthetic structures on a small $4 \times 4 \times 4$ voxel grid with limited object category, appearance, and geometry diversity, and perception and reasoning remain partly entangled (Appendix~\ref{app:limitations}). Future work includes mixed-object and irregular structures, varied object sizes, more target tasks and support structures, larger real-image sets, and a broader investigation of whether action-grounded training in vision-language-action models systematically benefits spatial reasoning.

\begin{ack}
This work was funded by NVIDIA. Alex Wong was supported by the grant SFI-AN-NC-SCN-00007276 from the Simons Foundation International as part of the SCENE collaboration and by NSF under the NSF AI Institute for Edge Computing and Systems (Athena), Grant \#2625580. Authors affiliated with NVIDIA are NVIDIA employees. Patrick Rim is a cofounder and CEO of Build AI.
\end{ack}

{
    \small
    \bibliographystyle{plain}
    \bibliography{main}

@inproceedings{yang2025thinking,
  title={Thinking in space: How multimodal large language models see, remember, and recall spaces},
  author={Yang, Jihan and Yang, Shusheng and Gupta, Anjali W and Han, Rilyn and Fei-Fei, Li and Xie, Saining},
  booktitle={Proceedings of the Computer Vision and Pattern Recognition Conference},
  pages={10632--10643},
  year={2025}
}

@article{yang2025mmsi,
  title={{MMSI-Bench}: A benchmark for multi-image spatial intelligence},
  author={Yang, Sihan and Xu, Runsen and Xie, Yiman and Yang, Sizhe and Li, Mo and Lin, Jingli and Zhu, Chenming and Chen, Xiaochen and Duan, Haodong and Yue, Xiangyu and others},
  journal={arXiv preprint arXiv:2505.23764},
  year={2025}
}

@article{xu2025spatialbench,
  title={{SpatialBench}: Benchmarking multimodal large language models for spatial cognition},
  author={Xu, Peiran and Wang, Sudong and Zhu, Yao and Li, Jianing and Qi, Gege and Zhang, Yunjian},
  journal={arXiv preprint arXiv:2511.21471},
  year={2025}
}

@article{chen2026babyvision,
  title={{BabyVision}: Visual Reasoning Beyond Language},
  author={Chen, Liang and Xie, Weichu and Liang, Yiyan and He, Hongfeng and Zhao, Hans and Yang, Zhibo and Huang, Zhiqi and Wu, Haoning and Lu, Haoyu and Bao, Yiping and others},
  journal={arXiv preprint arXiv:2601.06521},
  year={2026}
}

@article{ray2024sat,
  title={{SAT}: Dynamic spatial aptitude training for multimodal language models},
  author={Ray, Arijit and Duan, Jiafei and Brown, Ellis and Tan, Reuben and Bashkirova, Dina and Hendrix, Rose and Ehsani, Kiana and Kembhavi, Aniruddha and Plummer, Bryan A and Krishna, Ranjay and others},
  journal={arXiv preprint arXiv:2412.07755},
  year={2024}
}

@article{he2025egoexobench,
  title={{EgoExoBench}: A benchmark for first-and third-person view video understanding in {MLLMs}},
  author={He, Yuping and Huang, Yifei and Chen, Guo and Pei, Baoqi and Xu, Jilan and Lu, Tong and Pang, Jiangmiao},
  journal={arXiv preprint arXiv:2507.18342},
  year={2025}
}

@article{li2025worldmodelbench,
  title={{WorldModelBench}: Judging video generation models as world models},
  author={Li, Dacheng and Fang, Yunhao and Chen, Yukang and Yang, Shuo and Cao, Shiyi and Wong, Justin and Luo, Michael and Wang, Xiaolong and Yin, Hongxu and Gonzalez, Joseph E and others},
  journal={arXiv preprint arXiv:2502.20694},
  year={2025}
}

@article{meng2024towards,
  title={Towards world simulator: Crafting physical commonsense-based benchmark for video generation},
  author={Meng, Fanqing and Liao, Jiaqi and Tan, Xinyu and Shao, Wenqi and Lu, Quanfeng and Zhang, Kaipeng and Cheng, Yu and Li, Dianqi and Qiao, Yu and Luo, Ping},
  journal={arXiv preprint arXiv:2410.05363},
  year={2024}
}

@inproceedings{duan2025worldscore,
  title={{WorldScore}: A unified evaluation benchmark for world generation},
  author={Duan, Haoyi and Yu, Hong-Xing and Chen, Sirui and Fei-Fei, Li and Wu, Jiajun},
  booktitle={Proceedings of the IEEE/CVF International Conference on Computer Vision},
  pages={27713--27724},
  year={2025}
}

@article{qin2024worldsimbench,
  title={{WorldSimBench}: Towards video generation models as world simulators},
  author={Qin, Yiran and Shi, Zhelun and Yu, Jiwen and Wang, Xijun and Zhou, Enshen and Li, Lijun and Yin, Zhenfei and Liu, Xihui and Sheng, Lu and Shao, Jing and others},
  journal={arXiv preprint arXiv:2410.18072},
  year={2024}
}

@article{singh2025openai,
  title={{OpenAI GPT-5} system card},
  author={Singh, Aaditya and Fry, Adam and Perelman, Adam and Tart, Adam and Ganesh, Adi and El-Kishky, Ahmed and McLaughlin, Aidan and Low, Aiden and Ostrow, AJ and Ananthram, Akhila and others},
  journal={arXiv preprint arXiv:2601.03267},
  year={2025}
}

@misc{deepmind2026gemini31pro,
  author = {{Google DeepMind}},
  title = {{Gemini 3 Pro}},
  url = {https://deepmind.google/technologies/gemini/},
  year = {2026},
  note = {Multimodal large language model}
}

@misc{anthropic2026claude,
  author = {{Anthropic}},
  title = {{Claude Opus 4.6}},
  url = {https://claude.ai},
  year = {2026},
  note = {Large language model}
}

@misc{vteam2025glm45v,
  title={{GLM-4.5V} and {GLM-4.1V-Thinking}: Towards Versatile Multimodal Reasoning with Scalable Reinforcement Learning},
  author={{V Team} and Hong, Wenyi and Yu, Wenmeng and others},
  howpublished={arXiv:2507.01006},
  year={2025},
  eprint={2507.01006},
  archivePrefix={arXiv},
  primaryClass={cs.CV}
}

@article{team2026kimi,
  title={{Kimi K2.5}: Visual Agentic Intelligence},
  author={Team, Kimi and Bai, Tongtong and Bai, Yifan and Bao, Yiping and Cai, SH and Cao, Yuan and Charles, Y and Che, HS and Chen, Cheng and Chen, Guanduo and others},
  journal={arXiv preprint arXiv:2602.02276},
  year={2026}
}

@article{goyal2025vla,
  title={{VLA-0}: Building state-of-the-art {VLAs} with zero modification},
  author={Goyal, Ankit and Hadfield, Hugo and Yang, Xuning and Blukis, Valts and Ramos, Fabio},
  journal={arXiv preprint arXiv:2510.13054},
  year={2025}
}

@article{wu2025qwen,
  title={{Qwen-Image} technical report},
  author={Wu, Chenfei and Li, Jiahao and Zhou, Jingren and Lin, Junyang and Gao, Kaiyuan and Yan, Kun and Yin, Sheng-ming and Bai, Shuai and Xu, Xiao and Chen, Yilei and others},
  journal={arXiv preprint arXiv:2508.02324},
  year={2025}
}

@article{cao2025hunyuanimage,
  title={{HunyuanImage 3.0} technical report},
  author={Cao, Siyu and Chen, Hangting and Chen, Peng and Cheng, Yiji and Cui, Yutao and Deng, Xinchi and Dong, Ying and Gong, Kipper and Gu, Tianpeng and Gu, Xiusen and others},
  journal={arXiv preprint arXiv:2509.23951},
  year={2025}
}

@misc{deepmind2026geminiflashimage,
  author = {{Google DeepMind}},
  title = {{Gemini 3.1 Flash Image}},
  url = {https://deepmind.google/technologies/gemini/},
  year = {2026},
  note = {Generative text-to-image model (Internal codename: Nano Banana 2)}
}

@misc{nvidia_isaacsim_5_1,
  author = {{NVIDIA}},
  title = {{NVIDIA Isaac Sim}},
  url = {https://developer.nvidia.com/isaac/sim},
  version = {5.1},
  year = {2024},
  note = {Robotics simulation platform}
}

@misc{nvidia_isaac_replicator_420,
  author = {{NVIDIA}},
  title = {{Replicator Tutorials --- Omniverse Isaac Sim 4.2.0 Documentation}},
  howpublished = {\url{https://docs.isaacsim.omniverse.nvidia.com/4.2.0/replicator_tutorials/index.html}},
  year = {2024},
  note = {Accessed: April 27, 2026}
}

@book{piaget1956child,
  title={The Child's Conception of Space},
  author={Piaget, Jean and Inhelder, B{\"a}rbel},
  year={1956},
  publisher={Routledge \& Kegan Paul},
  address={London},
  note={Translated by F. J. Langdon and J. L. Lunzer; original French edition 1948}
}

@book{piaget1969psychology,
  title={The Psychology of the Child},
  author={Piaget, Jean and Inhelder, B{\"a}rbel},
  year={1969},
  publisher={Basic Books},
  address={New York}
}

@article{yu2025dapo,
  title={{DAPO}: An open-source {LLM} reinforcement learning system at scale},
  author={Yu, Qiying and Zhang, Zheng and Zhu, Ruofei and Yuan, Yufeng and Zuo, Xiaochen and Yue, Yu and Dai, Weinan and Fan, Tiantian and Liu, Gaohong and Liu, Lingjun and others},
  journal={arXiv preprint arXiv:2503.14476},
  year={2025}
}

@article{shepard1971mental,
  title={Mental rotation of three-dimensional objects},
  author={Shepard, Roger N. and Metzler, Jacqueline},
  journal={Science},
  volume={171},
  number={3972},
  pages={701--703},
  year={1971}
}

@inproceedings{xiao2026spatialtree,
  title={{SpatialTree}: How spatial abilities branch out in {MLLMs}},
  author={Xiao, Yuxi and Li, Longfei and Yan, Shen and Liu, Xinhang and Peng, Sida and Wei, Yunchao and Zhou, Xiaowei and Kang, Bingyi},
  booktitle={The First Workshop on Efficient Spatial Reasoning},
  year={2026}
}

@article{zhan20263viewsense,
  title={{3ViewSense}: Spatial and Mental Perspective Reasoning from Orthographic Views in Vision-Language Models},
  author={Zhan, Shaoxiong and Lai, Yanlin and Liu, Zheng and Lin, Hai and Li, Shen and Cai, Xiaodong and Lin, Zijian and Huang, Wen and Zheng, Hai-Tao},
  journal={arXiv preprint arXiv:2603.07751},
  year={2026}
}

@article{Gong2025SpaCE10AC,
  title={{SpaCE-10}: A Comprehensive Benchmark for Multimodal Large Language Models in Compositional Spatial Intelligence},
  author={Ziyang Gong and Wenhao Li and Olivera Mart{\'i}nez Ma and Songyuan Li and Jiayi Ji and Xue Yang and Gen Luo and Junchi Yan and Rongrong Ji},
  journal={ArXiv},
  year={2025},
  volume={abs/2506.07966},
  url={https://api.semanticscholar.org/CorpusID:279251735}
}

@inproceedings{koenderink2011vision,
  title={Vision as a user interface},
  author={Koenderink, Jan},
  booktitle={Human vision and electronic imaging XVI},
  volume={7865},
  pages={18--30},
  year={2011},
  organization={SPIE}
}

@article{battaglia2013simulation,
  title={Simulation as an engine of physical scene understanding},
  author={Battaglia, Peter W and Hamrick, Jessica B and Tenenbaum, Joshua B},
  journal={Proceedings of the national academy of sciences},
  volume={110},
  number={45},
  pages={18327--18332},
  year={2013},
  publisher={National Academy of Sciences}
}

@article{fischer2016functional,
  title={Functional neuroanatomy of intuitive physical inference},
  author={Fischer, Jason and Mikhael, John G and Tenenbaum, Joshua B and Kanwisher, Nancy},
  journal={Proceedings of the national academy of sciences},
  volume={113},
  number={34},
  pages={E5072--E5081},
  year={2016},
  publisher={National Academy of Sciences}
}

@article{smith2013sources,
  title={Sources of uncertainty in intuitive physics},
  author={Smith, Kevin A and Vul, Edward},
  journal={Topics in cognitive science},
  volume={5},
  number={1},
  pages={185--199},
  year={2013},
  publisher={Wiley Online Library}
}

@book{kaufman1983kaufman,
  author={Kaufman, Alan S. and Kaufman, Nadeen L.},
  title={Kaufman Assessment Battery for Children ({K-ABC})},
  publisher={American Guidance Service},
  address={Circle Pines, MN},
  year={1983}
}

@article{drozdick2018kaufman,
  title={The {Kaufman Assessment Battery for Children---Second Edition} and {KABC-II Normative Update}},
  author={Drozdick, Lisa Whipple and Singer, Jennie Kaufman and Lichtenberger, Elizabeth O and others},
  journal={Contemporary intellectual assessment: Theories, tests, and issues},
  pages={333--359},
  year={2018},
  publisher={Guilford Publications}
}

@inproceedings{dumery2025counting,
  title={Counting Stacked Objects},
  author={Dumery, Corentin and Ett{\'e}, Noa and Fan, Aoxiang and Li, Ren and Xu, Jingyi and Le, Hieu and Fua, Pascal},
  booktitle={Proceedings of the IEEE/CVF International Conference on Computer Vision},
  pages={19774--19783},
  year={2025}
}

@article{baillargeon1987object,
  title={Object permanence in 3$1/2$-and 4$1/2$-month-old infants.},
  author={Baillargeon, Renee},
  journal={Developmental psychology},
  volume={23},
  number={5},
  pages={655},
  year={1987},
  publisher={American Psychological Association}
}

@article{spelke1992origins,
  title={Origins of knowledge.},
  author={Spelke, Elizabeth S and Breinlinger, Karen and Macomber, Janet and Jacobson, Kristen},
  journal={Psychological review},
  volume={99},
  number={4},
  pages={605},
  year={1992},
  publisher={American Psychological Association}
}

@article{tatsuoka1983rule,
  title={Rule space: An approach for dealing with misconceptions based on item response theory},
  author={Tatsuoka, Kikumi K},
  journal={Journal of educational measurement},
  pages={345--354},
  year={1983},
  publisher={JSTOR}
}

@article{leighton2004attribute,
  title={The attribute hierarchy method for cognitive assessment: A variation on {Tatsuoka's} rule-space approach},
  author={Leighton, Jacqueline P and Gierl, Mark J and Hunka, Stephen M},
  journal={Journal of educational measurement},
  volume={41},
  number={3},
  pages={205--237},
  year={2004},
  publisher={Wiley Online Library}
}

@article{templin2014hierarchical,
  title={Hierarchical diagnostic classification models: A family of models for estimating and testing attribute hierarchies},
  author={Templin, Jonathan and Bradshaw, Laine},
  journal={Psychometrika},
  volume={79},
  number={2},
  pages={317--339},
  year={2014},
  publisher={Springer US}
}

@article{tarr1989mental,
  title={Mental rotation and orientation-dependence in shape recognition},
  author={Tarr, Michael J and Pinker, Steven},
  journal={Cognitive psychology},
  volume={21},
  number={2},
  pages={233--282},
  year={1989},
  publisher={Elsevier}
}

@inproceedings{lightman2023let,
  title={Let's verify step by step},
  author={Lightman, Hunter and Kosaraju, Vineet and Burda, Yuri and Edwards, Harrison and Baker, Bowen and Lee, Teddy and Leike, Jan and Schulman, John and Sutskever, Ilya and Cobbe, Karl},
  booktitle={The twelfth international conference on learning representations},
  year={2024}
}

@inproceedings{pothiraj2025capture,
  title={{CAPTURe}: Evaluating spatial reasoning in vision language models via occluded object counting},
  author={Pothiraj, Atin and Stengel-Eskin, Elias and Cho, Jaemin and Bansal, Mohit},
  booktitle={Proceedings of the IEEE/CVF International Conference on Computer Vision},
  pages={8001--8010},
  year={2025}
}

@article{shao2024deepseekmath,
  title={{DeepSeekMath}: Pushing the limits of mathematical reasoning in open language models},
  author={Shao, Zhihong and Wang, Peiyi and Zhu, Qihao and Xu, Runxin and Song, Junxiao and Bi, Xiao and Zhang, Haowei and Zhang, Mingchuan and Li, YK and Wu, Yang and others},
  journal={arXiv preprint arXiv:2402.03300},
  year={2024}
}

@inproceedings{li2026spatialladder,
  title={{SpatialLadder}: Progressive training for spatial reasoning in vision-language models},
  author={Li, Hongxing and Li, Dingming and Wang, Zixuan and Yan, Yuchen and Wu, Hang and Zhang, Wenqi and Shen, Yongliang and Lu, Weiming and Xiao, Jun and Zhuang, Yueting},
  booktitle={International Conference on Learning Representations},
  year={2026}
}

@article{starkey2014emergence,
  title={The emergence of ``groupitizing'' in children's numerical cognition},
  author={Starkey, Gillian S and McCandliss, Bruce D},
  journal={Journal of Experimental Child Psychology},
  volume={126},
  pages={120--137},
  year={2014}
}

@inproceedings{liu2023libero,
  title={{LIBERO}: Benchmarking knowledge transfer for lifelong robot learning},
  author={Liu, Bo and Zhu, Yifeng and Gao, Chongkai and Feng, Yihao and Liu, Qiang and Zhu, Yuke and Stone, Peter},
  booktitle={Advances in Neural Information Processing Systems},
  year={2023}
}

@inproceedings{kim2024openvla,
  title={{OpenVLA}: An open-source vision-language-action model},
  author={Kim, Moo Jin and Pertsch, Karl and Karamcheti, Siddharth and Xiao, Ted and Balakrishna, Ashwin and Nair, Suraj and Rafailov, Rafael and Foster, Ethan P and Sanketi, Pannag R and Vuong, Quan and Kollar, Thomas and Burchfiel, Benjamin and Tedrake, Russ and Sadigh, Dorsa and Levine, Sergey and Liang, Percy and Finn, Chelsea},
  booktitle={Proceedings of The 8th Conference on Robot Learning},
  volume={270},
  pages={2679--2713},
  year={2025}
}

@inproceedings{zhao2025cotvla,
  title={{CoT-VLA}: Visual chain-of-thought reasoning for vision-language-action models},
  author={Zhao, Qingqing and Lu, Yao and Kim, Moo Jin and Fu, Zipeng and Zhang, Zhuoyang and Wu, Yecheng and Li, Zhaoshuo and Ma, Qianli and Han, Song and Finn, Chelsea and Handa, Ankur and Lin, Tsung-Yi and Wetzstein, Gordon and Liu, Ming-Yu and Xiang, Donglai},
  booktitle={Proceedings of the IEEE/CVF Conference on Computer Vision and Pattern Recognition},
  pages={1702--1713},
  year={2025}
}

@inproceedings{qu2025spatialvla,
  title={{SpatialVLA}: Exploring spatial representations for visual-language-action models},
  author={Qu, Delin and Song, Haoming and Chen, Qizhi and Yao, Yuanqi and Ye, Xinyi and Gu, Jiayuan and Wang, Zhigang and Ding, Yan and Zhao, Bin and Wang, Dong and Li, Xuelong},
  booktitle={Robotics: Science and Systems},
  year={2025}
}

@inproceedings{wang2026univla,
  title={Unified vision-language-action model},
  author={Wang, Yuqi and Li, Xinghang and Wang, Wenxuan and Zhang, Junbo and Li, Yingyan and Chen, Yuntao and Wang, Xinlong and Zhang, Zhaoxiang},
  booktitle={International Conference on Learning Representations},
  year={2026}
}

@misc{google2026gemma4,
  author = {{Google DeepMind}},
  title = {{Gemma 4}},
  year = {2026},
  note = {Open-weights model family, google/gemma-4-31B-it}
}

@article{bai2025qwen25vl,
  title={{Qwen2.5-VL} Technical Report},
  author={Bai, Shuai and Chen, Keqin and Liu, Xuejing and Wang, Jialin and Ge, Wenbin and Song, Sibo and Dang, Kai and Wang, Peng and Wang, Shijie and Tang, Jun and others},
  journal={arXiv preprint arXiv:2502.13923},
  year={2025}
}

@misc{qwen35,
  title={{Qwen3.5}: Towards Native Multimodal Agents},
  author={{Qwen Team}},
  month={February},
  year={2026},
  url={https://qwen.ai/blog?id=qwen3.5}
}

@article{lambert2024tulu,
  title={Tulu 3: Pushing frontiers in open language model post-training},
  author={Lambert, Nathan and Morrison, Jacob and Pyatkin, Valentina and Huang, Shengyi and Ivison, Hamish and Brahman, Faeze and Miranda, Lester James V and Liu, Alisa and Dziri, Nouha and Lyu, Shane and others},
  journal={arXiv preprint arXiv:2411.15124},
  year={2024}
}

@article{guo2025deepseek,
  title={{DeepSeek-R1}: Incentivizing reasoning capability in {LLMs} via reinforcement learning},
  author={Guo, Daya and Yang, Dejian and Zhang, Haowei and Song, Junxiao and Wang, Peiyi and Zhu, Qihao and Xu, Runxin and Zhang, Ruoyu and Ma, Shirong and Bi, Xiao and others},
  journal={arXiv preprint arXiv:2501.12948},
  year={2025}
}
}


\newpage
\appendix

\section{Prompts, Definitions, and Task Specifications}
\label{app:prompts}

This appendix reports the exact text of all prompts used in the \projectname{} benchmark. Every text free-response, multiple-choice, and image-editing query is composed by concatenating a fixed critical-instruction header, the six recurring concept definitions, and a task-specific question string. We reproduce all three components verbatim below.

\subsection{Critical-Instruction Header}

The critical-instruction header sets the test framing and is appended at the start of every prompt. The text is modality-specific because the response demand differs (a single value or word for text, a single capital letter for MCQ, an edited image for image editing).

\begin{tcolorbox}[
    colback=blue!3,
    colframe=blue!40!black,
    title=\textbf{Text free-response and vision-language-action models},
    fonttitle=\bfseries,
    breakable
]
\small\ttfamily
CRITICAL INSTRUCTION: You are undergoing a precision capability test. The input image shows a 3D structure built from objects of the same type, positioned at the center of the image. The objects are primarily boxes or cubes, but may also include stackable objects such as cylindrical objects, cans, and mugs. Answer the following questions using the absolute minimum number of words possible. Output only the exact value, count, or object name. Do not include full sentences, explanations, pleasantries, or context.
\end{tcolorbox}

\begin{tcolorbox}[
    colback=blue!3,
    colframe=blue!40!black,
    title=\textbf{Multiple-choice with image options (single-view tasks)},
    fonttitle=\bfseries,
    breakable
]
\small\ttfamily
CRITICAL INSTRUCTION: You are undergoing a precision capability test. The input image shows a 3D structure built from objects of the same type, positioned at the center of the image. The objects are primarily boxes or cubes, but may also include stackable objects such as cylindrical objects, cans, and mugs. You will receive exactly 6 images in sequence:

- Image 1 is the REFERENCE structure.

- Images 2, 3, 4, 5, and 6 are the MULTIPLE-CHOICE OPTIONS corresponding to choices A, B, C, D, and E respectively. Each option depicts a possible coloring of the REFERENCE structure.

Select the option image that correctly answers the question by applying the described coloring to the structure. Output ONLY a single capital letter (A, B, C, D, or E) corresponding to the correct option. Do not include full sentences, explanations, pleasantries, or context.
\end{tcolorbox}

\begin{tcolorbox}[
    colback=blue!3,
    colframe=blue!40!black,
    title=\textbf{Multiple-choice with image options (mental rotation tasks)},
    fonttitle=\bfseries,
    breakable
]
\small\ttfamily
CRITICAL INSTRUCTION: You are undergoing a precision capability test. The images show a 3D structure built from objects of the same type, positioned at the center of the image. The objects are primarily boxes or cubes, but may also include stackable objects such as cylindrical objects, cans, and mugs. You will receive exactly 7 images in sequence:

- Image 1 is VIEW A of an object structure.

- Image 2 is VIEW B of the exact same object structure.

- Images 3, 4, 5, 6, and 7 are the MULTIPLE-CHOICE OPTIONS corresponding to choices A, B, C, D, and E respectively. Each option depicts a possible coloring of VIEW B.

Select the option image that correctly answers the question by applying the described coloring to VIEW B. Output ONLY a single capital letter (A, B, C, D, or E) corresponding to the correct option. Do not include full sentences, explanations, pleasantries, or context.
\end{tcolorbox}

The mental rotation header refers to VIEW B, but the options color VIEW A, as the question states. Models received the prompt as shown.

\begin{tcolorbox}[
    colback=blue!3,
    colframe=blue!40!black,
    title=\textbf{Image editing},
    fonttitle=\bfseries,
    breakable
]
\small\ttfamily
CRITICAL INSTRUCTION: You are undergoing a precision capability test. The input image shows a 3D structure built from objects of the same type, positioned at the center of the image. The objects are primarily boxes or cubes, but may also include stackable objects such as cylindrical objects, cans, and mugs. Only coloring of the objects in the original image is allowed. Do NOT change, add, or move the objects and thus the structure in the image. Do NOT remove or color any part of the background of the original image.
\end{tcolorbox}

\subsection{Concept Definitions}

The following six recurring definitions are appended after the critical-instruction header on every task that depends on them (i.e., every sub-task and target task except the open-ended object-categorization task that asks only for the object type).

\begin{tcolorbox}[
    colback=green!3,
    colframe=green!40!black,
    title=\textbf{Definitions},
    fonttitle=\bfseries,
    breakable
]
\small
\begin{enumerate}[leftmargin=*, itemsep=4pt]
    \item A \textbf{COLUMN} is a vertical stack of one or more objects in which each object lies either on the ground or solely on top of the object immediately beneath it.
    \item A \textbf{LAYER} is a horizontal group of one or more objects at the same vertical height in the structure. The objects within the same layer could be laterally detached or attached. A single object also counts as a layer.
    \item A \textbf{VISIBLE OBJECT} is an object with at least one face fully or partially visible in the image.
    \item A \textbf{CLUSTER} is a group of one or more objects that are connected through direct contact, possibly across multiple layers. Two objects are considered connected if they are in direct physical contact (i.e., directly adjacent horizontally or vertically with no gap between them); diagonal placement that results in no contact does not count. This applies to both box-shaped and cylindrical objects. A cluster includes all objects that are connected directly or indirectly through such contacts.
    \item A \textbf{SUPPORTING OBJECT} is any object that is beneath a given object in the same column. A \textbf{DIRECTLY SUPPORTING OBJECT} is the supporting object in immediate contact beneath a given object. If an object is directly supported by the ground, then it has no supporting objects.
    \item A \textbf{HIDDEN OBJECT} is an object with no visible faces in the image. Any hidden object must be a supporting object of at least one visible object. Otherwise, such an object does not exist in a valid structure.
\end{enumerate}
\end{tcolorbox}

\subsection{Task-Specific Queries}

The body of each prompt is the task-specific query, appended after the critical-instruction header and (where relevant) the concept definitions. Table~\ref{tab:task_queries} reports the question text for all nine sub-tasks, the two target tasks, and the two mental rotation variants, in each of the three response formats.

\begin{table}[ht]
\centering
\small
\caption{Task-specific query text for the nine sub-tasks (S1--S9), the \textit{Object Counting} target task (T1), and the two \textit{Mental Rotation} variants (T2). Text columns are slightly compressed for typography, and the full strings are in the released code.}
\label{tab:task_queries}
\begin{adjustbox}{max width=\textwidth}
\begin{tabular}{l p{0.55\textwidth} p{0.30\textwidth}}
\toprule
\textbf{Task} & \textbf{Text free-response} & \textbf{Image-editing instruction} \\
\midrule
S1: Object Categorization & What type of object(s) make up the structure at the center of the image? Give the object type name only. & Color the structure with a unique color. \\
\midrule
S2: Cluster Count & How many distinct clusters are there in the object structure? & Color each distinct cluster with a unique color. \\
\midrule
S3: Column Count & How many columns are in the object structure? & Color all objects within each column with a new unique color. \\
\midrule
S4: Layer Count & How many layers are in the object structure? & Color all objects within each layer with a new unique color. \\
\midrule
S5: Visible Object Count & How many VISIBLE objects are there in the object structure? & Color each object with a new unique color. \\
\midrule
S6: Top Layer & How many objects are in the top-most layer of the whole structure? & Color the top-most object(s) with a unique color\\
\midrule
S7: Direct Support & How many VISIBLE OBJECTS are directly supporting the top-most object(s)? Answer 0 if all columns have height 1. & Color the object(s) directly supporting the top-most object(s). \\
\midrule
S8: Support Column & How many VISIBLE OBJECTS are in the column(s) that contain the top-most object(s), excluding the top-most object(s) themselves? & Color all objects in the column(s) that contain the top-most object(s), excluding the top-most object(s) themselves. \\
\midrule
S9: Hidden Object Count & For each inferred HIDDEN OBJECT, identify the single VISIBLE OBJECT in immediate contact directly above it. How many such visible objects are there? & Color each VISIBLE OBJECT in immediate contact directly above a HIDDEN OBJECT. \\
\midrule
T1: Object Counting & How many TOTAL objects are present in the object structure? & Generate an image with the integer answer in black on a white background. \\
\midrule
T2 ($90^\circ$): Mental Rotation & Given two views of the same structure, how many objects are visible in the second view but not the first? & Color all objects visible in the second view but not the first. \\
\midrule
T2 ($180^\circ$): Mental Rotation & Given two views of the same structure, how many objects are visible in the second view but not the first? & Color all objects visible in the second view but not the first. \\
\bottomrule
\end{tabular}
\end{adjustbox}
\end{table}

The MCQ format reuses the same task content as the text free-response format, rephrased as ``Which option (A--E) correctly depicts...'' against the image options described in the modality framing. We omit the rephrased MCQ strings here for space. They can be derived mechanically from the text free-response queries by substituting the option-selection wording, except for S1, which asks for a single-color rendering, and T2, which asks for objects visible in the first view but not the second.

\subsection{Multiple-Choice Wrong-Answer Taxonomy}
\label{app:mcq_distractors}

\begin{figure}[t]
\centering
\begin{minipage}{\textwidth}
\centering
\includegraphics[width=\textwidth]{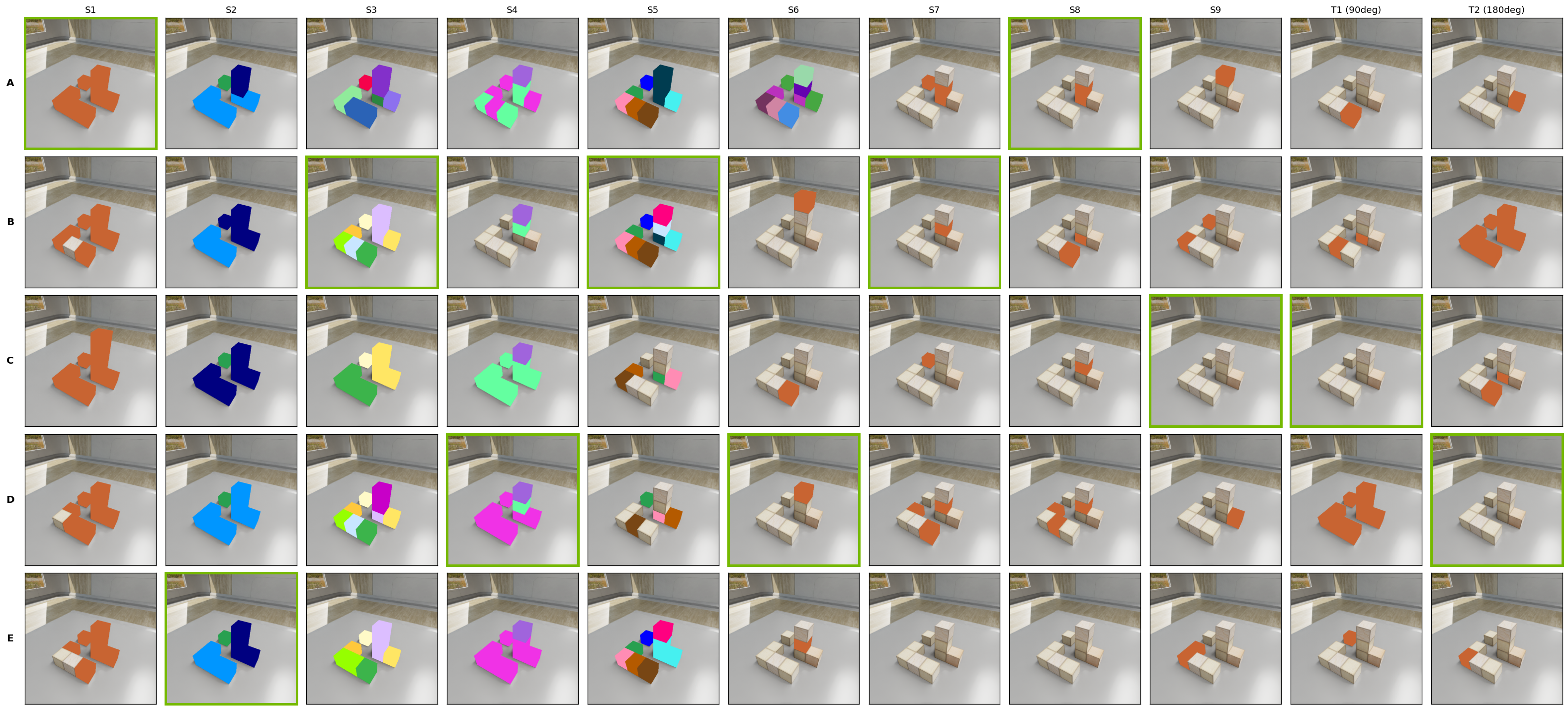}
\\[0.5em]
{\small (a) All multiple-choice answers for all tasks with image options (Example 1).}
\end{minipage}
\\[1em]
\begin{minipage}{\textwidth}
\centering
\includegraphics[width=\textwidth]{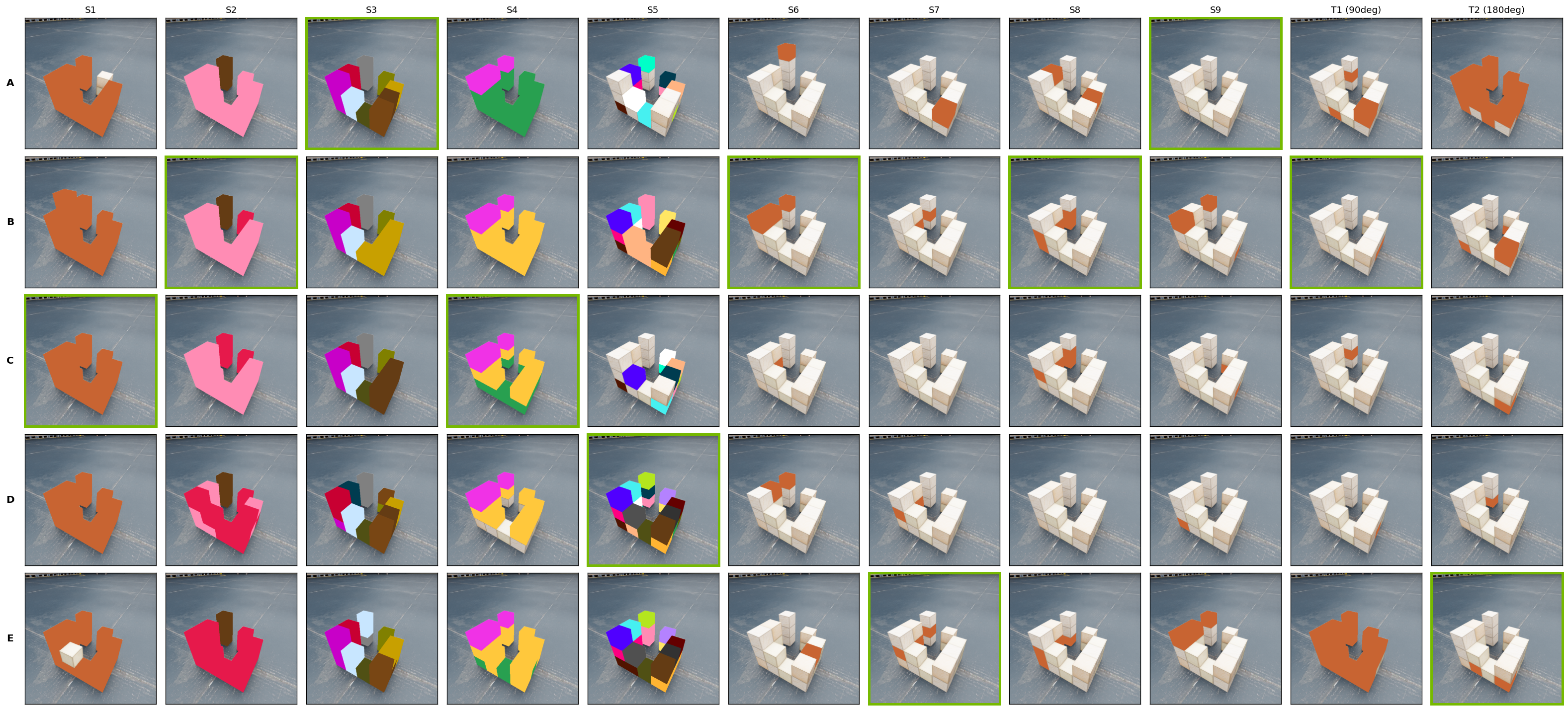}
\\[0.5em]
{\small (b) All multiple-choice answers for all tasks with image options (Example 2).}
\end{minipage}
\caption{Examples of the five-choice MCQ option grids for two procedurally generated scenes. Columns correspond to the eleven sub-task and target-task panels (S1--S9, T2 at $90^\circ$, T2 at $180^\circ$); rows label the five answer choices A--E. The correct option for each task is outlined in green. The four wrong options per cell are constructed according to the per-task taxonomy in Appendix~\ref{app:mcq_distractors}.}
\label{fig:mcq_examples}
\end{figure}

For every sub-task and target task, we generate four distinct wrong-answer images per scene rather than perturbing a single answer integer. Distractors are constructed deterministically from the scene's voxel ground truth so that most wrong choices are visually well-formed renderings of a specific named counting confusion, and the rest are the random fallbacks described below. Distractor types are designed task-by-task, and we name each category after the underlying confusion it represents (e.g., merge-two-columns, phantom-block-above-top). Most categories are specific to the task they diagnose, although phantom-block-above-top is shared by S1 and S6. Color labels and palette assignments are permutation-invariant under our scoring (a coloring that swaps two color slots is judged equivalent to the ground truth), so distractor types that differ only by such swaps are not used. Below we report the four wrong options per task, drawn from two to four error types, and the procedural rules used to construct them. Figure~\ref{fig:mcq_examples} shows two example scenes with their full grids of five rendered MCQ options across all eleven tasks.

\paragraph{\textit{Object Categorization} (S1).}
(1) \textbf{phantom-block-above-top}: an extra cube is spawned on top of a randomly chosen non-empty column and highlighted along with the rest of the visible structure. (2) \textbf{omit-one-block}: highlights all visible blocks except one. (3) \textbf{omit-two-blocks}: highlights all visible blocks except two. (4) \textbf{drop-one-column-highlight}: highlights all visible blocks except those in one randomly chosen column.

\paragraph{\textit{Cluster Count} (S2).}
The four distractors depend on the number of true clusters $C$ in the scene. When $C{=}1$: \textbf{multicolor-single-cluster}, \textbf{single-cluster-one-color-a}, \textbf{single-cluster-one-color-b}, and \textbf{fake-split-into-two-colors}. When $C{=}2$: \textbf{fake-split-one-cluster}, two directions of \textbf{merge-two-clusters} (cluster $0 \to$ cluster $1$'s color, and the reverse), and \textbf{collapse-to-one-cluster-color}. When $C \geq 3$: \textbf{fake-split-largest-cluster}, and three directions of \textbf{merge-two-clusters} between the three largest clusters. The one-color constructions (single-cluster-one-color-a and -b, and collapse-to-one-cluster-color) duplicate another option, so they always fall back to a random coloring of the visible blocks.

\paragraph{\textit{Column Count} (S3).}
(1)--(2) \textbf{same-column-two-colors}: two or three tall columns are split such that blocks within the same column receive different colors. (3)--(4) \textbf{merge-two-columns}: one or two disjoint groups of two to four spatially adjacent columns are assigned the same color. The two instances of each type use different random columns or merge groups, and permuting color labels alone is not used because column-coloring ground truth is permutation-invariant.

\paragraph{\textit{Layer Count} (S4).}
(1) \textbf{same-layer-two-colors}: two blocks in the same layer receive different colors. (2) \textbf{merge-two-layers}: the bottom two layers are assigned the same color. (3) \textbf{missing-one-layer}: one layer is left uncolored. (4) \textbf{far-layer-confusion}: the same two bottom layers are assigned one color, taken from the other layer of the pair.

\paragraph{\textit{Visible Object Count} (S5).}
(1)--(2) \textbf{merge-two-adjacent-same-color} / \textbf{merge-three-adjacent-same-color}: spatially adjacent pairs or triplets of visible blocks are assigned the same color, with the two instances using different random pairings. (3) \textbf{drop-half-visible-blocks}: roughly half of the visible blocks (a random subset of size $\max(1, \lfloor |V|/2 \rfloor)$) are colored and the rest are left uncolored, with the two instances using different random subsets.

\paragraph{\textit{Top Layer} (S6).}
(1) \textbf{phantom-block-above-top}: an extra cube is placed directly above an existing top-layer column and highlighted as if it were the new top. (2) \textbf{missing-one-top-block}: highlights all top-layer blocks except one. (3) \textbf{one-below-top}: highlights the block immediately below a top-layer block instead of the top-layer block itself. (4) \textbf{other-column-top}: highlights the top-most block of a non-top-layer column.

\paragraph{\textit{Direct Support} (S7).}
(1) \textbf{miss-one-support}: highlights all directly supporting blocks except one. (2) \textbf{wrong-support-subset}: highlights an alternative incorrect subset of the support set. (3) \textbf{empty-support}: highlights no blocks (claiming no blocks support the top-most). (4) \textbf{non-support-highlight}: highlights a single non-supporting block.

\paragraph{\textit{Support Column} (S8).}
(1) \textbf{remove-support-column-blocks}: highlights all support-column blocks except one. (2) \textbf{wrong-column-empty}: highlights nothing. (3) \textbf{remove-two-column-blocks}: highlights all support-column blocks except two. (4) \textbf{support-blocks-not-column}: intended to highlight the directly supporting blocks (the \textit{Direct Support} set) rather than the full \textit{Support Column} set. The \textit{Direct Support} set lies inside the \textit{Support Column} set, so this option duplicates another option and always falls back to a random highlight.

\paragraph{\textit{Hidden Object Count} (S9).}
(1) \textbf{miss-one-supported}: highlights all visible blocks above hidden objects except one. (2) \textbf{empty-set}: highlights no blocks. (3) \textbf{wrong-visible-block}: highlights a randomly chosen visible block that is not above any hidden object. (4) \textbf{alt-incomplete-subset}: intended as an alternative incomplete subset of the same set. It duplicates miss-one-supported, so it always falls back to a random highlight.

\paragraph{\textit{Object Counting} target task (T1).}
The target-task answer is a positive integer, so the five MCQ choices are integers rather than recolored images. We sample a pivot integer $P = N + d$ where $N$ is the correct count and $d$ is drawn uniformly from $\{-3, -2, -1, 0, 0, 1, 2, 3\}$ (zero weighted twice), biasing the pivot toward but not pinning it to the correct answer. We then sample four distinct positive spreads from $\{1, 2, 3, 4, 5\}$ and form a candidate pool of $\{P - s_i,\, P + s_i\}$. The correct answer $N$ is always included; the five final options are drawn from the pool, with any leftover slots filled by random integers in $[N - 8,\, N + 8]$. All options are positive and distinct, and option order is randomized per scene. This construction weakens two heuristics that a simpler $\{N - 2, N - 1, N, N + 1, N + 2\}$ design would invite. It reduces but does not eliminate the chance that the correct answer sits at the median of the displayed options ($28.7\%$ of views, against $20\%$ expected), and it asks the model to commit to a specific count rather than to a small window of values around the median.

\paragraph{\textit{Mental Rotation} (T2, $90^\circ$ and $180^\circ$).}
(1) \textbf{random-wrong-highlight-blocks}: highlights a random subset of one to three visible blocks that are not in the cross-view set. (2) \textbf{random-single-block}: highlights one random visible block from the first view. (3) \textbf{both-views-blocks}: highlights the blocks visible in both views (rather than only the first). (4) \textbf{missing-nonmatch}: highlights all but one of the cross-view-only blocks. All four are drawn on the first view.

\paragraph{Deduplication and fallback.}
Distractor renderings can collide with each other or with the ground-truth coloring when the underlying scene admits a degenerate construction (e.g., a scene with too few tall columns for two distinct same-column-two-colors instances). After the four distractor images for a task are rendered, we compute an RGBA fingerprint per image and check for duplicates among the four distractors and against the ground-truth coloring. If a duplicate is detected, we replace it with a procedurally generated fallback distractor: a random highlight subset of one to three blocks for support-style sub-tasks, or a random color partition of the visible blocks for clustering-style sub-tasks. The replacement is retried up to a fixed budget. If no unique distractor is found within the budget, the code inserts a black dummy image, a case that does not occur in the data. Overall, $22.4\%$ of image distractors are fallbacks, and $55.4\%$ of task-views contain at least one.

\paragraph{Per-scene reproducibility.}
Each task's distractor RNG is seeded by a deterministic hash of the scene identifier and task name, so re-rendering the same scene produces the same five choice images and the same correct-answer letter. Option letter order is randomized within this seed, so the correct answer is not systematically located at a single position.

\subsection{Prompt validation on easy cases.}

Before running large-scale inference, we validate the full pipeline on a small set of representative easy structures (a single object, two stacked objects, and a $2 \times 2 \times 2$ cube). A correctly functioning model is expected to score at ceiling on all sub-tasks for these structures, and a deviation from ceiling is treated as a prompt defect rather than a genuine model failure. The frontier text-output models all clear the easy-case threshold, which we take as evidence that the prompts are unambiguous and sufficiently instructive. We use this same check at every stage of the study, including when adding new sub-tasks or revising prompts.

\section{Human Study Protocol}
\label{app:human_study}

\paragraph{Recruitment and compensation.}
All human responses in this paper were collected from NVIDIA's in-house annotation team. Annotators are NVIDIA employees compensated by salary through standard internal channels; no external recruitment platform was used.

\begin{figure}[t]
\centering
\includegraphics[width=0.9\textwidth]{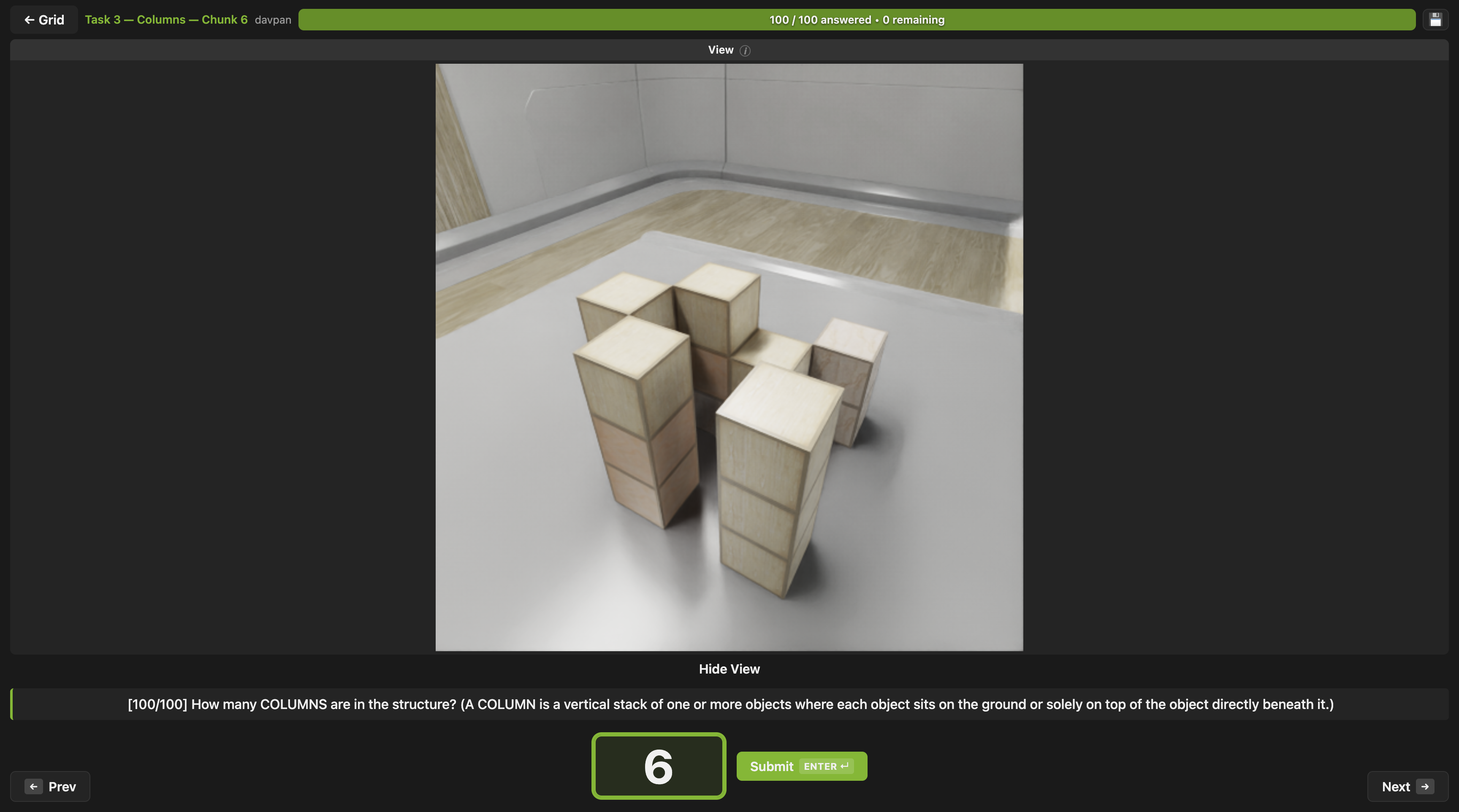}
\caption{Human-study interface for the text free-response modality. The annotator is shown the same reference image and the same task-specific question that the language models receive, and submits an integer or short label. This example shows a \textit{Column Count} (S3) trial, with the task-specific definition of "column" included verbatim in the prompt. Progress and chunk metadata appear in the top header.}
\label{fig:human_text_interface}
\end{figure}

\begin{figure}[t]
\centering
\includegraphics[width=0.9\textwidth]{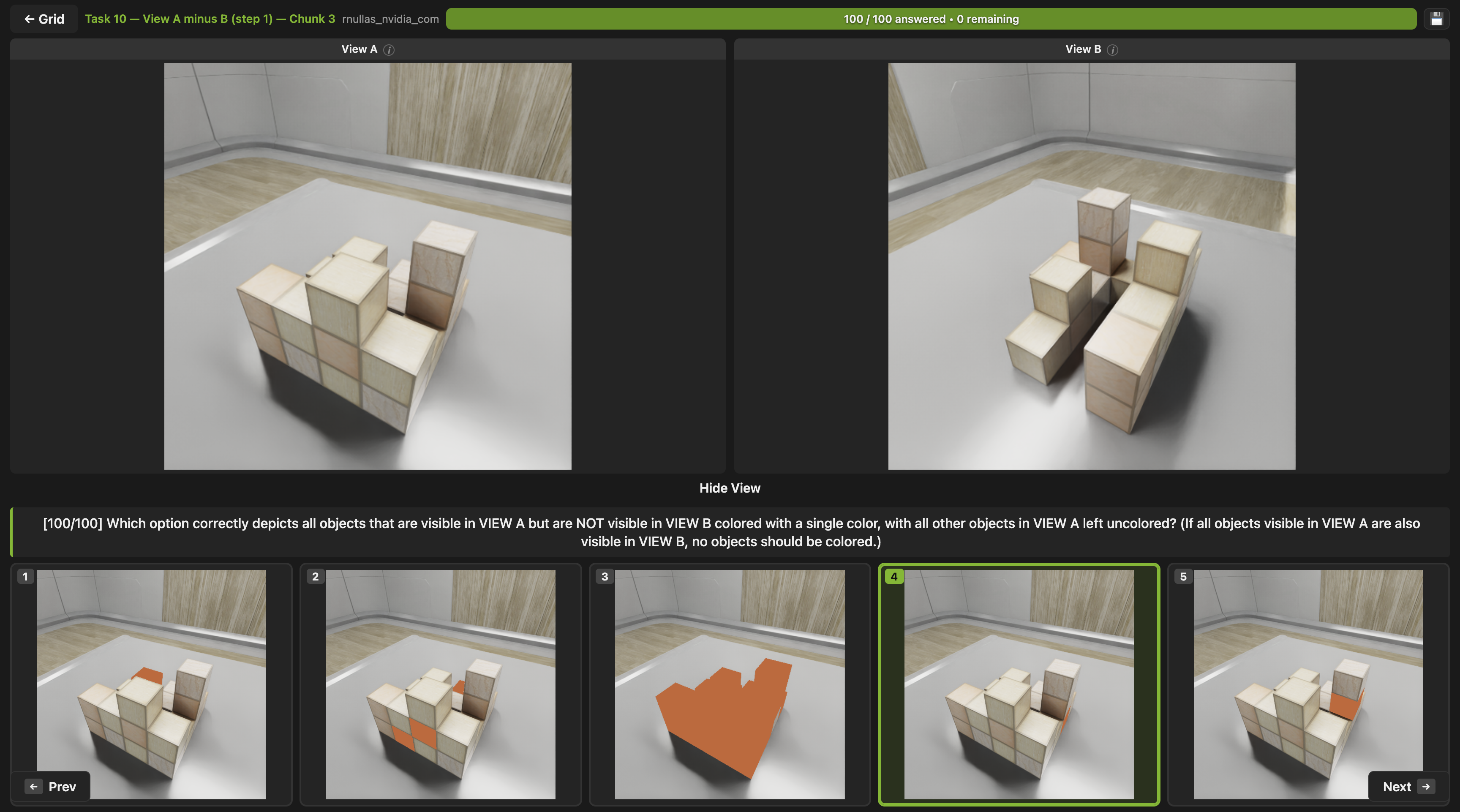}
\caption{Human-study interface for the multiple-choice modality, illustrated on a five-choice \textit{Mental Rotation} (T2) trial. The annotator is shown VIEW A and VIEW B of the same structure (top), the task-specific question (middle), and five candidate option images (bottom) corresponding to choices A--E. The annotator clicks the option image that correctly answers the question; the selected option is outlined in green.}
\label{fig:human_mcq_interface}
\end{figure}

\begin{figure}[t]
\centering
\begin{minipage}[t]{0.49\textwidth}
\centering
\includegraphics[width=\textwidth]{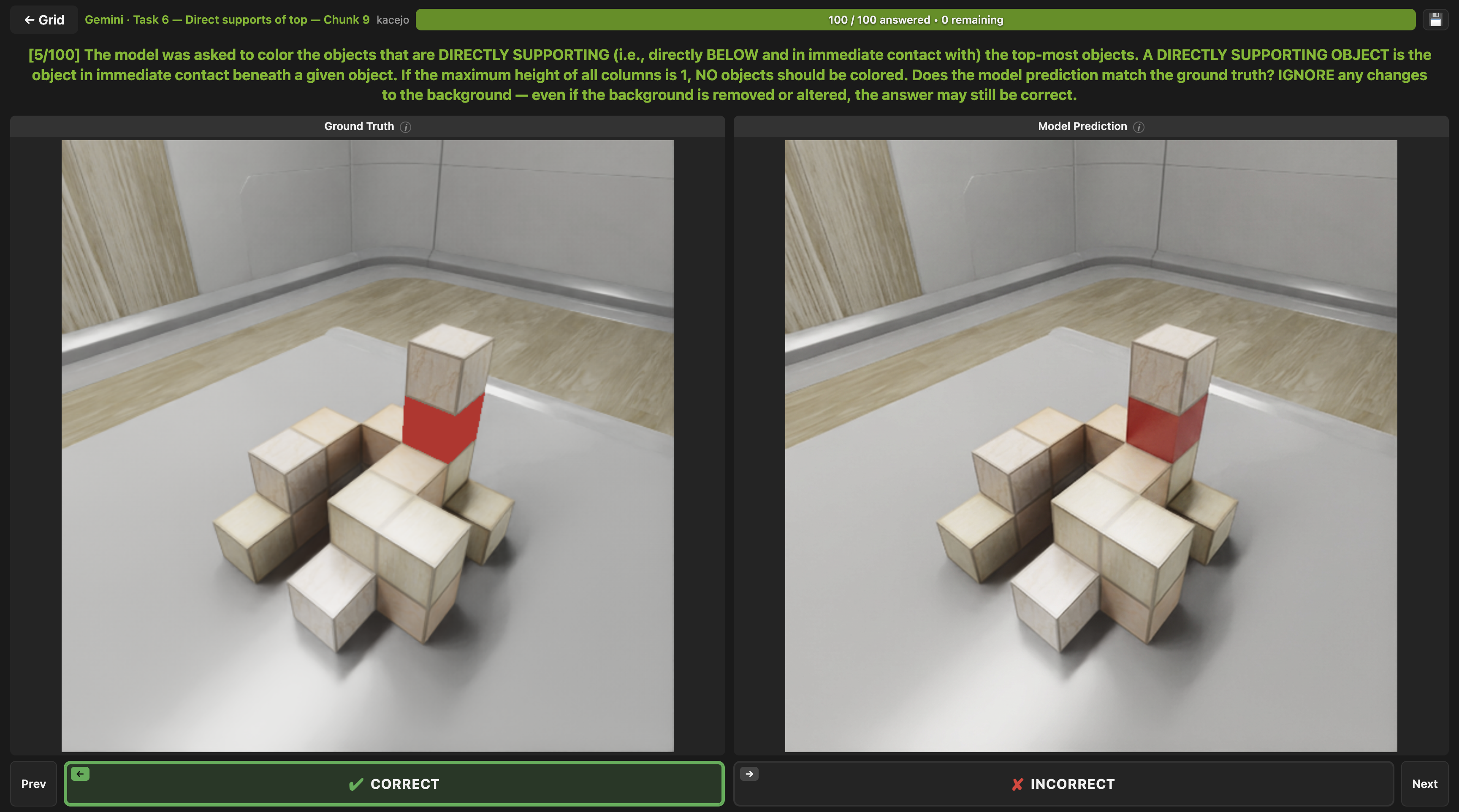}
\caption*{(a) Annotator correctly scores this CORRECT.}
\end{minipage}
\hfill
\begin{minipage}[t]{0.49\textwidth}
\centering
\includegraphics[width=\textwidth]{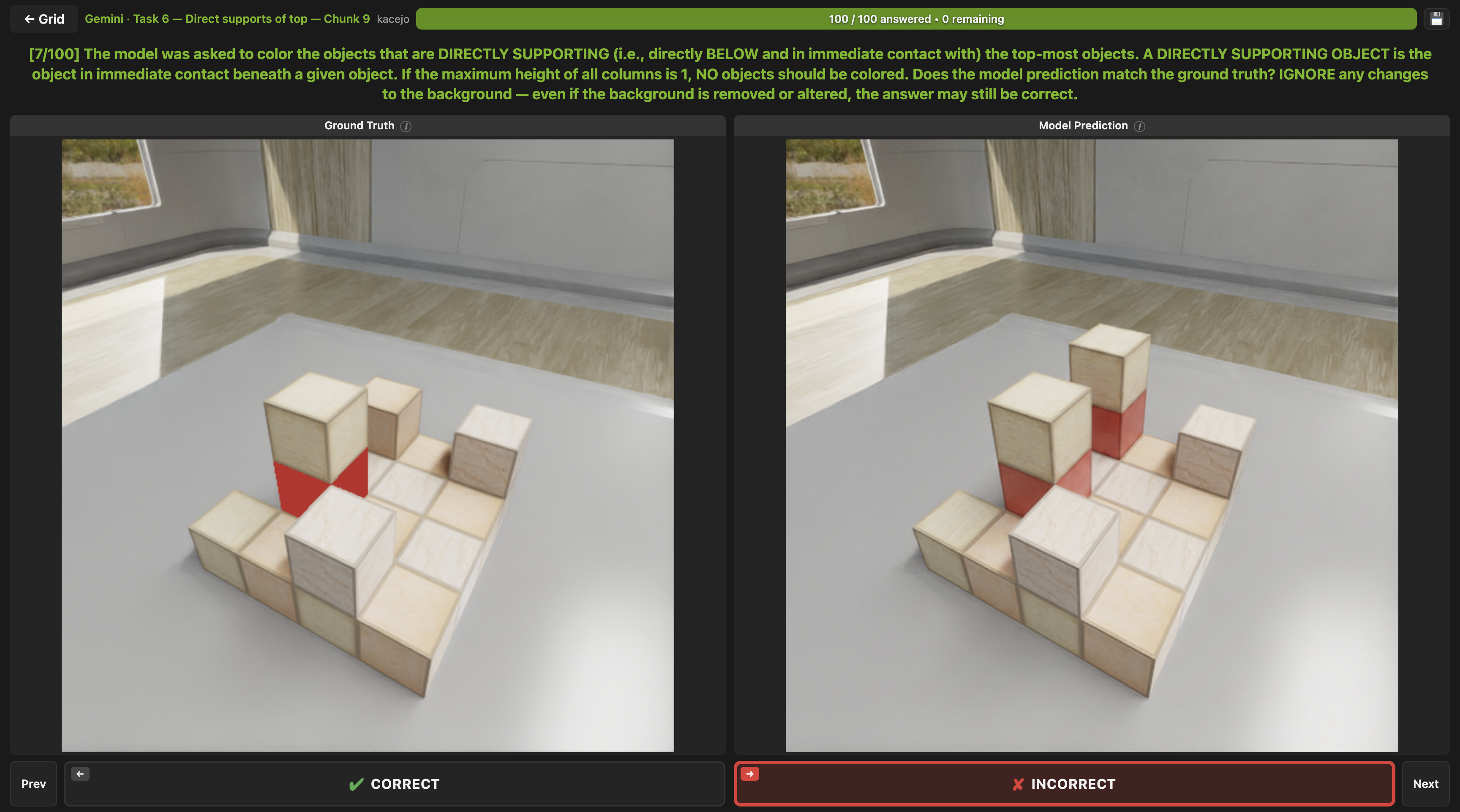}
\caption*{(b) Annotator correctly scores this INCORRECT.}
\end{minipage}
\caption{Human-scoring interface for the image-editing modality, shown on \textit{Direct Support} (S7) trials produced by Gemini Flash Image. The annotator views the ground-truth coloring (left) and the model's predicted coloring (right) side by side, and judges whether the prediction matches the ground truth. The instruction explicitly tells annotators to ignore changes to the background, since image-editing models occasionally alter or remove background pixels even when their object-level coloring is correct. Panel (a) shows a trial scored as Correct; panel (b) shows a trial scored as Incorrect.}
\label{fig:human_image_interface}
\end{figure}

\paragraph{Study structure.}
Each annotator is assigned a chunk of 100 questions at a time, with all 100 questions belonging to the same (sub-)task. This blocked design lets the annotator internalize the task definition once per chunk rather than context-switching between tasks. Each chunk begins with three warm-up questions for which the ground-truth answer is revealed immediately after submission, so the annotator can calibrate on the task's specific instructions before the 100 graded responses begin. Warm-up responses are not included in the human baseline; only the 100 graded responses per chunk are used in the analysis. Three different annotator interfaces are used, one for each of the three response formats we evaluate (Figures~\ref{fig:human_text_interface}--\ref{fig:human_image_interface}).

\paragraph{Text free-response interface.}
Figure~\ref{fig:human_text_interface} shows the interface used to collect human responses for the text modality. The annotator views the reference image and reads the same task-specific prompt that text-output models receive, including the verbatim concept definitions where applicable. They submit a single integer or short label via a numeric keypad, with Enter advancing to the next trial. This interface is used for eight sub-tasks (S2--S9) plus the \textit{Object Counting} target task, and mental rotation trials are presented in the same interface with two reference views displayed side by side rather than one.

\paragraph{Multiple-choice interface.}
Figure~\ref{fig:human_mcq_interface} shows the MCQ interface, illustrated on a five-choice \textit{Mental Rotation} trial. The annotator is shown the reference image (or the two views, for mental rotation), the same task-specific question that MCQ-condition models receive, and the same five candidate option images (A--E). They select the correct option by clicking. Only the five-choice condition is collected from human annotators; the four- and three-choice numbers in the main paper come from the model evaluations.

\paragraph{Image-editing scoring interface.}
The image-editing modality requires a different protocol because the model output is a full image rather than an integer or a discrete selection. Figure~\ref{fig:human_image_interface} shows the interface used to score image-editing model outputs. The annotator views the ground-truth coloring and the model's predicted coloring side by side, and judges whether the prediction matches the ground truth at the level of which objects were colored. The instruction explicitly tells annotators to ignore changes to the background, because image-editing models occasionally alter or remove background pixels even when their object-level coloring is correct, and we want to score the spatial-reasoning content of the response rather than its incidental edit fidelity. This protocol is used to score Gemini Flash Image, Qwen-Image-Edit, and HunyuanImage-Instruct.

\paragraph{Easy-case warm-up.}
Before the main benchmark trials, annotators encounter the same easy-case validation set used to validate the model prompts (Section~\ref{app:prompts}, "Prompt validation on easy cases"): a single object, two stacked objects, and a $2 \times 2 \times 2$ cube. An annotator who fails any of these is given the task definition and concept definitions a second time before the chunk continues. In practice, all annotators cleared the easy cases on the first attempt, which we take as evidence that the task definitions are unambiguous from the human side as well as from the model side.

\section{Hierarchy Details}
\label{app:hierarchy_details}

This appendix expands the four-layer organization of the sub-task hierarchy summarized in Section~\ref{sec:dataset_generation}, which Figure~\ref{fig:task_overview} also conveys schematically.

\subsection{Developmental Frameworks}
\label{app:development_stages}

\begin{figure}[t]
\centering
\includegraphics[width=\textwidth]{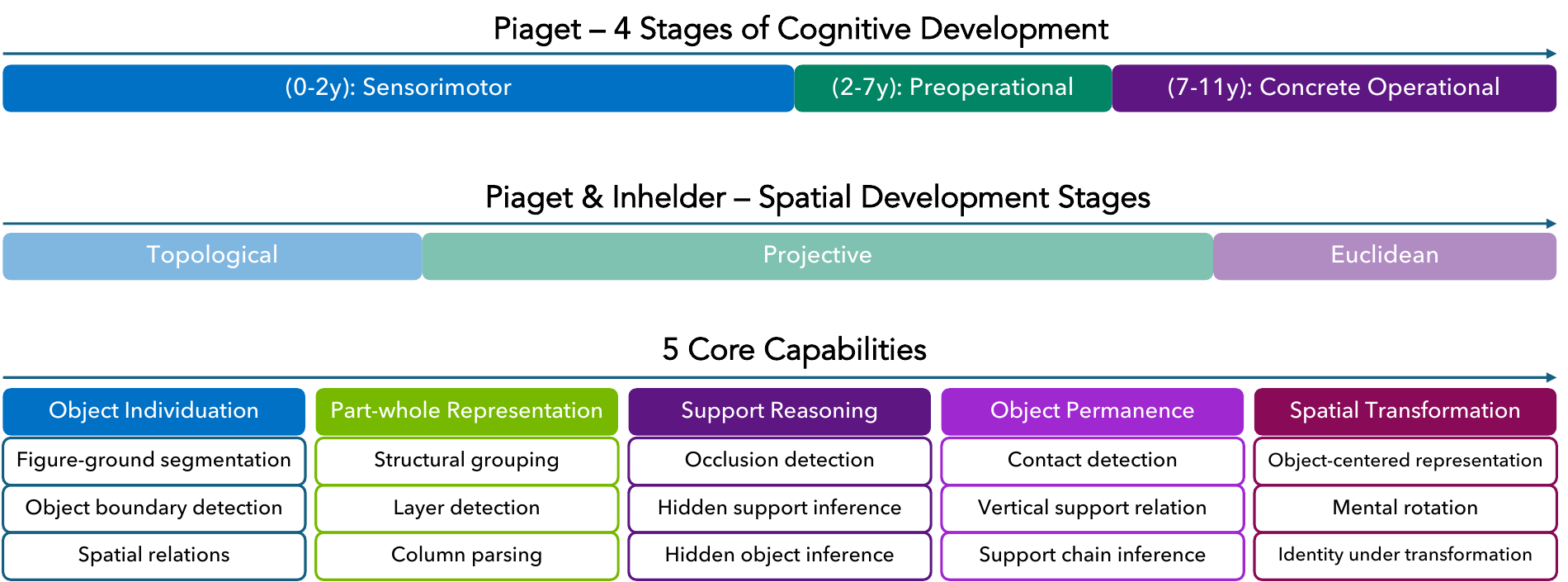}
\caption{The three developmental frameworks tagged on each task panel of Figure~\ref{fig:task_overview}: the three of Piaget's four stages of cognitive development relevant here, Piaget and Inhelder's spatial development stages, and our five core spatial capabilities and their constituent primitives.}
\label{fig:development_stages}
\end{figure}

Each task panel in Figure~\ref{fig:task_overview} is tagged with its position on three developmental frameworks, laid out together in Figure~\ref{fig:development_stages}. The first two are drawn directly from the developmental psychology literature: Piaget's four stages of cognitive development~\citep{piaget1969psychology}, and Piaget and Inhelder's three spatial development stages of topological, projective, and Euclidean reasoning~\citep{piaget1956child}. The third is a framework of our own that groups spatial competence into five core capabilities (object individuation, part-whole representation, object permanence, support reasoning, and spatial transformation), each realized by a small set of perceptual or cognitive primitives that the corresponding sub-tasks are designed to probe.

\subsection{Four-Layer Sub-Task Organization}

\paragraph{Primitive perceptual organization.}
To establish the visible structure of the scene, \textit{Object Categorization} requires identifying the foreground objects that compose the stacked structure. \textit{Column Count}, \textit{Layer Count}, and \textit{Cluster Count} require grouping those visible objects into spatial units defined by alignment, height, and local connectedness. These sub-tasks are intended to provide the structural primitives from which higher-level reasoning is built and thus indirectly support \textit{Visible Object Count}.

\paragraph{Local structural reasoning.}
To support the hidden-object branch of the target task, we test two related competencies: referential chaining and gravity-governed support understanding. \textit{Top Layer} requires identifying the objects in the highest occupied layer, establishing the initial structural referent. \textit{Direct Support} then requires identifying the objects immediately supporting those top objects, testing whether the model can follow that referent downward through a physically meaningful support relation. \textit{Support Column} extends this reasoning by localizing the relevant support structure within each supporting column. These sub-tasks therefore do more than probe generic local structure: they test whether the model can preserve a reference (\textit{Top Layer}) across steps while also respecting the directional logic of stacked objects under gravity. Thus, they form the \textit{Internal Referential Chain} rather than a loose collection of related sub-tasks.

\paragraph{Occlusion-sensitive counting.}
\textit{Visible Object Count} requires exact enumeration of the visible foreground objects and is grounded primarily in the perceptual-organization branch. \textit{Hidden Object Count} requires inferring the existence of occluded objects that must be present to support the visible arrangement and is grounded primarily in the support-reasoning branch. Whereas \textit{Visible Object Count} depends on correctly individuating and organizing the foreground structure, \textit{Hidden Object Count} depends on reconstructing latent 3D structure from support, adjacency, and occlusion constraints. The benchmark is designed so that these two branches can be evaluated separately before they are recombined.

\paragraph{Target-level integration.}
The primary target task, \textit{Object Counting}, requires reporting the total number of objects in the stacked structure. By benchmark construction, the total equals the visible objects plus the hidden objects, and \textit{Hidden Object Count} probes the hidden term through the visible objects resting directly on a hidden object. The hidden term and this probe coincide unless hidden objects stack on one another ($17.9\%$ of benchmark views). We treat the visible-plus-hidden relation as the \textit{Summation Mechanism}: it defines how the final answer can be assembled once its two main components are available, while leaving open whether the model genuinely used the intended upstream competencies to produce those component counts.

\section{Dataset Generation Details}
\label{app:dataset_details}

This appendix expands the procedural generation summary in Section~\ref{sec:dataset_generation}.

\paragraph{Structure sampling distribution.}
We sample a target object count $n$ from a piecewise probability density that rises linearly from a small weight at $n{=}3$ to a peak at $n{=}11$, halves at $n{=}12$, declines linearly through $n{=}30$, and then decays exponentially. We then draw $n$ random grid locations on the $4 \times 4 \times 4$ voxel grid and apply gravity-based collapse: each column is independently consolidated into a contiguous stack of objects supported by the ground plane, so any voxel positions that would have been suspended in midair are pulled downward to rest on the column below them. The result is a structure of exactly $n$ objects in which every column is ground-supported. Each generation run uses one object type, and the evaluation set is $55\%$ textured cubes, $25\%$ factory-style boxes, $10\%$ tomato soup cans, and $10\%$ potted meat cans. Within a single scene, all objects are of one type so that sub-task answers do not depend on type-level disambiguation.

\paragraph{Difficulty calibration.}
The object-count weight function described above was tuned during a piloting phase rather than chosen a priori. Naive uniform sampling over the $4 \times 4 \times 4$ grid concentrates probability mass at very sparse and very dense scenes, neither of which discriminates between models: sparse scenes are at ceiling for every frontier model we tested, while dense scenes are at floor and produce noisy correlations with the sub-tasks. We piloted on the frontier text-output models and observed that scenes around eleven objects were the regime in which model accuracy began to separate, which set the peak of our piecewise weight. The slow falloff through thirty objects and the exponential decay beyond was chosen to retain enough hard scenes to measure tail behavior without dominating the eval set with structures that no current model solves.

\paragraph{Camera and environment variation.}
We render every scene from four base views spaced $90^\circ$ apart, at two azimuth offsets ($3^\circ$ and $12.5^\circ$) and under two perspective conditions (a default lens and a wide-angle dolly zoom), giving four camera variations per base view. Backgrounds consist of four textured indoor and warehouse environments with minimal foreground clutter, and lighting is path-traced via NVIDIA RTX Interactive Path Tracing. The two camera factors are designed to isolate two known sources of perceptual difficulty. The first is camera azimuth: the base camera sits about $42.6^\circ$ off the grid axis, and the offset changes how much of each object's side faces is exposed (especially for cube and factory-box objects). The second source of difficulty concerns the dolly-zoom effect: the size of the structure in the image can be approximately preserved while strongly varying perspective distortion. We render each scene under the default lens and again after widening the field of view threefold (shorter focal length) and moving the camera to $0.3\times$ the distance, chosen so that the projected structure size is approximately preserved between conditions. The cross-product of azimuth offset and dolly setting gives the four camera variations per base view. As reported in Appendix~\ref{app:difficulty_modalities}, both factors produce near-zero point-biserial correlations with \textit{Object Counting} correctness across models and the human baseline, confirming that they function as robustness axes rather than primary difficulty axes.

\paragraph{Ambiguity removal: the depth-buffer occlusion test.}
Whether a hidden object exists is undecidable from a single image alone: an object placed behind a visible object at a depth where it could not have been inferred by support is observationally equivalent to its absence. To identify hidden objects unambiguously, we apply a pixel-level depth-buffer occlusion test using the Replicator toolkit~\cite{nvidia_isaac_replicator_420}. For each candidate object in the structure, we query the per-pixel depth buffer of the rendered camera view and test whether any pixel belonging to that object is the nearest occupied surface along its ray. An object is retained if and only if at least one of its pixels is the nearest surface along some camera ray (i.e., it is camera-visible) or it lies beneath a retained object in the same column and is therefore necessary for physical support of a retained object. Any voxel that would otherwise be hidden behind a visible neighbor without supporting a retained object is removed from the scene. As an example, consider a row of three objects in front of a single object behind the middle of the row: the rear object is fully occluded by the visible row, contributes no pixels to the rendered image, and is not required to support any retained object. Our procedure removes it. By contrast, a single visible object resting on top of an otherwise invisible object in the same column passes the test on the lower object via the support criterion, because the lower object is required to keep the upper object physically supported. This guarantees that every hidden object in the final scene is logically required by support and is never merely concealed behind a visible neighbor at an unrecoverable depth.

\paragraph{Ground-truth generation.}
Because we control the scene at the voxel level, all sub-task and target-task answers are derived analytically from the same data structure rather than annotated post hoc. Integer and label answers (e.g., column count, layer count, total object count, object category) are computed from the retained-object voxel grid and written to a per-view JSON file together with the camera parameters. Image-editing ground-truth images are generated by re-rendering the same scene with task-specific colorings applied through the Replicator semantic-segmentation annotator: for each image-editing task, the relevant subset of objects (e.g., the directly supporting objects below the top-most layer, for \textit{Direct Support}) is colored according to the task specification, and the rendered image with those colorings replaces the standard texture pass. This construction guarantees that ground truth and the rendered scene are pixel-aligned and that the ground-truth coloring uses exactly the same object geometry, lighting, and camera as the standard render.

\paragraph{Dataset partitioning.}
The full dataset comprises approximately 80{,}000 scenes featuring distinct structures, object categories, and environments. From this pool, we draw two disjoint subsets. The primary benchmark is a 3{,}000-sample evaluation set used for all human and model evaluations reported in this paper. For training experiments (Section~\ref{sec:training}), we use a separate pool of 136{,}171 views of 8{,}521 structures, with no scene structure shared across the evaluation and training partitions.

\section{Benchmark Validation Details}
\label{app:validation}

This appendix section supplements Section~\ref{sec:validation} of the main paper with full figures and numerical tables for the MCQ and image-output modalities, camera and object robustness analyses, and the mental rotation results that motivated focusing the main-paper analysis on \textit{Object Counting}.

\subsection{Mental Rotation Results}
\label{app:mental_rotation}

\paragraph{Task accuracy and spread.}
We report per-task accuracy for both the $90^\circ$ and $180^\circ$ mental rotation conditions across all text models and the human baseline (cross-view columns of Table~\ref{tab:text_task_accuracy_full}). The human baseline sits at $46.4\%$ and $45.8\%$ respectively, well below every counting sub-task for humans (lowest: \textit{Column Count}, $63.4\%$), providing the empirical basis for treating \textit{Mental Rotation} as a complementary rather than primary probe in Section~\ref{sec:mental_rotation_complement}. Among models, \textit{Mental Rotation} ranges from $3.5\%$ in GLM to $40.1\%$ in Claude on the $90^\circ$ condition, with a similar ordering at $180^\circ$. Notably, this is a different model ranking than \textit{Object Counting}: Claude leads the cross-view branch despite ranking third on the counting target, while Qwen leads the counting target but sits near the bottom on cross-view (seventh of eight). The two arms therefore separate models on different competencies.

\begin{figure}[t]
\centering
\includegraphics[width=\textwidth]{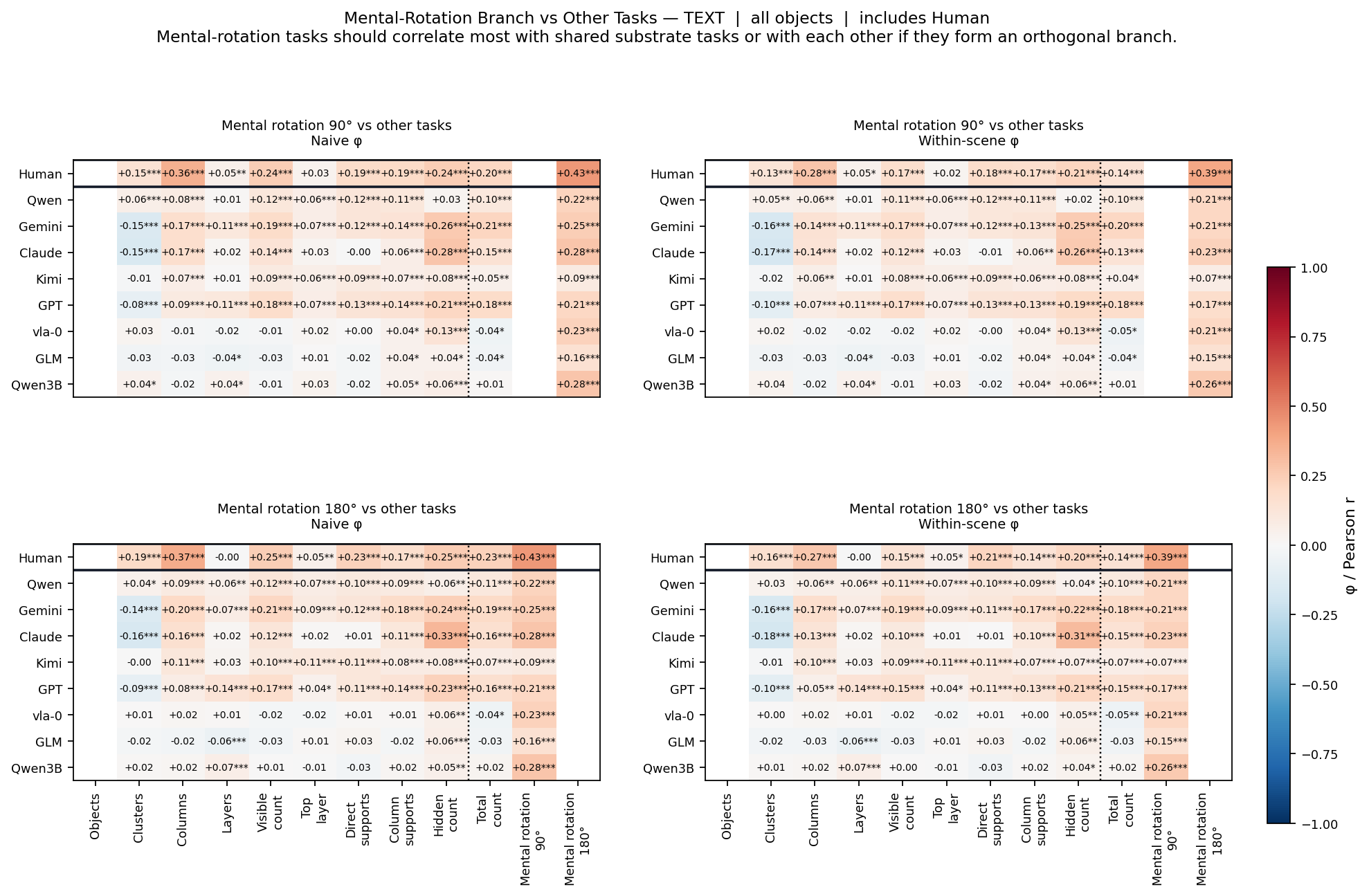}
\caption{Cross-view dependency correlations. Naive $\phi$ (left panels) and within-scene $\phi$ (right panels) between the two \textit{Mental Rotation} tasks ($90^\circ$ and $180^\circ$) and each counting sub-task, per model. Within-scene correlations control for shared scene difficulty by varying the parent across responders within the same scene. Persistent high values indicate a genuinely shared cognitive basis rather than scene-level confounding. For the human baseline and four of the eight models, the two \textit{Mental Rotation} tasks correlate most strongly with each other (median within-scene $\phi \approx 0.21$) and only weakly with the counting sub-tasks. For Claude, Gemini, GPT and Kimi, the within-scene $\phi$ with \textit{Hidden Object Count} is comparable to that value or larger.}
\label{fig:mr1_crossview_dependencies}
\end{figure}

\paragraph{Cross-view tasks correlate with each other more than with most counting tasks.}
Figure~\ref{fig:mr1_crossview_dependencies} shows naive and within-scene $\phi$ between the cross-view tasks and every other task in the benchmark, separately for each model. For the human baseline and four of the eight models, the two \textit{Mental Rotation} variants correlate most strongly with each other: the within-scene $\phi$ between $90^\circ$ and $180^\circ$ is $+0.39$ for the human baseline and remains around $+0.17$--$0.23$ for the frontier models (Kimi $0.07$). For Claude, Gemini, GPT and Kimi, the within-scene $\phi$ with \textit{Hidden Object Count} is comparable to the $90^\circ$--$180^\circ$ value or larger. Associations with the other counting sub-tasks are weaker, with most cells in the within-scene panels sitting between $-0.05$ and $+0.20$. The strongest single-task association at the within-scene level is with \textit{Hidden Object Count}, but no multi-task chain reaches the level of coupling observed for \textit{Object Counting}.

\begin{figure}[t]
\centering
\includegraphics[width=\textwidth]{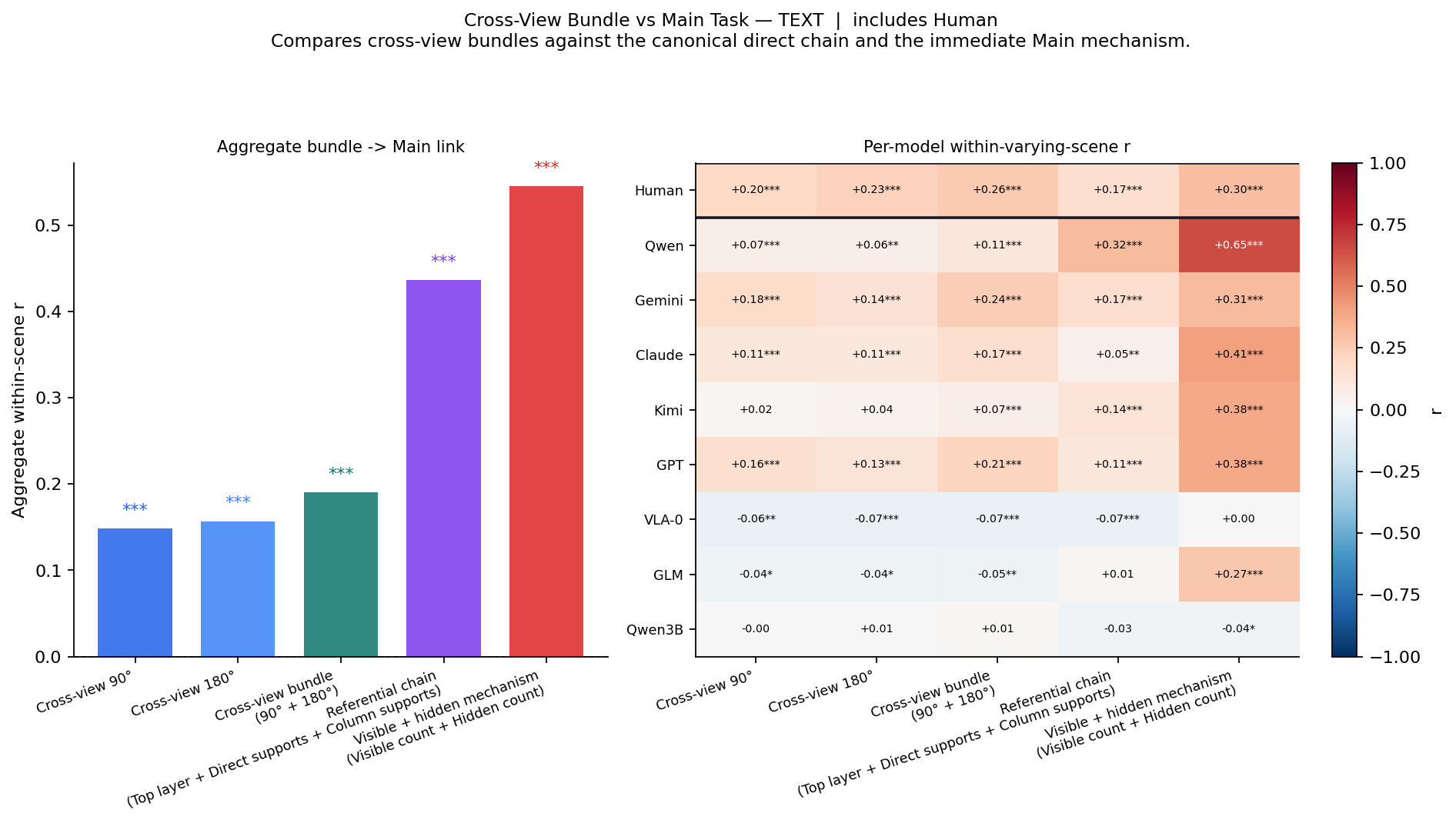}
\caption{Cross-view bundle versus \textit{Object Counting} mechanism comparison. Aggregate within-scene Pearson $r$ between each task bundle and \textit{Object Counting} (left: pooled within-scene correlation, right: per-model heatmap). The \textit{Mental Rotation} bundle ($90^\circ + 180^\circ$, $r \approx 0.19$) is roughly 2--3$\times$ weaker than the Internal Referential Chain ($r \approx 0.44$) and the Summation Mechanism ($r \approx 0.54$), with all five aggregate effects FDR-significant.}
\label{fig:mr2_crossview_bundle}
\end{figure}

\paragraph{Hierarchy coupling.}
Figure~\ref{fig:mr2_crossview_bundle} compares the within-scene coupling of each pre-specified bundle to \textit{Object Counting}. The aggregate within-scene correlations are $r_{ws} = 0.148$ for \textit{Mental Rotation} $90^\circ$ alone, $0.157$ for $180^\circ$ alone, and $0.190$ for the combined cross-view bundle, against $0.436$ for the Internal Referential Chain (\textit{Top Layer} + \textit{Direct Support} + \textit{Support Column}) and $0.545$ for the \textit{Summation Mechanism} (\textit{Visible Object Count} + \textit{Hidden Object Count}). For comparison, in the human baseline, all four counting bundles correlate positively with \textit{Object Counting}: the Visible Count Hierarchy bundle at within-responder $r = 0.40$, the Hidden Count Hierarchy bundle at $r = 0.30$, the Summation Mechanism bundle at $r = 0.30$, and the Internal Referential Chain at $r = 0.17$, rising to $r = 0.59$, $0.50$, $0.54$, and $0.44$ respectively at the aggregate within-scene level. The cross-view bundle is therefore substantially weaker than every counting bundle in aggregate. Per model, it exceeds the Internal Referential Chain bundle for Claude, Gemini, GPT, Qwen3B and the human baseline. We additionally ran a within-structure fixed-effects regression (not shown in figure form) testing whether the cross-view tasks add any incremental \textit{Object Counting} signal beyond the Internal Referential Chain and camera fixed effects: the per-model coefficients are uniformly small (mean $|\beta| \approx 0.015$) and almost all non-significant, with only one model showing a marginally negative \textit{Mental Rotation} $180^\circ$ coefficient. These coefficients support the reading that responses align with the proposed hierarchy for \textit{Object Counting} but not for \textit{Mental Rotation}, and that the two arms probe largely separable competencies.

\subsection{Task Accuracy: Text, MCQ, and Image Modalities}
\label{app:task_accuracy_modalities}

This appendix subsection provides the per-task accuracy figures and full numerical tables for all three response modalities, supplementing Figure~\ref{fig:task_accuracy_text} in the main paper.

\paragraph{Text-modality task accuracy: full numerical table.}
Table~\ref{tab:text_task_accuracy_full} reports per-task accuracy with Wilson 95\% confidence intervals on \textit{Object Counting} and per-task means for each model in the text modality, providing the raw data underlying Figure~\ref{fig:task_accuracy_text} of the main paper. Across the text free-response models, sub-task performance varies considerably. \textit{Layer Count} is typically the easiest sub-task, reaching $77.0\%$ in Qwen and at or above $63\%$ in five of the eight models. \textit{Column Count} and \textit{Visible Object Count} are among the hardest, with most models below $20\%$. Between these extremes, \textit{Direct Support} and \textit{Support Column} sit in a middle tier, and \textit{Top Layer} produces unusually wide between-model spread (from $15.5\%$ in VLA-0 to $44.9\%$ in Gemini, with Qwen lowest among the frontier models at $21.1\%$ despite leading on \textit{Object Counting}). The fact that models with similar \textit{Object Counting} accuracy (e.g., Gemini and Claude) differ substantially on these lower-level competencies is the property exploited in the analyses of Section~\ref{sec:analysis}.

\begin{table}[t]
\centering
\small
\caption{Text-modality per-task accuracy by model. Raw data underlying main-paper Figure~\ref{fig:task_accuracy_text}. \textit{Object Counting} is reported with Wilson 95\% confidence intervals; sub-task and \textit{Mental Rotation} columns report point accuracy. All cells are evaluated on the same $n=3000$ views. Wilson intervals are computed at the view level and do not account for dependence among views derived from the same structure. \textit{Object Categorization} (S1) is not scored in the text modality. Pairwise rank differences are tested with paired exact McNemar (two-sided, BH FDR corrected across model pairs).}
\label{tab:text_task_accuracy_full}
\begin{adjustbox}{max width=\textwidth}
\begin{tabular}{l c c c c c c c c c c c}
\toprule
\textbf{Responder} & \textbf{Obj. Count [95\% CI]} & \textbf{Cluster} & \textbf{Column} & \textbf{Layer} & \textbf{Vis. Count} & \textbf{Top Layer} & \textbf{Dir. Sup.} & \textbf{Sup. Col.} & \textbf{Hid. Count} & \textbf{MR $90^\circ$} & \textbf{MR $180^\circ$} \\
\midrule
Human   & 82.1\% [80.7, 83.5] & 71.3\% & 63.4\% & 96.2\% & 76.8\% & 96.4\% & 80.0\% & 83.5\% & 79.9\% & 46.4\% & 45.8\% \\
Qwen    & 17.7\% [16.4, 19.1] & 45.8\% & 26.1\% & 77.0\% & 19.2\% & 21.1\% & 50.9\% & 42.1\% & 49.2\% & 15.9\% & 14.1\% \\
Gemini  & 14.8\% [13.6, 16.2] & 46.7\% & 14.4\% & 70.2\% & 17.7\% & 44.9\% & 47.7\% & 37.9\% & 43.6\% & 37.8\% & 35.1\% \\
Claude  & 14.1\% [12.9, 15.4] & 41.1\% & 19.1\% & 63.0\% & 15.7\% & 26.5\% & 28.8\% & 30.2\% & 51.4\% & 40.1\% & 36.4\% \\
Kimi    & 13.6\% [12.4, 14.9] & 39.4\% &  7.8\% & 66.6\% & 11.5\% & 28.1\% & 26.8\% & 24.7\% & 33.1\% & 18.9\% & 17.6\% \\
GPT     &  8.3\% [7.4, 9.4]   & 42.9\% &  9.5\% & 65.2\% &  8.7\% & 36.6\% & 33.2\% & 31.9\% & 47.4\% & 35.1\% & 31.7\% \\
VLA-0   &  4.2\% [3.5, 4.9]   & 23.0\% &  0.1\% & 17.6\% &  1.2\% & 15.5\% & 37.3\% & 15.2\% & 19.0\% & 22.0\% & 21.4\% \\
GLM     &  3.5\% [2.9, 4.2]   & 21.0\% &  2.1\% & 21.0\% &  2.9\% & 24.2\% & 16.8\% & 14.6\% & 16.4\% &  3.5\% &  2.8\% \\
Qwen3B  &  2.1\% [1.7, 2.7]   & 22.9\% &  0.1\% &  9.8\% &  0.2\% & 20.0\% & 36.8\% & 13.0\% & 19.0\% & 25.7\% & 24.1\% \\
\bottomrule
\end{tabular}
\end{adjustbox}
\end{table}

\begin{table}[t]
\centering
\small
\caption{MCQ-modality (5CQ) per-task accuracy by model. Companion to Table~\ref{tab:text_task_accuracy_full}, computed on the same $n=3000$ views. \textit{Object Counting} is reported with view-level Wilson 95\% confidence intervals; sub-task and \textit{Mental Rotation} columns report point accuracy. Models are ordered by \textit{Object Counting} accuracy. Per-cell counts and Wilson CIs are exported in \texttt{table\_07\_mcq5\_task\_accuracy.csv}.}
\label{tab:mcq5_task_accuracy_full}
\begin{adjustbox}{max width=\textwidth}
\begin{tabular}{l c c c c c c c c c c c c}
\toprule
\textbf{Responder} & \textbf{Obj. Count [95\% CI]} & \textbf{Cluster} & \textbf{Column} & \textbf{Layer} & \textbf{Vis. Count} & \textbf{Top Layer} & \textbf{Dir. Sup.} & \textbf{Sup. Col.} & \textbf{Hid. Count} & \textbf{MR $90^\circ$} & \textbf{MR $180^\circ$} \\
\midrule
Human   & 86.2\% [84.9, 87.4] & 91.2\% & 93.6\% & 99.4\% & 99.4\% & 99.6\% & 97.6\% & 97.7\% & 91.5\% & 96.5\% & 97.1\% \\
GPT     & 20.7\% [19.3, 22.2] & 35.1\% & 23.8\% & 31.6\% & 39.2\% & 22.2\% & 42.1\% & 43.7\% & 30.1\% & 22.3\% & 20.7\% \\
Kimi    & 20.5\% [19.1, 21.9] & 34.8\% & 40.0\% & 55.4\% & 46.6\% & 38.2\% & 49.9\% & 50.4\% & 29.7\% & 22.6\% & 23.2\% \\
GLM     & 20.3\% [18.9, 21.8] & 19.0\% & 20.4\% & 20.3\% & 20.3\% & 19.0\% & 20.5\% & 19.6\% & 20.7\% & 19.4\% & 20.0\% \\
VLA-0   & 20.3\% [18.9, 21.8] & 20.6\% & 19.9\% & 19.5\% & 21.1\% & 20.0\% & 19.3\% & 19.8\% & 20.3\% & 19.7\% & 19.0\% \\
Qwen    & 20.2\% [18.8, 21.6] & 20.0\% & 22.8\% & 15.6\% & 26.4\% & 27.8\% & 29.1\% & 26.4\% & 18.4\% & 15.6\% & 16.6\% \\
Qwen3B  & 20.2\% [18.8, 21.6] & 20.7\% & 20.1\% & 20.4\% & 20.8\% & 20.7\% & 19.7\% & 19.8\% & 20.3\% & 19.3\% & 20.0\% \\
Claude  & 20.0\% [18.6, 21.5] & 39.8\% & 31.5\% & 37.8\% & 52.6\% & 35.7\% & 36.0\% & 39.1\% & 29.4\% & 41.7\% & 40.2\% \\
Gemini  & 19.5\% [18.2, 21.0] & 38.0\% & 28.2\% & 48.0\% & 46.1\% & 31.1\% & 36.5\% & 33.3\% & 31.3\% & 29.8\% & 27.1\% \\
\bottomrule
\end{tabular}
\end{adjustbox}
\end{table}

\begin{table}[t]
\centering
\small
\caption{Image-editing modality per-task accuracy by model. Companion to Table~\ref{tab:text_task_accuracy_full}, computed on the same $n=3000$ views. \textit{Object Counting} is reported with view-level Wilson 95\% confidence intervals; sub-task and \textit{Mental Rotation} columns report point accuracy. Per-cell details are exported in \texttt{table\_08\_image\_task\_accuracy.csv}.}
\label{tab:image_task_accuracy_full}
\begin{adjustbox}{max width=\textwidth}
\begin{tabular}{l c c c c c c c c c c c c}
\toprule
\textbf{Responder} & \textbf{Obj. Count [95\% CI]} & \textbf{Cluster} & \textbf{Column} & \textbf{Layer} & \textbf{Vis. Count} & \textbf{Top Layer} & \textbf{Dir. Sup.} & \textbf{Sup. Col.} & \textbf{Hid. Count} & \textbf{MR $90^\circ$} & \textbf{MR $180^\circ$} \\
\midrule
Gemini Flash Image    & 14.7\% [13.5, 16.0] & 47.1\% & 29.3\% & 24.1\% & 36.1\% & 34.3\% & 31.2\% & 22.6\% & 46.5\% & 27.5\% & 26.4\% \\
Qwen-Image-Edit       & 12.0\% [10.9, 13.2] &  6.9\% &  5.5\% &  5.6\% &  3.3\% & 15.3\% & 11.9\% &  7.6\% & 38.7\% & 31.5\% & 31.3\% \\
HunyuanImage-Instruct &  5.9\% [5.1, 6.8]   & 18.7\% &  6.3\% &  4.0\% & 10.8\% & 29.5\% & 19.0\% & 12.7\% & 30.5\% & 15.8\% & 16.0\% \\
\bottomrule
\end{tabular}
\end{adjustbox}
\end{table}

\begin{figure}[t]
\centering
\includegraphics[width=\textwidth]{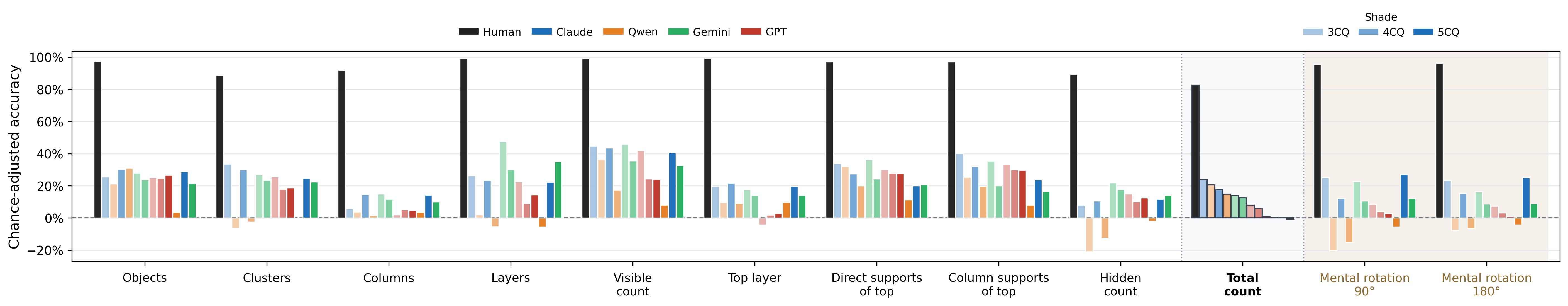}
\caption{Chance-adjusted per-task accuracy under five-, four-, and three-choice MCQ for the four overlapping models, with the human five-choice baseline (black). Shading darkness encodes answer-space size (palest = three-choice; darkest = five-choice). \textit{Object Counting}, \textit{Visible Object Count}, and \textit{Direct Support} show monotone recovery as the answer space shrinks; \textit{Hidden Object Count} shows model-specific recovery with Qwen's anomalous downward trajectory documented in Appendix~\ref{app:mcq_qwen_t8}.}
\label{fig:mcq_c1a_chance_adjusted}
\end{figure}

\paragraph{Task accuracy in the five-, four-, and three-choice MCQ conditions.}
Figure~\ref{fig:mcq_c1a_chance_adjusted} reports chance-adjusted per-task accuracy for the four overlapping models (Claude, Qwen, Gemini, GPT) under five-, four-, and three-choice MCQ conditions, alongside the human five-choice baseline. The directional summary is that reducing the answer space substantially recovers \textit{Object Counting}, but the recovery is uneven across sub-tasks. For \textit{Object Counting}, chance-adjusted accuracy rises monotonically for all four models: Claude moves from $0.000$ ($5$CQ) to $0.178$ ($4$CQ) to $0.237$ ($3$CQ); Qwen from $0.002 \to 0.148 \to 0.205$; GPT from $0.009 \to 0.059 \to 0.079$; and Gemini from $-0.006 \to 0.128 \to 0.140$. The clearest sub-task recovery occurs on \textit{Visible Object Count}: under three-choice MCQ, Gemini reaches $0.459$, Claude $0.446$, GPT $0.421$, and Qwen $0.365$. The hidden-side branch behaves differently. Claude changes only weakly on \textit{Hidden Object Count} ($0.117 \to 0.107 \to 0.080$), GPT remains modest ($0.126 \to 0.104 \to 0.150$), and Qwen remains below chance throughout ($-0.020 \to -0.126 \to -0.209$); Gemini is the notable exception, with a positive monotone trajectory ($0.141 \to 0.177 \to 0.219$). This branch-asymmetric recovery pattern suggests that visible-object aggregation benefits most from answer-space reduction while hidden-object reasoning remains the more persistent bottleneck.

\begin{figure}[t]
\centering
\includegraphics[width=\textwidth]{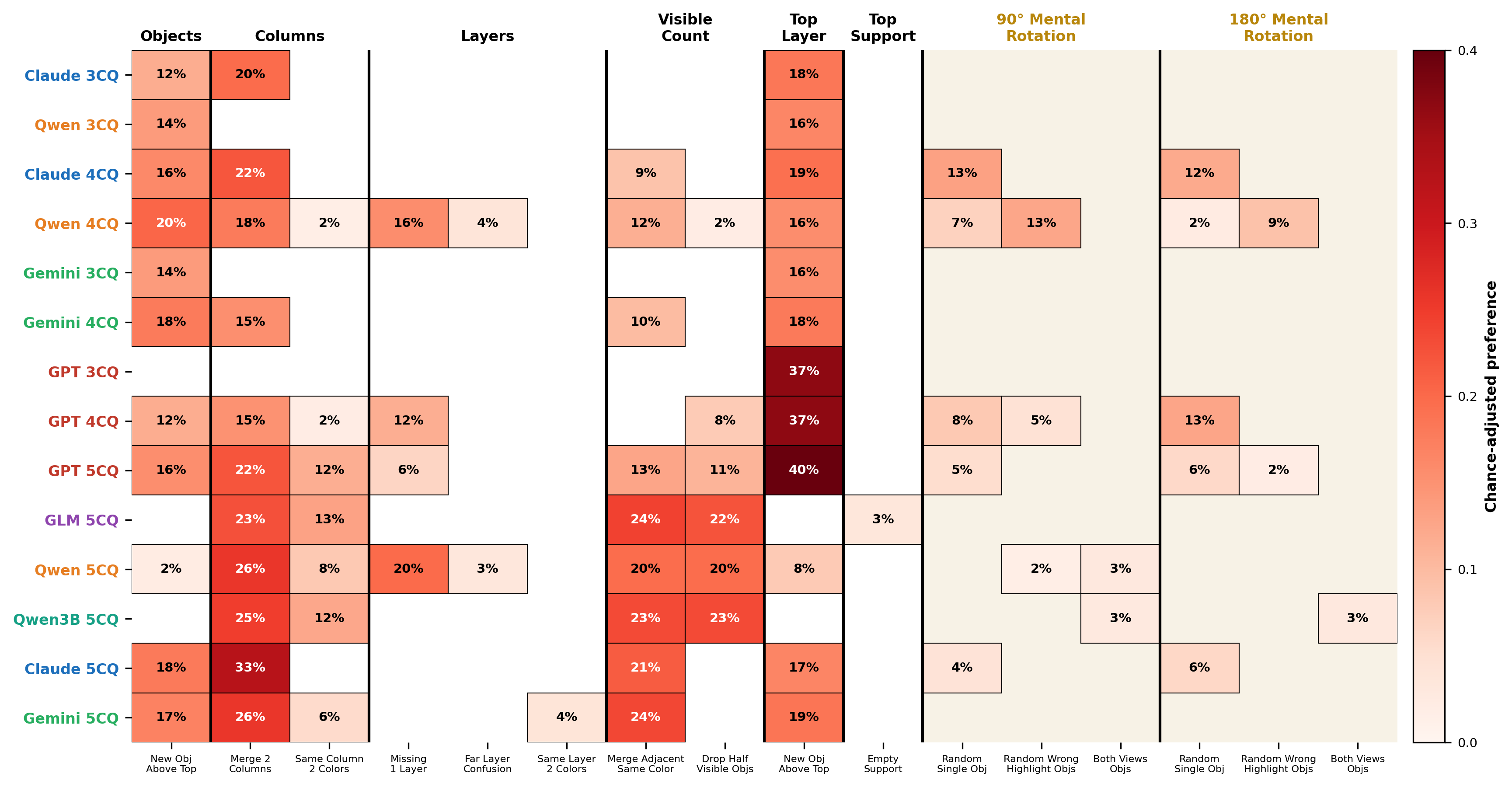}
\caption{Mixed-CQ wrong-answer preference heatmap. Each cell reports the chance-adjusted preference for a given distractor for a given model and CQ condition. Tasks are separated by solid vertical lines; wrong-answer reasons within a task are separated by dotted lines. The dominant per-task distractor preferences are stable across $5$CQ, $4$CQ, and $3$CQ.}
\label{fig:mcq_m1_345}
\end{figure}

\paragraph{Wrong-answer preferences across reduced answer spaces.}
Figure~\ref{fig:mcq_m1_345} repeats the wrong-answer heatmap of Figure~\ref{fig:mcq_wrong_answers} on a chance-adjusted scale across all three MCQ answer-space conditions. The dominant distractor preferences persist across answer-space reductions: \textit{Object Categorization} and \textit{Top Layer} remain dominated by \textit{phantom block above top}; \textit{Column Count} remains dominated by \textit{merge two columns} with \textit{same column, two colors} as the secondary confusion; \textit{Visible Object Count} continues to be dominated by \textit{merge adjacent same color} and \textit{drop half visible blocks}; and the hidden-side tasks remain concentrated on the same narrow confusion families. This stability across answer-space conditions is the basis for treating these distractors as stable counting confusions rather than artifacts of the five-choice format. GPT shows a large \textit{phantom block above top} preference on \textit{Top Layer} at every answer-set size ($37$--$40\%$, about twice any other model), while its \textit{Object Categorization} preference fades to chance at three choices.

\paragraph{MCQ-modality task accuracy: full numerical table.}
Table~\ref{tab:mcq5_task_accuracy_full} reports per-task accuracy for the 5CQ condition with Wilson 95\% confidence intervals on \textit{Object Counting} and per-task means for each model, providing the raw data underlying the five-choice column of Figure~\ref{fig:mcq_c1a_chance_adjusted}. Two patterns stand out. First, the human row remains near ceiling on every sub-task ($\geq 91\%$ across the board), while every model row collapses to chance on \textit{Object Counting} (between 19.5\% and 20.7\%), reproducing the headline finding from the main paper. Second, the sub-task structure is preserved across models even when the target task is at chance: \textit{Visible Object Count} ranges from $20.3\%$ in GLM and $20.8\%$ in Qwen3B (at chance) to $52.6\%$ in Claude, and \textit{Top Layer}, \textit{Direct Support}, and \textit{Support Column} similarly separate models into a recoverable mid-tier (Claude, Gemini, GPT, Kimi) and a chance-only tier (GLM, Qwen3B, VLA-0). Qwen sits between these, with above-chance \textit{Visible Object Count} ($26.4\%$) but below-chance \textit{Hidden Object Count} ($18.4\%$), the same below-chance pattern discussed in Appendix~\ref{app:mcq_qwen_t8}.

\begin{figure}[t]
\centering
\includegraphics[width=\textwidth]{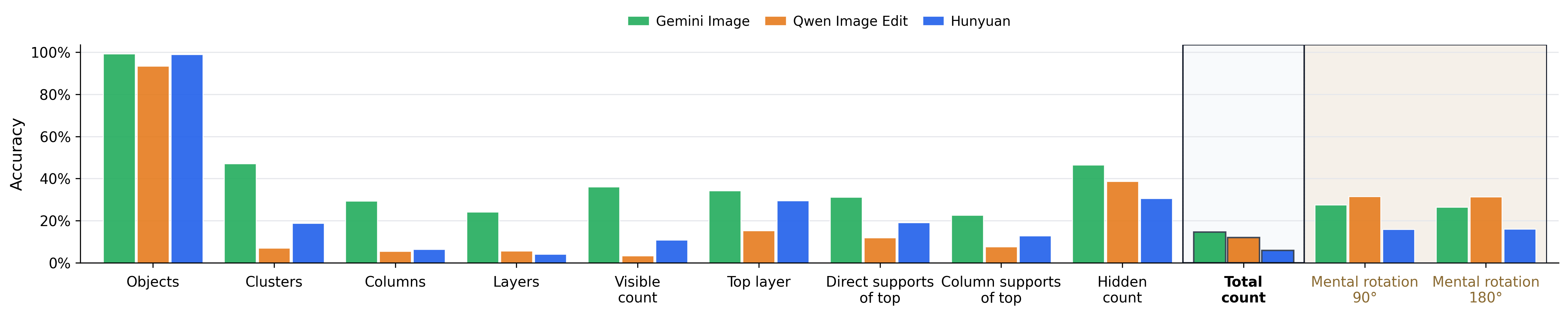}
\caption{Image-output per-task accuracy for Gemini Flash Image, Qwen-Image-Edit, and HunyuanImage-Instruct, arranged in the same Piagetian task order as Figure~\ref{fig:task_accuracy_text} in the main paper. \textit{Object Counting} (boxed at right of the counting region) is the hardest target for all three responders; the two \textit{Mental Rotation} tasks (shaded at far right) form the parallel cross-view evaluation arm.}
\label{fig:ig_c1_task_accuracy}
\end{figure}

\paragraph{Image-editing modality task accuracy: full numerical table.}
Table~\ref{tab:image_task_accuracy_full} reports per-task accuracy with Wilson 95\% confidence intervals on \textit{Object Counting} and per-task means for the three image-editing models, providing the raw data underlying Figure~\ref{fig:ig_c1_task_accuracy}. The three models occupy different competence profiles. Gemini Flash Image leads on the perceptual sub-tasks (\textit{Cluster Count} $47.1\%$, \textit{Visible Object Count} $36.1\%$, \textit{Hidden Object Count} $46.5\%$) and on the target task ($14.7\%$). Qwen-Image-Edit collapses on the visible-counting branch (\textit{Visible Object Count} $3.3\%$, \textit{Cluster Count} $6.9\%$) but remains competitive on \textit{Hidden Object Count} ($38.7\%$) and \textit{Mental Rotation} ($31.5\%$ at $90^\circ$, $31.3\%$ at $180^\circ$). HunyuanImage-Instruct sits between these on most sub-tasks but trails on the target task ($5.9\%$). The image modality therefore separates models by which branch of the proposed hierarchy they preserve, rather than along a single axis of overall competence.

\paragraph{Task accuracy in the image-output modality.}
Figure~\ref{fig:ig_c1_task_accuracy} reports per-task accuracy for the three image-output models. The image-generation profile is strongly differentiated across both tasks and responders. \textit{Object Categorization}, the easiest task in this modality, sits near ceiling for all three responders ($99.3\%$ for Gemini Flash Image, $93.4\%$ for Qwen-Image-Edit, $98.9\%$ for HunyuanImage-Instruct), confirming that simple foreground-object identification is broadly tractable. The differences emerge on the structured sub-tasks. Gemini Flash Image is the most balanced, with comparatively strong \textit{Cluster Count} ($47.1\%$), \textit{Hidden Object Count} ($46.5\%$), and \textit{Visible Object Count} ($36.1\%$). Qwen-Image-Edit is more polarized: it remains reasonably competitive on \textit{Hidden Object Count} ($38.7\%$) and the two \textit{Mental Rotation} tasks ($31.5\%$, $31.3\%$), but collapses on the visible-count branch (\textit{Visible Object Count} = $3.3\%$). HunyuanImage-Instruct fills in the third profile: it reaches $30.5\%$ on \textit{Hidden Object Count} and $29.5\%$ on \textit{Top Layer}, but remains weaker on \textit{Visible Object Count} ($10.8\%$) and \textit{Mental Rotation} ($15.8\%$, $16.0\%$). \textit{Object Counting} is the hardest target for all three: Gemini Flash Image reaches $14.7\%$, Qwen-Image-Edit $12.0\%$, and HunyuanImage-Instruct $5.9\%$. The modality is therefore not flat in either direction: it separates models by which branch of the task graph they preserve, and by how much of that local structure survives into the final count.

\subsection{Difficulty Controls: Text, MCQ, and Image Modalities}
\label{app:difficulty_modalities}

This appendix subsection supplements Figure~\ref{fig:difficulty_text} of the main paper with the full difficulty-bin numerical table for the text modality, the per-task camera robustness analysis, and the parallel difficulty-control analysis for the image modality.

\paragraph{Pooled difficulty-bin accuracies in text.}
Table~\ref{tab:text_difficulty_bins} reports pooled \textit{Object Counting} accuracy across the model set as a function of each candidate difficulty control, providing the raw data underlying Figure~\ref{fig:difficulty_text} of the main paper. Pooled \textit{Object Counting} falls from $31.5\%$ in the easiest \textit{Number of objects} bin to $1.0\%$ in the hardest, from $17.6\%$ at zero hidden objects to $0.8\%$ at four or more, and from $35.2\%$ at one layer to $0.9\%$ at four. \textit{Fill ratio} moves in the opposite direction (sparse $5.0\%$, dense $13.9\%$), consistent with denser scenes reducing the visible-hidden gap. The point-biserial correlation between each structural factor and target-task correctness is consistently negative for total objects, hidden objects, columns, and layers, with model-by-model values ranging from approximately $-0.10$ to $-0.44$ in text. \textit{Fill ratio} produces positive correlations in the same range ($+0.110$ to $+0.219$), consistent with the dense-scene-as-rectangular-prism interpretation in Section~\ref{sec:validation_difficulty}.

\begin{table}[t]
\centering
\small
\caption{Pooled \textit{Object Counting} accuracy by difficulty bin in the text modality. Raw data underlying main-paper Figure~\ref{fig:difficulty_text}. Pooled across the eight text models reported in Table~\ref{tab:text_task_accuracy_full}.}
\label{tab:text_difficulty_bins}
\begin{tabular}{l l r r r r}
\toprule
\textbf{Factor} & \textbf{Bin} & \textbf{Pooled acc.} & \textbf{SE} & \textbf{95\% CI} & \textbf{$n$} \\
\midrule
\multirow{5}{*}{Number of objects}
& 4--8   & 31.5\% & 0.007 & [30.1, 32.9] &  4{,}288 \\
& 9--12  & 11.2\% & 0.004 & [10.5, 12.0] &  7{,}128 \\
& 13--18 &  2.6\% & 0.002 & [2.2, 3.1]   &  5{,}288 \\
& 19--25 &  0.8\% & 0.001 & [0.6, 1.1]   &  4{,}528 \\
& 26+    &  1.0\% & 0.002 & [0.7, 1.5]   &  2{,}768 \\
\midrule
\multirow{4}{*}{Number of hidden objects}
& 0   & 17.6\% & 0.003 & [16.9, 18.2] & 12{,}152 \\
& 1   &  3.6\% & 0.003 & [3.1, 4.2]   &  4{,}088 \\
& 2--3 &  1.0\% & 0.002 & [0.8, 1.4]  &  3{,}448 \\
& 4+  &  0.8\% & 0.001 & [0.6, 1.1]   &  4{,}312 \\
\midrule
\multirow{4}{*}{Number of layers}
& 1 & 35.2\% & 0.012 & [33.0, 37.5] &  1{,}696 \\
& 2 & 15.4\% & 0.004 & [14.6, 16.1] &  9{,}312 \\
& 3 &  3.4\% & 0.002 & [3.0, 3.8]   &  8{,}352 \\
& 4 &  0.9\% & 0.001 & [0.7, 1.2]   &  4{,}640 \\
\midrule
\multirow{4}{*}{Number of columns}
& 3--5   & 40.6\% & 0.011 & [38.4, 42.9] &  1{,}888 \\
& 6--8   & 13.6\% & 0.004 & [12.9, 14.4] &  8{,}320 \\
& 9--11  &  4.1\% & 0.002 & [3.7, 4.5]   &  9{,}728 \\
& 12--14 &  1.3\% & 0.002 & [1.0, 1.7]   &  4{,}064 \\
\midrule
\multirow{3}{*}{Fill ratio}
& Dense  & 13.9\% & 0.004 & [13.1, 14.7] &  6{,}760 \\
& Medium & 10.1\% & 0.003 & [9.6, 10.7]  & 10{,}840 \\
& Sparse &  5.0\% & 0.003 & [4.4, 5.5]   &  6{,}400 \\
\bottomrule
\end{tabular}
\end{table}

\begin{figure}[t]
\centering
\includegraphics[width=\textwidth]{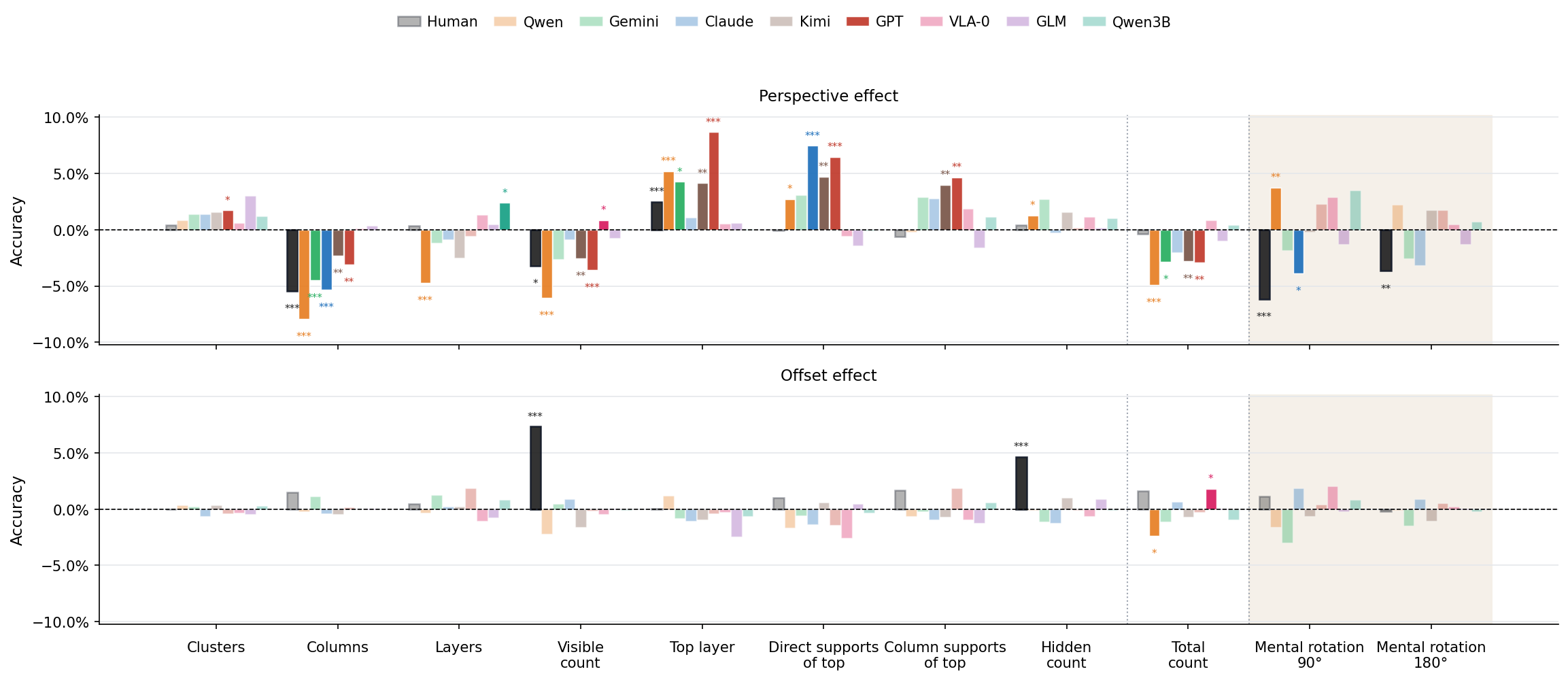}
\caption{Within-structure camera-parameter effects per task in text. Top: perspective deltas (strong $-$ weak perspective). Bottom: offset deltas ($12.5^\circ - 3.0^\circ$ azimuth). Asterisks denote BH-FDR-corrected significance per responder. Visible-structure tasks (\textit{Column Count}, \textit{Visible Object Count}) absorb the largest negative perspective effects. Perspective gains are positive for \textit{Top Layer} in every responder and for \textit{Direct Support} and \textit{Support Column} in most, and \textit{Hidden Object Count} is near-flat under perspective (at most $+2.7$\,pp).}
\label{fig:c13_camera_within_structure}
\end{figure}

\paragraph{Camera variation effects: per-task analysis.}
Although camera factors produce near-zero point-biserial correlations with \textit{Object Counting} correctness across models and the human baseline, this aggregate verdict masks meaningful task-level heterogeneity. Figure~\ref{fig:c13_camera_within_structure} reports within-structure perspective and offset deltas separately for each task and model. Within paired 3D structures, stronger perspective (raw $\textsf{fov}=3$ versus $\textsf{fov}=1$) significantly hurts accuracy on the visible-structure tasks: \textit{Column Count} and \textit{Visible Object Count} show the largest negative perspective effects (down to $-7.9$\,pp on \textit{Column Count} for Qwen, FDR-significant for six of the nine responders, including the human baseline), while \textit{Hidden Object Count} effects are small (at most $+2.7$\,pp). Perspective effects are positive for \textit{Top Layer} in every responder and for \textit{Direct Support} and \textit{Support Column} in most: stronger perspective makes the top of the structure easier to read out, with GPT improving by $+8.7$\,pp on \textit{Top Layer} and Claude by $+7.5$\,pp on \textit{Direct Support} under strong perspective. Camera offset effects are mostly non-significant across tasks and models, with few exceptions: the human baseline ($+7.3$\,pp on \textit{Visible Object Count}, $+4.7$\,pp on \textit{Hidden Object Count}) and small \textit{Object Counting} effects for Qwen ($-2.4$\,pp) and VLA-0 ($+1.8$\,pp). This per-task pattern fits a reading that separates perception from reasoning: perspective mainly changes how objects appear in a single rendered view, so the perceptual visible-structure tasks are most affected, while hidden-object inference moves little.

\paragraph{Distinct-color control.}
\label{app:distinct_color}
We re-rendered a subset of benchmark views with distinctly colored objects, keeping structures, counts and viewpoints identical and changing only appearance, which removes most of the individuation difficulty. \textit{Object Counting} rises from $16.5\%$ (the same views in standard rendering) to $24.3\%$ for Qwen, from $15.0\%$ to $21.5\%$ for Gemini, and from $14.3\%$ to $21.6\%$ for Claude, and models remain far below the human $82.1\%$ in the standard condition. We do not adopt distinct coloring as a default because it also removes part of the reasoning demand the task is meant to probe. On the standard benchmark, we can instead condition on correct \textit{Cluster Count}, \textit{Column Count}, \textit{Layer Count} and \textit{Visible Object Count}. \textit{Object Counting} then reaches $78.9\%$ for Qwen ($n = 123$) and $71.2\%$ for Gemini ($n = 80$), well above their unconditioned accuracy, which confirms that many counting errors co-occur with perceptual errors. The remaining failure rate is $37.0\%$ for GPT ($n = 27$) against $4.4\%$ for humans ($n = 1310$), so a reasoning gap survives on scenes where perception is verified correct. Perception and reasoning remain partly entangled in the raw task, and neither control isolates them completely.

\begin{figure}[t]
\centering
\includegraphics[width=\textwidth]{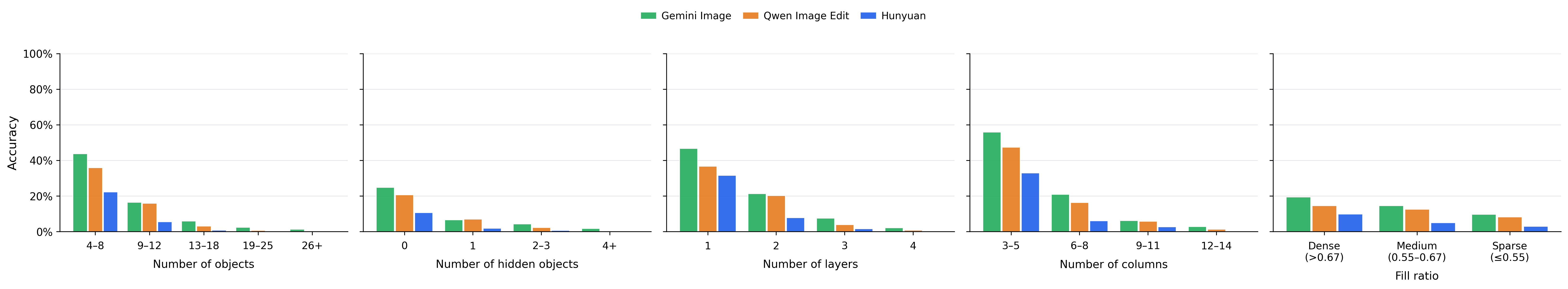}
\caption{\textit{Object Counting} accuracy as a function of the five candidate difficulty controls in the image-output modality (Gemini Flash Image, Qwen-Image-Edit, HunyuanImage-Instruct), paralleling Figure~\ref{fig:difficulty_text} in the main paper. The directional pattern (negative for object count, hidden count, layers, columns; positive for fill ratio) is preserved from text, confirming that the difficulty controls are not modality-specific artifacts.}
\label{fig:ig_c2_difficulty}
\end{figure}

\paragraph{Difficulty controls in the image-output modality.}
Figure~\ref{fig:ig_c2_difficulty} reports the parallel five-axis difficulty breakdown for the three image-output models. The structural difficulty controls remain coherent across all three responders. For Gemini Flash Image, the strongest negative associations with \textit{Object Counting} are \textit{Number of columns} ($r_{pb} = -0.360$), \textit{Number of objects} ($-0.337$), \textit{Number of layers} ($-0.310$), and \textit{Number of hidden objects} ($-0.228$); \textit{Fill ratio} is positive ($+0.177$). Qwen-Image-Edit shows the same directional profile (\textit{Number of objects} $-0.330$, \textit{Number of columns} $-0.329$, \textit{Number of layers} $-0.314$, \textit{Number of hidden objects} $-0.223$, \textit{Fill ratio} $+0.141$). HunyuanImage-Instruct shows the same direction for every factor at lower magnitudes. The camera factors remain weaker overall, but they are no longer negligible across the board: \textit{offset} is effectively null for all three image responders, whereas \textit{perspective} is mildly negative for Qwen-Image-Edit ($-0.040$) and more clearly negative for HunyuanImage-Instruct ($-0.088$). The same core structural axes that matter in text therefore also matter in image output: larger, taller, wider, and more occluded scenes are harder, while denser scenes are easier.

\subsection{Object-Category Results}
\label{app:object_results}

\begin{figure}[t]
\centering
\includegraphics[width=\textwidth]{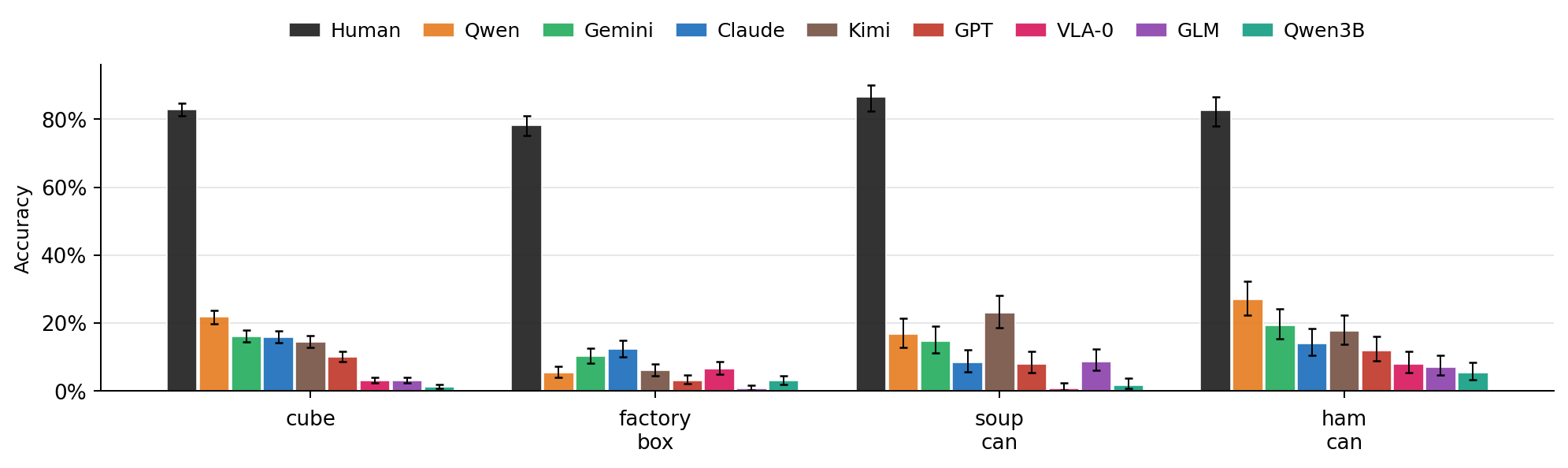}
\caption{\textit{Object Counting} accuracy by object type for the human baseline and all text models, with view-level Wilson 95\% confidence intervals. Object identity modulates accuracy but does not reverse the overall human-versus-model ordering. The human baseline remains above $78\%$ on every object, while no model exceeds $30\%$ on any object. The ``ham can'' label denotes the potted meat can.}
\label{fig:x1_accuracy_by_object}
\end{figure}

\paragraph{Main-task accuracy by object type.}
Figure~\ref{fig:x1_accuracy_by_object} reports responder-level \textit{Object Counting} accuracy broken down by the four object categories (textured cube, factory box, tomato soup can, potted meat can) for all text models and the human baseline, with Wilson 95\% confidence intervals. The human baseline remains strong across all four objects: $83.0\%$ on cube, $78.2\%$ on factory box, $86.7\%$ on soup can, and $82.7\%$ on potted meat can, for an object spread of $8.4$ percentage points. By contrast, pooled model accuracy across total object count drops from $31.5\%$ to $1.0\%$ (Table~\ref{tab:text_difficulty_bins}), so object identity is a secondary robustness axis rather than the dominant organizing variable. Some models do show object-specific brittleness: for instance, Qwen drops from $21.7\%$ on cubes to $5.4\%$ on factory boxes, and Kimi is unusually strong on soup can ($23.0\%$) relative to its other object categories. These object-specific dips are themselves diagnostic of which models are sensitive to object-level visual cues: factory-box scenes have distinctive textured rectangular surfaces that deviate from the canonical cuboid schema, and Qwen's drop is consistent with a surface-affordance effect, although the current data do not distinguish this from other object-specific failure modes.

\begin{figure}[t]
\centering
\includegraphics[width=\textwidth]{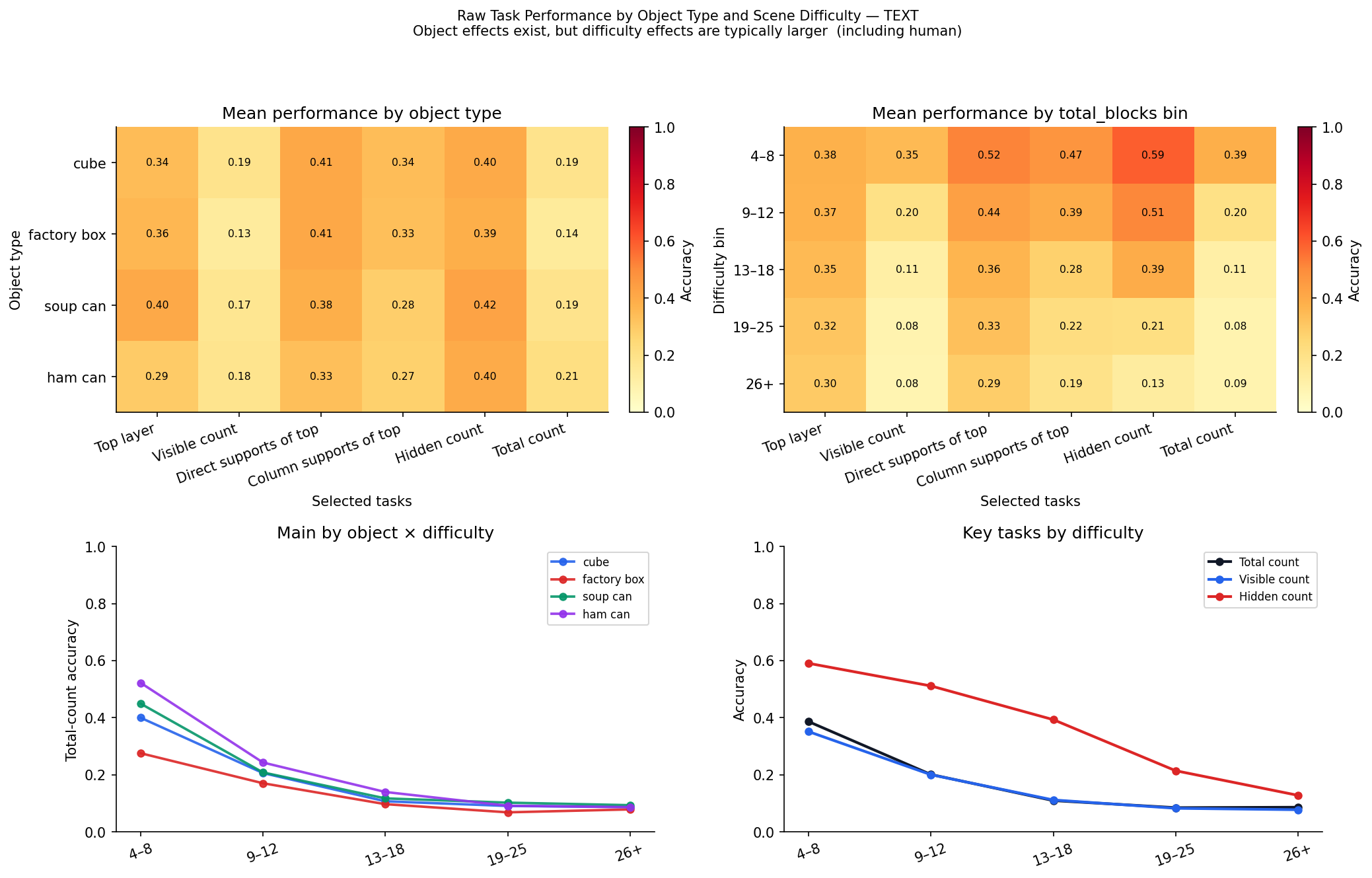}
\caption{Aggregated text-modality task performance by object type (left) and by \textit{Number of objects} difficulty bin (right), with paired \textit{Object Counting} curves at the bottom. Difficulty effects (right) are typically larger than object effects (left). The \textit{Object Counting} curves at bottom-left fall with difficulty in every object category, although their slopes differ (factory box $-19.7$ points, potted meat cans $-43.6$ points). The ``ham can'' label denotes the potted meat can.}
\label{fig:c12f_object_difficulty}
\end{figure}

\paragraph{Object-type effects versus difficulty effects.}
Figure~\ref{fig:c12f_object_difficulty} compares aggregated text task performance by object category alongside the same performance stratified by the \textit{Number of objects} difficulty bins. The two heatmaps show that object identity changes magnitude without reversing direction for most tasks: the per-object means on \textit{Visible Object Count}, \textit{Direct Support}, \textit{Support Column}, \textit{Hidden Object Count}, and \textit{Object Counting} differ by at most $\sim 8$ percentage points across object categories, and \textit{Top Layer} differs by $11.0$ points. The difficulty bins have a larger effect on most rows, which drop by $20$--$50$ percentage points from the easiest to hardest bin, although \textit{Top Layer} drops only $7.5$ points. The bottom-left line plot shows that the \textit{Object Counting} accuracy curve falls with difficulty for all four object categories, with different slopes: factory box drops $19.7$ points against $43.6$ points for potted meat cans. The bottom-right line plot also makes visible the interaction between \textit{Hidden Object Count} and the difficulty axis: hidden-count accuracy starts at $59\%$ for $4$--$8$-object scenes, falls to $51\%$ at $9$--$12$, and continues down through the hardest bins, demonstrating that the hidden-object branch responds to scene complexity in the same direction as the target task.

\paragraph{Object-type effects in the image modality.}
\textit{Factory box} is the hardest object for all three image responders (Gemini Flash Image $10.7\%$, Qwen-Image-Edit $2.0\%$, HunyuanImage-Instruct $2.6\%$), with the remaining ordering being mild and model-specific. The pooled object-difficulty fit gives a standardized total-block coefficient of $-0.106$ versus object offsets of $-0.066$ (factory box), $+0.035$ (soup can), and $+0.039$ (potted meat can) relative to cube, again confirming that the difficulty gradient is the larger axis.

\section{Full Hierarchy Dependency Analysis}
\label{app:hierarchy_dependency}

This appendix section supplements Section~\ref{sec:analysis} of the main paper with the complete statistical dependency evidence underlying all three evaluative summary tables, the per-cell mechanism breakdowns, edge-pass and within-scene lift statistics, and the additional analyses suppressed from the main text for space.

\paragraph{Inference protocol.}
Each model is queried once per (sample, task, modality) tuple, with a fixed prompt and decoding temperature appropriate to the modality. Closed frontier models are accessed through their respective API keys, and open-weights models are served locally via vLLM. We do not perform prompt ensembling for the main results: prompts were finalized during piloting and then frozen for the full sweep.

\paragraph{Confirmatory relations and their motivation.}
The pre-specified edges and bundles are those in Figure~\ref{fig:chains}. The Summation Mechanism is an arithmetic composition, and the Internal Referential Chain is deductive and built into the task wording. The Visible and Hidden Count Hierarchies are diagnostic hypotheses about how counting typically proceeds.

\paragraph{Statistical methodology.}
The Summation Mechanism analysis reports conditional \textit{Object Counting} accuracy in the four cells of the $\textit{Visible Object Count} \times \textit{Hidden Object Count}$ correctness joint, together with the prediction-level correlation $r(\hat{C}_{\mathrm{vis}} + \hat{C}_{\mathrm{hid}}, \hat{C}_{\mathrm{tot}})$ and the exact-match rate where the modality permits prediction-level access. The exact-match rate is the share of scenes on which the error of the reported total equals the sum of the errors of the reported visible and hidden counts, $(\hat{C}_{\mathrm{tot}} - C_{\mathrm{tot}}) - [(\hat{C}_{\mathrm{vis}} - C_{\mathrm{vis}}) + (\hat{C}_{\mathrm{hid}} - C_{\mathrm{hid}})] = 0$. Scoring errors rather than the raw identity accounts for scenes where the ground-truth counts themselves do not sum, since S9 counts visible objects resting on hidden ones ($17.9\%$ of benchmark views). The chain-and-supports analysis reports row-level relation integrity $I(P \rightarrow C) = P(P{=}1 \mid C{=}1)$ for each pre-specified sub-task edge, the perfect-chain rate for the Internal Referential Chain $\textit{Top Layer} \rightarrow \textit{Direct Support} \rightarrow \textit{Support Column}$, and within-scene lift $P(C{=}1 \mid P{=}1) - P(C{=}1 \mid P{=}0)$ on the subset of scenes where the parent variable varies across responders. Within-scene lift partially controls for shared scene difficulty by holding the scene fixed while varying the parent across responders. Significance uses a permutation procedure with Benjamini-Hochberg FDR correction across all within-scene comparisons in this section. Bootstrap confidence intervals resample rendered views while retaining their responder observations; they do not account for dependence among different views of the same structure.

\subsection{Text Modality: Full Hierarchy Dependency Evidence}
\label{app:text_hierarchy}
\label{app:text_full}
\label{app:text_mechanism}

\paragraph{Per-cell mechanism breakdown.}
Reading off the both-correct cell of Figure~\ref{fig:text_mechanism}, Qwen reaches $88.9\%$, Claude $66.9\%$, GPT $56.2\%$, and Gemini $48.0\%$, against $23.5\%$, $7.8\%$, and $1.9\%$ in the other three Qwen cells. The exact-match rate is $69.1\%$ for the human baseline, $64.2\%$ for Qwen, and $18.6\%$ for Gemini. Two models can reach similar headline accuracy by very different means: Qwen by approximating the human path through the explicit visible-plus-hidden composition, Gemini by some other route that succeeds at comparable rates without leaving the same component-conditioned signature.

\begin{figure}[t]
\centering
\includegraphics[width=\textwidth]{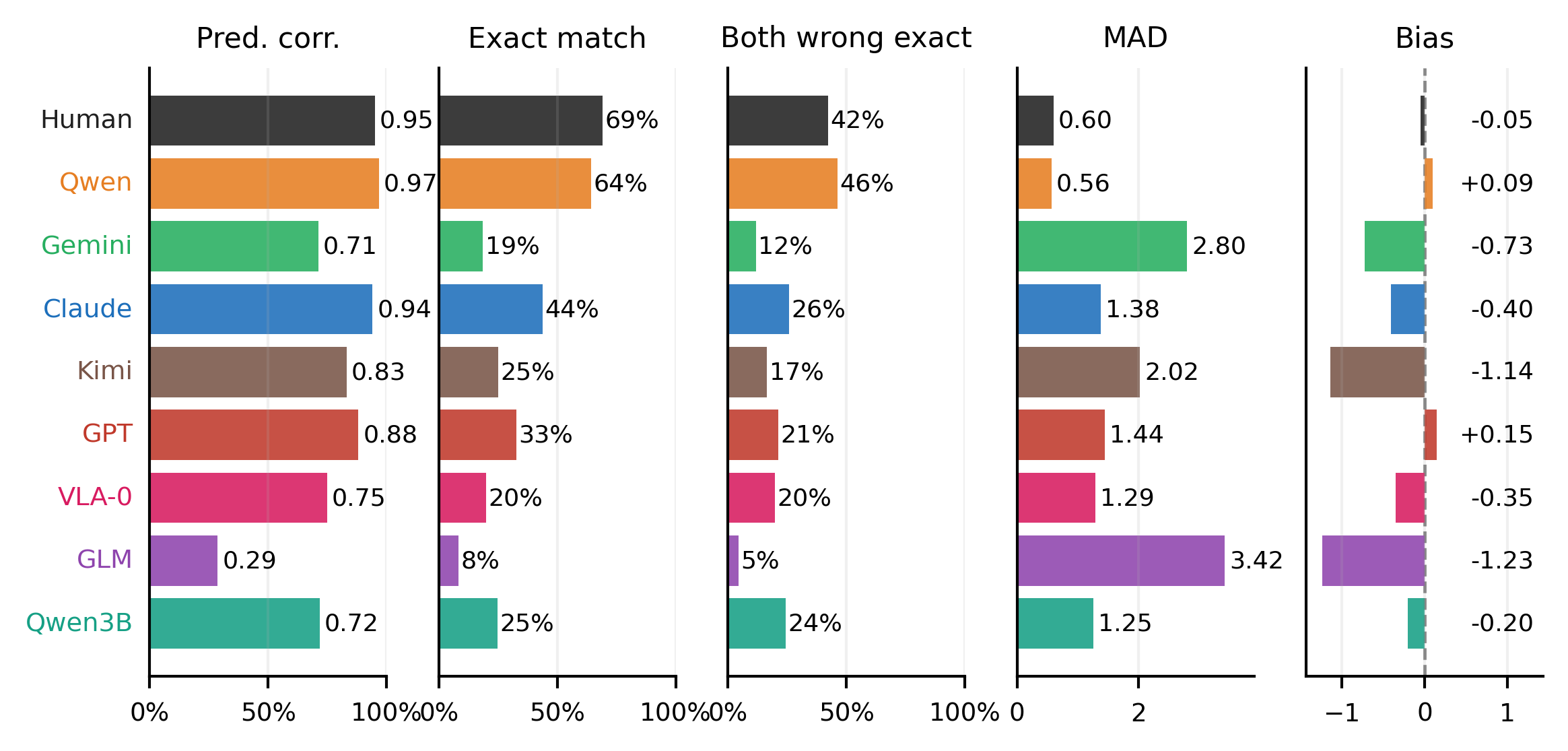}
\caption{Composition diagnostics for the Summation Mechanism analysis in text. Per responder: prediction correlation $r(\hat{C}_{\mathrm{vis}} + \hat{C}_{\mathrm{hid}}, \hat{C}_{\mathrm{tot}})$, \textit{Object Counting}-versus-component exact-match rate, the same rate restricted to the both-component-wrong subset, mean absolute deviation of the residual, and signed residual bias. Qwen leads the models on prediction correlation and exact match, but its high both-wrong-exact rate ($46\%$) indicates that compositional alignment can persist even with grounded errors. Gemini shows the opposite dissociation with high \textit{Object Counting} accuracy but low explicit composition alignment.}
\label{fig:text_e1aa_composition}
\end{figure}

\paragraph{Composition diagnostics.}
Figure~\ref{fig:text_e1aa_composition} reports the full set of composition diagnostics that complement the conditional cell view in Figure~\ref{fig:text_mechanism} of the main paper. Five quantities are reported per responder. The \textit{prediction correlation} is $r(\hat{C}_{\mathrm{vis}} + \hat{C}_{\mathrm{hid}}, \hat{C}_{\mathrm{tot}})$. The \textit{exact-match rate} is the fraction of scenes on which the error of the reported total equals the sum of the errors of the reported visible and hidden counts. The \textit{both-wrong-exact} rate is the same exact-match rate restricted to the subset of scenes where the model's component counts are themselves both wrong. The \textit{mean absolute deviation (MAD)} and \textit{Bias} are the mean absolute and signed values of the residual between the error of the reported total and the sum of the component errors, the latter reporting whether the model systematically over- or under-counts relative to its own component sum. The ordering by prediction correlation is Qwen ($0.97$) $>$ Human ($0.95$) $\approx$ Claude ($0.94$) $>$ GPT ($0.88$) $>$ Kimi ($0.83$) $>$ VLA-0 ($0.75$) $\approx$ Qwen3B ($0.72$) $\approx$ Gemini ($0.71$) $\gg$ GLM ($0.29$). The exact-match ordering is similar at the top (Human $69\%$, Qwen $64\%$, Claude $44\%$, GPT $33\%$). The double-failure exact-match rate is the stricter diagnostic: Qwen still reaches $46\%$ exact match when both component tasks are wrong, indicating that its high compositional alignment can persist even when the component counts themselves are inaccurate. Gemini's MAD ($2.80$) and bias ($-0.73$) are the largest among the proprietary models, consistent with its weaker exact-match rate and with a pattern of systematic under-counting relative to its own component sum.

\paragraph{Edge-wise integrity and within-scene lift for the Internal Referential Chain.}
We report per-edge pass rates, perfect-chain rates, and within-scene lift values for $\textit{Top Layer} \to \textit{Direct Support} \to \textit{Support Column}$. In the human baseline, edge-pass rates are $97.0\%$ for $\textit{Top Layer} \rightarrow \textit{Direct Support}$ and $83.6\%$ for $\textit{Direct Support} \rightarrow \textit{Support Column}$, with a perfect-chain rate of $82.8\%$. Among models, the highest perfect-chain rate is $49.8\%$ in Gemini, with Internal Referential Chain rates higher than Hidden Count Hierarchy rates in every model and than Visible Count Hierarchy rates in most. Within-scene lift for $\textit{Top Layer} \rightarrow \textit{Direct Support}$ is $+28.1$ percentage points and for $\textit{Direct Support} \rightarrow \textit{Support Column}$ is $+32.7$ percentage points, both significant under permutation tests with FDR correction.

\paragraph{Visible and Hidden Count Hierarchies.}
We report row-level integrity and within-scene lift for $\textit{Column Count} \land \textit{Layer Count} \to \textit{Visible Object Count}$ ($+56.0$ pp within-scene lift) and $\textit{Column Count} \land \textit{Support Column} \to \textit{Hidden Object Count}$ ($+49.7$ pp). Both relations survive FDR correction. Their relation integrity is lower than the chain's, although they rank higher on the scene-controlled dependence measures below. Their joint integrity, $P(\text{both prerequisites correct} \mid \text{dependent correct})$, is $70.6\%$ and $63.0\%$ in the human row (the edge-mean integrity in Table~\ref{tab:text_summary} is $84.5\%$ and $78.5\%$).

\paragraph{Bundle-to-Main coupling.}
We report within-scene correlations $r_{ws}$ between each of the four pre-specified bundles and \textit{Object Counting}. The Visible Count Hierarchy bundle has the highest aggregate coupling ($r_{ws} = 0.592$), the Summation Mechanism bundle sits at $r_{ws} = 0.545$, the Hidden Count Hierarchy bundle at $r_{ws} = 0.498$, and the Internal Referential Chain at $r_{ws} = 0.436$, with all four aggregate effects surviving BH FDR correction.

\paragraph{Scene-and-responder fixed effects.}
Within-scene lift and $r_{ws}$ hold the scene fixed but not the responder, so average performance differences across responders could inflate them. We therefore estimate a two-way model with scene and responder fixed effects, pooled across the eight models and the human baseline. The estimates shrink but stay positive, at $+20.2$ pp (95\% CI $[18.3, 22.3]$) for the Visible Count Hierarchy and $+15.7$ pp ($[13.4, 17.8]$) for the Hidden Count Hierarchy, with view-clustered bootstrap confidence intervals. Both stay positive without the human baseline ($+19.5$ and $+21.0$ pp), and every responder with usable variation shows a positive slope (7 of 7 and 6 of 6). The evidence remains associational: it is consistent with the hypothesized decomposition but does not establish that a model computes the count through these sub-tasks.

\paragraph{Object-conditioned hierarchy stability.}
We report object-stratified Internal Referential Chain edge-pass rates, within-scene lifts, and perfect-chain rates for the four object categories. The Internal Referential Chain remains positive in every object family, with soup can the strongest and textured cube the weakest in edge-pass and perfect-chain rates (within-scene lifts are strongest for potted meat cans), and no relation sign reverses across objects.

\paragraph{Action grounding and spatial reasoning: VLA-0 versus its backbone.}
VLA-0 is a vision-language-action model (which natively retains its ability to generate text) that uses Qwen2.5-VL-3B as its language backbone and is trained on robot trajectory data rather than on any task directly related to block counting or spatial reasoning. Even so, VLA-0 outperforms its own backbone on \textit{Object Counting} ($4.2\%$ versus $2.1\%$) and on the Hidden Count Hierarchy. The gain is small in absolute terms, and the backbone is higher on the Internal Referential Chain and the Summation Mechanism, but action-grounded training on tasks unrelated to counting may transfer positively to the spatial reasoning the benchmark probes. We are cautious about drawing strong conclusions from a single backbone-versus-VLA pair, particularly at this performance regime, but the directional finding is worth flagging because it bears on a longstanding hypothesis about how spatial intelligence develops. Human infants acquire much of their early spatial competence not through linguistic supervision but through embodied interaction with the physical world, manipulating and reorienting objects in three dimensions. The VLA-0 result, however small, is consistent with the same effect appearing in models: that grounding learning in physical interaction may benefit spatial reasoning even on tasks the model was never trained to perform. Confirming this would require comparing more (backbone, VLA) pairs and more spatial benchmarks, which we leave to future work.

\subsection{MCQ Modality: Full Hierarchy Dependency Evidence}
\label{app:mcq_hierarchy}
\label{app:mcq_full}

\begin{figure}[t]
\centering
\includegraphics[width=\textwidth]{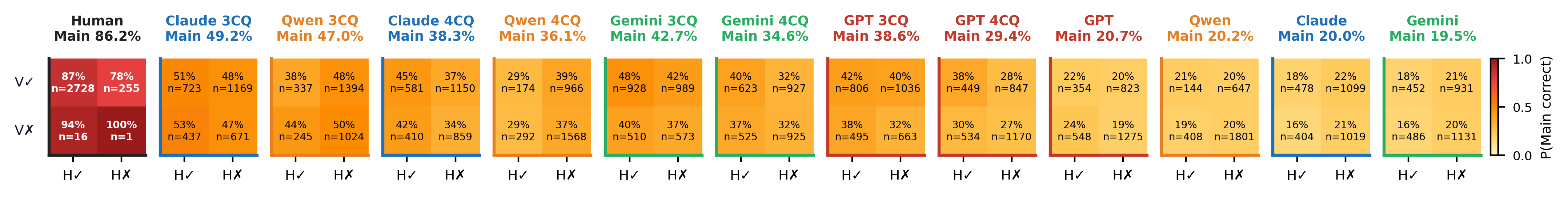}
\caption{Conditional \textit{Object Counting} accuracy under the $\textit{Visible Object Count} \times \textit{Hidden Object Count} \rightarrow \textit{Object Counting}$ mechanism in MCQ. Each panel is a single responder's $2 \times 2$ joint, ordered left-to-right by chance-adjusted \textit{Object Counting} accuracy. Each cell reports both the conditional accuracy and the cell sample size. Human five-choice (leftmost) concentrates almost all scenes in the both-correct cell ($87\%$, $n = 2728$), and the off-diagonal and both-wrong cells are nearly empty. Model rows under five-choice (rightmost four) collapse the structure, and three- and four-choice rows partially recover it for Claude, Gemini, and GPT.}
\label{fig:mcq_e1a_mechanism}
\end{figure}

\paragraph{Per-cell mechanism breakdown across answer-space conditions.}
Figure~\ref{fig:mcq_e1a_mechanism} reports the $\textit{Visible Object Count} \times \textit{Hidden Object Count}$ joint at five-, four-, and three-choice for each of the four overlapping models, alongside the human five-choice anchor. The human baseline preserves the expected mechanism: \textit{Object Counting} reaches $87\%$ when both component sub-tasks are correct ($n = 2728$), and $78\%$ when only \textit{Visible Object Count} is correct, with the off-diagonal cells correspondingly small in scene share. Under five-choice MCQ, every model row collapses with the both-correct cell and the both-wrong cell within three percentage points of each other (five-choice both-correct vs both-wrong: Claude $18\%$ vs $21\%$, Qwen $21\%$ vs $20\%$, Gemini $18\%$ vs $20\%$, GPT $22\%$ vs $19\%$). Reducing the answer space partially restores the expected mechanism. The both-correct cell rises monotonically from five-choice through three-choice for Claude ($18\%, 45\%, 51\%$) and Gemini ($18\%, 40\%, 48\%$), while the both-wrong cell rises more slowly, and GPT shows the same pattern more weakly ($22\%, 38\%, 42\%$). Qwen does not: its \textit{Object Counting} accuracy improves with answer-space reduction but the both-correct cell ($21\%, 29\%, 38\%$) remains close to or below the both-wrong cell, and at three-choice the off-diagonal cells (e.g., V$\checkmark$H$\times$ at $48\%$) actually exceed the both-correct cell ($38\%$). Better final-answer selection under a smaller answer space does not in itself imply that the model is recovering the explicit visible-plus-hidden decomposition.

\paragraph{Edge-wise integrity under answer-space variation.}
We report per-edge pass rates and within-scene lifts for the Internal Referential Chain and the Visible and Hidden Count Hierarchies under five-, four-, and three-choice conditions. The Internal Referential Chain integrity is generally well-preserved across responders, with the human baseline reaching above $90\%$ on each edge under five-choice. Model edge-pass rates and perfect-chain rates are lower in absolute terms but largely follow the same ordering as in text: the Internal Referential Chain is cleaner than the Hidden Count Hierarchy in every reported MCQ condition and than the Visible Count Hierarchy except for Gemini. The chain edges are significantly positive after FDR correction in the pooled analysis. At three and four choices, only the $\textit{Direct Support} \rightarrow \textit{Support Column}$ edge remains significant.

\paragraph{Bundle-to-Main coupling under MCQ.}
We report pooled within-scene $r_{ws}$ values for each bundle at each answer-space size. MCQ correlations are one-third to one-half the magnitude of text correlations. Five-choice MCQ compresses differences in \textit{Object Counting} accuracy across models, but the present analysis does not establish the cause of this attenuation. Pooled values are $r_{ws} = 0.211$ for the Internal Referential Chain bundle, $0.196$ for the Visible Count Hierarchy bundle, $0.192$ for the Hidden Count Hierarchy bundle, and $0.163$ for the Summation Mechanism bundle. The human five-choice rows contribute much of the pooled value ($0.04$--$0.07$ without them).

\paragraph{Qwen \textit{Hidden Object Count} below chance.}
\label{app:mcq_qwen_t8}
Qwen's \textit{Hidden Object Count} accuracy is below chance in every MCQ condition (chance-adjusted $-0.209$, $-0.126$ and $-0.020$ at three, four and five choices) because its wrong answers concentrate on two distractors (alt-incomplete-subset $22.9\%$ and wrong-visible-block $20.7\%$ of trials at three choices), not because hidden-object reasoning reverses.

\subsection{Image Modality: Full Hierarchy Dependency Evidence}
\label{app:image_hierarchy}
\label{app:image_full}

\paragraph{\textit{Hidden Object Count} in image output.}
In every modality, \textit{Hidden Object Count} targets the visible objects resting directly on a hidden object.

\paragraph{Task accuracy, difficulty controls, and object robustness.}
We provide the per-task accuracy figure (Figure~\ref{fig:ig_c1_task_accuracy} above) and the five-axis difficulty breakdown (Figure~\ref{fig:ig_c2_difficulty} above) for the three image-output models, paralleling the text figures. We additionally provide the object-conditioned \textit{Object Counting} accuracy for all three image responders, showing that factory box is the hardest object for all three and that the difficulty gradient remains the dominant axis.

\begin{figure}[t]
\centering
\includegraphics[width=0.85\textwidth]{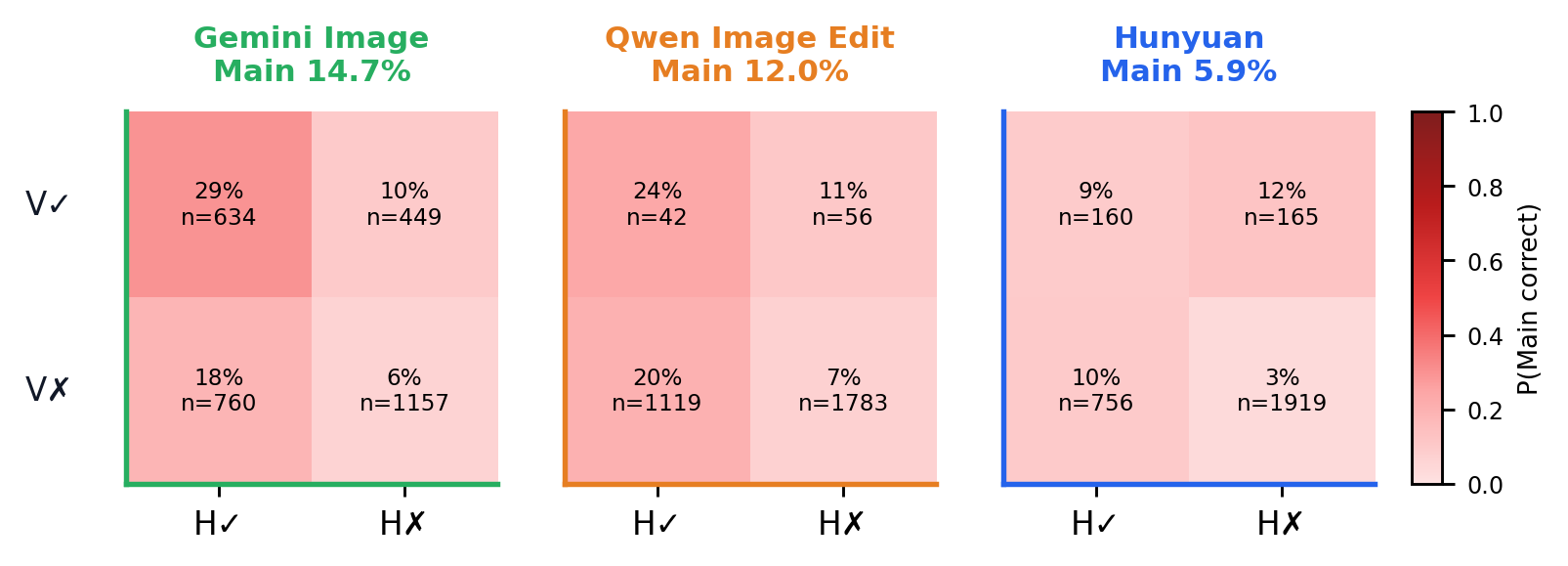}
\caption{Conditional \textit{Object Counting} accuracy under the $\textit{Visible Object Count} \times \textit{Hidden Object Count} \rightarrow \textit{Object Counting}$ mechanism in image output for the three image-editing models. Headline \textit{Object Counting} accuracy appears in the panel title; each cell reports the conditional accuracy and sample size. Gemini Flash Image and Qwen-Image-Edit preserve the expected ordering with both-correct $>$ both-wrong. HunyuanImage-Instruct also preserves both-correct $>$ both-wrong, but its visible-only cell ($12\%$) exceeds both-correct ($9\%$).}
\label{fig:ig_e1a_mechanism}
\end{figure}

\paragraph{Summation Mechanism in image output.}
Figure~\ref{fig:ig_e1a_mechanism} reports the $\textit{Visible Object Count} \times \textit{Hidden Object Count}$ conditional accuracy grid for Gemini Flash Image, Qwen-Image-Edit, and HunyuanImage-Instruct, paralleling Figure~\ref{fig:text_mechanism} of the main paper. The joint preserves the expected ordering for two of the three responders, but at substantially weaker magnitudes than in text. Gemini Flash Image reaches $29\%$ \textit{Object Counting} accuracy when both sub-tasks are correct ($n = 634$), dropping to $18\%$ when only \textit{Hidden Object Count} is correct, $10\%$ when only \textit{Visible Object Count} is correct, and $6\%$ when both are wrong. Qwen-Image-Edit shows the same endpoint ordering at slightly lower magnitudes ($24\%$ in the both-correct cell with $n = 42$, $7\%$ in the both-wrong cell), with the mixed cells reversed relative to text ($20\%$ when only \textit{Hidden Object Count} is correct and $11\%$ when only \textit{Visible Object Count} is correct), as for Gemini Flash Image. The very small $n = 42$ for Qwen-Image-Edit's both-correct cell reflects that its visible-count branch almost entirely collapses ($3.3\%$ raw \textit{Visible Object Count} accuracy), which limits the resolution of this cell. HunyuanImage-Instruct does not preserve a clean mechanism pattern; its both-correct cell is only $9\%$, its visible-only cell is slightly higher at $12\%$, its hidden-only cell sits at $10\%$, and its both-wrong cell at $3\%$. The image-generation mechanism is therefore real but weak, and it is less stable across responders than in text.

\paragraph{Internal Referential Chain and Count Hierarchies in image output.}
We report edge-wise integrity and within-scene lift for the Internal Referential Chain and the two Count Hierarchies across the three image responders. The Internal Referential Chain $\textit{Top Layer} \rightarrow \textit{Direct Support} \rightarrow \textit{Support Column}$ is visible in every responder, with edge-pass rates of $67.6\%$ for Qwen-Image-Edit, $53.7\%$ for Gemini Flash Image, and $50.1\%$ for HunyuanImage-Instruct, and corresponding perfect-chain rates of $48.5\%$, $35.9\%$, and $34.3\%$. Pooled within-scene lifts on the six pre-target relations are uniformly positive and FDR-significant: $\textit{Top Layer} \rightarrow \textit{Direct Support}$ at $+19.1$ pp, $\textit{Direct Support} \rightarrow \textit{Support Column}$ at $+38.7$, the two visible-side supports at $+40.2$ and $+43.7$, and the two hidden-side supports at $+18.9$ and $+30.5$. The Count Hierarchies follow the same ordering observed in text and MCQ, with the Internal Referential Chain consistently cleaner: in joint integrity, the Count Hierarchies sit at $20.8\%$ (Visible) and $18.1\%$ (Hidden) for Gemini Flash Image against an Internal Referential Chain edge-pass rate of $53.7\%$. Qwen-Image-Edit shows the largest split, with a $67.6\%$ Internal Referential Chain against a joint $42.9\%$ Visible Count Hierarchy and a near-absent $2.2\%$ Hidden Count Hierarchy, and HunyuanImage-Instruct splits $50.1\%$ Internal Referential Chain against joint $27.1\%$ Visible Count Hierarchy and $9.5\%$ Hidden Count Hierarchy. By contrast, the within-scene lifts on the target-adjacent edges are much smaller, at $+5.6$ pp for $\textit{Visible Object Count} \rightarrow \textit{Object Counting}$ and $+6.5$ pp for $\textit{Hidden Object Count} \rightarrow \textit{Object Counting}$, indicating that the upstream structure is more format-stable than the final integration step.

\paragraph{Bundle-to-\textit{Object Counting} coupling in image output.}
We provide bundle-to-\textit{Object Counting} coupling for the three image responders. Gemini Flash Image leads on \textit{Object Counting} ($0.147$) and on the conditional mechanism proxy ($0.295$), and its bundle-to-\textit{Object Counting} correlations are uniformly the highest in the modality, at $0.210$ for the Internal Referential Chain, $0.181$ for the Visible Count Hierarchy, $0.240$ for the Hidden Count Hierarchy, and $0.199$ for the Summation Mechanism bundle. Qwen-Image-Edit gives the clearest example of the local-versus-target dissociation: it has the strongest Internal Referential Chain ($0.676$) and the strongest edge-mean Visible Count Hierarchy integrity ($0.587$), but its raw \textit{Visible Object Count} accuracy is only $3.3\%$ and its visible-bundle-to-\textit{Object Counting} correlation is slightly negative ($-0.020$). Strong upstream structure does not, on its own, produce strong target-task coupling. HunyuanImage-Instruct completes the picture at the weak-coupling end: although its Internal Referential Chain remains positive ($0.501$), its bundle-to-\textit{Object Counting} correlations are uniformly small ($0.024$ to $0.119$).

\section{Cross-Modality Analysis}
\label{app:cross_modality}

\paragraph{Format differences.}
All three response formats ask for the same \textit{Hidden Object Count} quantity, the visible objects resting directly on a hidden object. The formats differ in how the answer is produced: a free-response integer, a selection among recolored images, or an edited image. The answer-selection burden in MCQ introduces the additional source of variance documented in Section~\ref{sec:validation_spread}.

\paragraph{Cross-modality effect size differences.}
We report quantitative comparisons of within-scene lifts and bundle-to-\textit{Object Counting} correlations across text, MCQ, and image output. MCQ within-scene lifts ($+1.4$ and $+4.3$ pp for the Internal Referential Chain at three-choice) are roughly eight to twenty times smaller than in text ($+28$ to $+33$ pp), and individual-responder bundle correlations are near zero in five-choice MCQ compared to $-0.09$ to $+0.65$ in text. The present analysis does not distinguish answer-selection effects from differences in the elicited dependency structure.

\paragraph{Format-stable versus format-sensitive findings.}
We summarize which conclusions replicate directionally across all three response formats and which are format-dependent. The Internal Referential Chain is the most robust hierarchy backbone across all modalities. Human performance respects the intended hierarchy ordering in both formats where it was measured (text and five-choice MCQ). The upstream local hierarchy is more format-stable than the target-task-proximal Summation Mechanism. By contrast, the visible/hidden branch asymmetry reverses between text and image output, and target-task-proximal bundle coupling appears weakest in five-choice MCQ.

\section{Training Details and Extended Results}
\label{app:training}

This appendix section supplements Section~\ref{sec:training} with the full setup, hyperparameters, sample-based decoding metrics, and extended analyses of object generalization and policy concentration behavior under the three training conditions and the zero-shot baseline.

\subsection{Setup}
\label{app:training_setup}

\paragraph{Models and data split.}
We train at two scales, Qwen2.5-VL-7B-Instruct and Qwen2.5-VL-32B-Instruct, both initialized from the public HuggingFace checkpoints. A pool of $136{,}171$ views of $8{,}521$ structures is split 90/10 by structure into a training set and a held-out validation set of $13{,}618$ views, so no view of any structure appears in both. Each split inherits the camera and environment diversity described in Appendix~\ref{app:dataset_details}. The SFT conditions train on the full training set for one epoch, and DAPO-tight continues from the SFT-CoT checkpoint on a $50\%$ subset of it. A replication on Gemma-4-31B~\citep{google2026gemma4} keeps this protocol and split unchanged.

\paragraph{Evaluation recipe.}
In Table~\ref{tab:training_summary}, the zero-shot and SFT-plain rows answer each task in its own query, as in Table~\ref{tab:text_summary}. The SFT-CoT and DAPO-tight rows answer once per benchmark view, and each sub-task score is read from the matching numbered line of that chain-of-thought response. \textit{Object Counting} is therefore comparable across all rows, while the sub-task, hierarchy and \textit{Summation Mechanism} columns of the SFT-CoT and DAPO-tight rows are not independently elicited. Validation-split results use $k{=}10$ completions per view at temperature $1.0$, top-$p$ $1.0$, and a maximum of $1024$ new tokens. On that split we report acc@1 (the accuracy of the first sampled completion), acc@10\_maj (the accuracy of the majority vote across the ten completions), and pass@10 (the fraction of views for which at least one of the ten completions is correct).

\paragraph{Chain-of-thought trace structure (Phase 1).}
For each scene in the SFT-CoT condition, the training target lists the answer to each sub-task in the order required to compute the target-task answer additively, followed by the final integer wrapped in answer tags. The model first reports the number of columns, layers, and clusters, then enumerates the visible objects, then derives the hidden-object count from the top-layer count, the directly supporting count, and the above-hidden count, and finally sums visible plus hidden to produce the total. The trace additionally supervises the total hidden-object count, which is distinct from the S9 proxy and supplies the hidden operand in the final addition. Table~\ref{tab:training_summary} and Figures~\ref{fig:tr_e1a_mechanism}--\ref{fig:tr_e1aa_composition} instead score the S9 line as \textit{Hidden Object Count}. All sub-task values are populated from ground truth, and we train with cross-entropy on the assistant tokens only. The SFT-plain condition uses the same scenes, but the assistant target consists of the final integer alone with no preceding sub-task trace.

\subsection{Training Hyperparameters}
\label{app:training_hparams}

\paragraph{SFT training configuration.}
Both scales train for one epoch with a maximum sequence length of $2048$ and cross-entropy on the assistant tokens only. Qwen2.5-VL-7B-Instruct uses a learning rate of $2\times10^{-5}$ and a global batch size of $512$, and Qwen2.5-VL-32B-Instruct uses $1\times10^{-5}$ and $256$. Each job runs on $8$ nodes ($32$ GPUs) of an NVIDIA GB200 cluster.

\paragraph{RLVR reward function and training configuration.}
The reward at each rollout is a weighted sum of a format component (weight $0.1$) checking for the presence of the structured answer tags and an integer-distance component (weight $0.9$) returning $\max(0, 1 - |\hat{y} - y| / \max(|y|, 5))$, where $\hat{y}$ is the model's predicted integer and $y$ is the ground-truth count. The integer-distance component returns zero whenever the structured answer tags are missing from the model output. The format gate prevents untagged answers from receiving the integer-distance reward; it does not itself require a reasoning trace. We optimize with GRPO under the DAPO-tight recipe, with a KL anchor on the SFT-CoT initialization (penalty coefficient $0.05$) to keep the policy on the fine-tuning manifold during exploration. Each step draws $8$ prompts with $8$ rollouts each at temperature $1.3$, and dynamic sampling drops groups with no reward variance. We use a learning rate of $1\times10^{-7}$, clip ratios of $0.2$ and $0.28$, and up to $500$ steps, and we select the best checkpoint by reward on $64$ validation samples.

\subsection{Extended Training Results}
\label{app:training_results}

\begin{figure}[t]
\centering
\includegraphics[width=\textwidth]{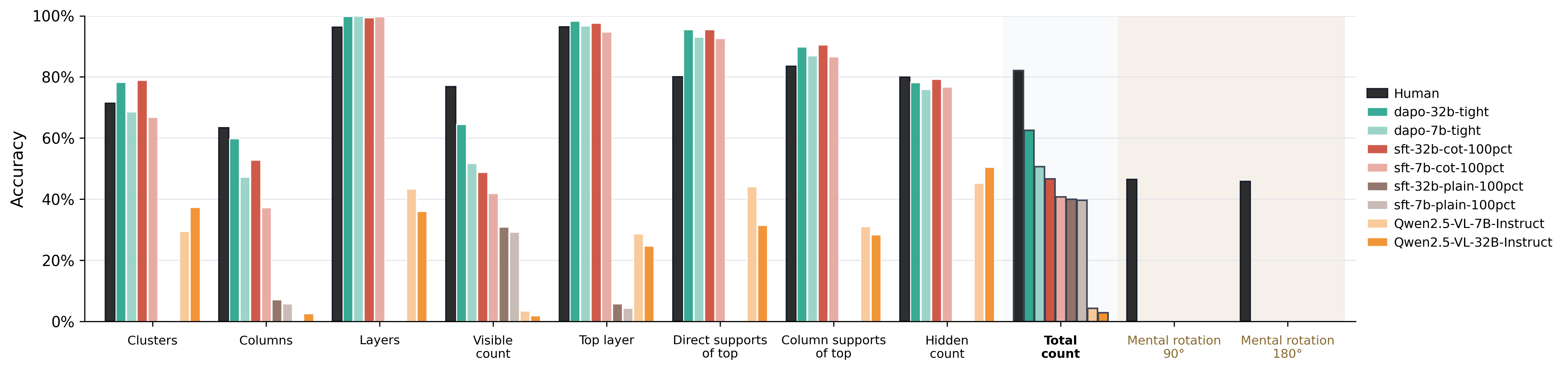}
\caption{Per-task accuracy across the trained-text intervention subset, redrawn in the same task-order format as Figure~\ref{fig:task_accuracy_text} of the main paper. Family hues identify DAPO-tight, SFT-CoT, SFT-plain, and Qwen2.5-VL backbones; within each family, $32$B keeps the base color and $7$B uses a $50\%$ shade. SFT-plain raises \textit{Object Counting} into the $40\%$ range while collapsing every sub-task involving spatial structure; SFT-CoT and DAPO-tight raise \textit{Object Counting} and the sub-tasks together.}
\label{fig:tr_c1_task_accuracy}
\end{figure}

\paragraph{Per-task accuracy under training interventions.}
Figure~\ref{fig:tr_c1_task_accuracy} reports per-task accuracy for the trained checkpoints alongside the human baseline and the two zero-shot Qwen2.5-VL backbones. The training intervention is large enough to change the benchmark ranking regime. Human accuracy remains the ceiling at $82.1\%$ on \textit{Object Counting}, but the best trained checkpoints close much more of the gap than any zero-shot model: DAPO-tight 32B reaches $62.6\%$, DAPO-tight 7B $50.7\%$, SFT-CoT 32B $46.7\%$, and SFT-CoT 7B $40.8\%$, all far above the strongest zero-shot text responder (Qwen at $17.7\%$, Table~\ref{tab:text_task_accuracy_full}). These gains are not confined to the final answer. The same checkpoints, whose sub-task scores are read from one chain-of-thought response, raise \textit{Visible Object Count} into the $42$--$65\%$ range and \textit{Hidden Object Count} into the $76$--$79\%$ range, and recover \textit{Direct Support} and \textit{Support Column} into the $87$--$96\%$ range, placing them much closer to the human profile than to the zero-shot regime. The SFT-plain checkpoints provide a contrasting pattern. SFT-plain 32B and SFT-plain 7B also raise \textit{Object Counting} (to $40.1\%$ and $39.7\%$), but they do so with much weaker component-task structure: \textit{Visible Object Count} remains only $30.9\%$ and $29.2\%$, while \textit{Hidden Object Count}, \textit{Cluster Count}, \textit{Layer Count}, \textit{Direct Support}, and \textit{Support Column} all collapse to near zero. The benchmark therefore remains diagnostic under training: two checkpoints can land in a similar \textit{Object Counting} range while exhibiting very different component profiles. The trained checkpoints are evaluated only in the text modality and therefore do not produce \textit{Mental Rotation} responses in the figure (the rightmost shaded region is empty for all trained rows).

\paragraph{Reinforcement learning concentrates the policy rather than expanding the candidate set.}
On the validation split, both the 7B and 32B SFT-CoT checkpoints already have pass@10 in the high eighties, meaning the correct answer is in the ten-sample budget on roughly nine out of ten views. Their acc@1 is much lower, between $29\%$ and $36\%$, so the correct answer is often present but not picked first. DAPO-tight narrows this gap: pass@10 moves only marginally ($+2.0$ at 7B, $+1.1$ at 32B), while acc@1 moves by $8$ to $18$ absolute points. This pattern fits a policy gradient that concentrates probability mass on trajectories already present in the SFT-CoT distribution. The pattern describes the output distribution and is not evidence about how the model computes its answer.

\paragraph{Controlled cold-start ablation.}
\label{app:cold_start}
Comparing SFT-plain with DAPO-tight changes the label type and the training algorithm at the same time. Table~\ref{tab:cold_start} varies one factor at a time on Qwen2.5-VL-7B, with the DAPO-tight recipe, the validation split and the evaluation protocol held fixed. The cold start uses either the chain-of-thought trace or the bare answer in the same tags, so the reward and parser are identical across arms, and the two arms of each data split use the same images. After identical RL, the chain-of-thought cold start leads by $9.4$, $24.2$ and $4.4$ points at the three splits. The SFT-only gaps are $+5.9$, $+19.6$ and $-0.6$ points, so at $100\%$ of the data the advantage of chain-of-thought labels appears only after RL. RL adds $8.0$ to $9.0$ points after a chain-of-thought cold start and $3.0$ to $4.9$ after a plain one. The $10/90$ split, which gives RL the most data, reaches $29.0\%$, below the $50/50$ ($39.6\%$) and $100/50$ ($37.2\%$) splits.

\begin{table}[t]
\centering
\small
\caption{Controlled cold-start ablation on Qwen2.5-VL-7B. Accuracies are validation-split acc@1 in percent, and RL gains are computed before rounding. The $10\%$ cold starts train for three epochs because one epoch did not teach the tag format. Evaluation coverage is $96$--$99.5\%$ of the $13{,}618$ validation views. The plain arms keep the answer tags, so the $100\%$ plain row differs from the untagged SFT-plain row of Table~\ref{tab:training_validation} ($28.63\%$).}
\label{tab:cold_start}
\begin{tabular}{llcccc}
\toprule
Cold start & Labels & RL data & SFT acc@1 & DAPO-tight acc@1 & RL gain \\
\midrule
$10\%$ (3 epochs) & CoT & $90\%$ & 20.6 & 29.0 & $+8.5$ \\
 & plain & $90\%$ & 14.7 & 19.6 & $+4.9$ \\
\midrule
$50\%$ & CoT & $50\%$ & 30.6 & 39.6 & $+9.0$ \\
 & plain & $50\%$ & 11.0 & 15.4 & $+4.4$ \\
\midrule
$100\%$ & CoT & $50\%$ & 29.2 & 37.2 & $+8.0$ \\
 & plain & $50\%$ & 29.8 & 32.8 & $+3.0$ \\
\bottomrule
\end{tabular}
\end{table}

\begin{figure}[t]
\centering
\includegraphics[width=\textwidth]{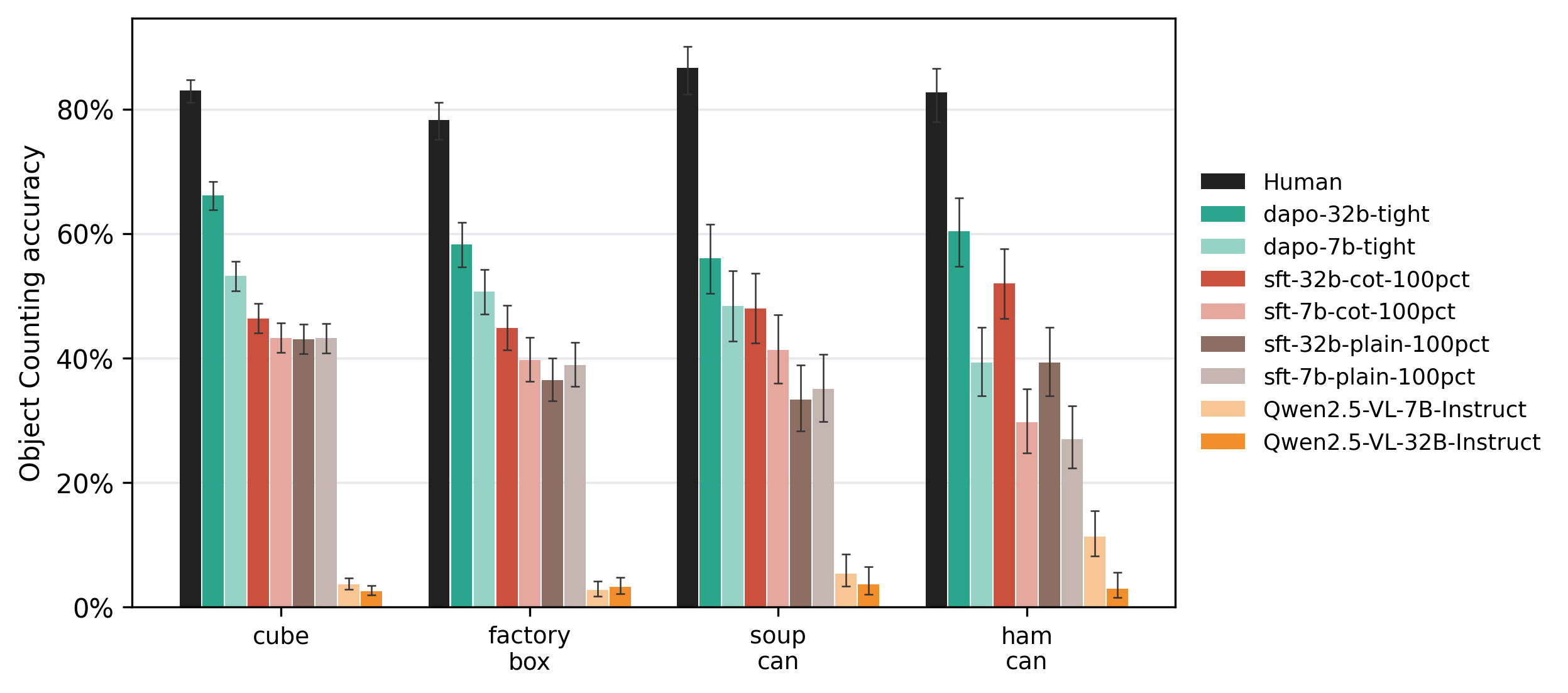}
\caption{\textit{Object Counting} accuracy by object type for the trained checkpoints (DAPO-tight and SFT-CoT/SFT-plain at 7B and 32B), the two zero-shot Qwen2.5-VL backbones, and the human baseline, with view-level 95\% Wilson confidence intervals. Cube is the only object category seen during training, and factory box, soup can, and potted meat can are out-of-distribution. The ``ham can'' label denotes the potted meat can.}
\label{fig:training_object_generalization}
\end{figure}

\paragraph{Object generalization for trained checkpoints.}
The training data contains only one object category, the textured cube. The benchmark itself spans four categories (cube, factory box, soup can, and potted meat can), three of which are absent from the training distribution. Figure~\ref{fig:training_object_generalization} reports DAPO-tight 32B's \textit{Object Counting} accuracy broken down by object category. In-distribution cube accuracy is $66\%$, and on the three out-of-distribution categories the accuracies are $58\%$ (factory box), $56\%$ (soup can), and $60\%$ (potted meat can). The model retains the bulk of its in-distribution performance on object categories it never saw during training. Qwen2.5-VL-32B (zero-shot) trails DAPO-tight 32B by $52$ points or more on every category, so the gap between trained and zero-shot is larger than the gap between in-distribution and out-of-distribution performance.

\paragraph{Sample-based decoding behavior and selection-based test-time scaling.}
Table~\ref{tab:training_validation} reports acc@1, acc@10\_maj and pass@10 on the validation split. On the trained Qwen2.5-VL checkpoints, majority vote adds only about $5$ to $8$ absolute points over acc@1, substantially less than the gap to pass@10. A small inspection of validation-split outputs found most errors off by one or two on the hidden-object step, whereas in the benchmark predictions most errors ($79$--$89\%$) lie in the visible step. Errors are systematic rather than stochastic, so naive test-time scaling that re-samples without selection inherits the same bias. Selection-based test-time scaling, where a verifier or reward model re-ranks the $k$ rollouts, is therefore a more promising avenue for further gains than simply increasing $k$.

\begin{table}[t]
\centering
\small
\caption{Validation-split results in percent ($k{=}10$). Each row is computed on the $n$ validation views (of $13{,}618$) its evaluation completed; unparseable responses are scored as incorrect. The Qwen2.5-VL-7B zero-shot row uses the chain-of-thought prompt.}
\label{tab:training_validation}
\begin{tabular}{llcccr}
\toprule
Model & Condition & acc@1 & acc@10\_maj & pass@10 & $n$ \\
\midrule
Qwen2.5-VL-7B & Zero-shot & 5.31 & 4.44 & 31.39 & 13{,}438 \\
 & SFT-plain & 28.63 & 33.14 & 84.29 & 13{,}536 \\
 & SFT-CoT & 29.24 & 35.80 & 86.87 & 12{,}867 \\
 & DAPO-tight & 37.21 & 45.10 & 88.84 & 13{,}276 \\
\midrule
Qwen2.5-VL-32B & Zero-shot & 4.95 & 4.66 & 17.48 & 13{,}542 \\
 & SFT-plain & 25.00 & 30.09 & 80.66 & 13{,}618 \\
 & SFT-CoT & 36.45 & 43.18 & 88.79 & 13{,}608 \\
 & DAPO-tight & 54.28 & 58.79 & 89.86 & 12{,}979 \\
\midrule
Gemma-4-31B & Zero-shot & 7.82 & 7.50 & 15.12 & 13{,}562 \\
 & SFT-plain & 22.59 & 25.52 & 74.10 & 13{,}618 \\
 & SFT-CoT & 50.37 & 58.08 & 92.92 & 13{,}601 \\
 & DAPO-tight & 60.38 & 62.81 & 85.45 & 13{,}002 \\
\bottomrule
\end{tabular}
\end{table}

\paragraph{Replication on Gemma-4-31B.}
The Gemma-4-31B rows of Table~\ref{tab:training_validation} follow the same ordering as the Qwen2.5-VL rows on acc@1. SFT-CoT gains $+27.8$ points over SFT-plain, against $+11.5$ at Qwen2.5-VL-32B and $+0.6$ at Qwen2.5-VL-7B, and DAPO-tight adds a further $+10.0$ points. Unlike in the Qwen2.5-VL rows, pass@10 falls after RL ($92.9\%$ to $85.5\%$). The acc@1 gaps above are validation-split values and cannot be compared with the benchmark values in Table~\ref{tab:training_summary}.

\paragraph{32B evaluation coverage note.}
The 32B DAPO-tight validation-split evaluation timed out with ranks about $96\%$ complete, leaving $12{,}979$ of $13{,}618$ views ($95.3\%$). Results are computed on the completed evaluation views; selection bias from incomplete coverage cannot be excluded. Table~\ref{tab:training_summary} is not affected and covers all $3{,}000$ benchmark views.

\begin{figure}[t]
\centering
\includegraphics[width=\textwidth]{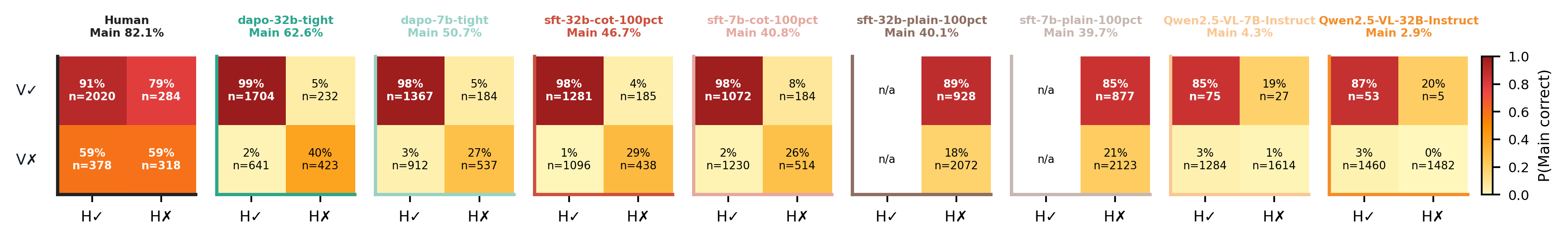}
\caption{Conditional \textit{Object Counting} accuracy under the $\textit{Visible Object Count} \times \textit{Hidden Object Count} \rightarrow \textit{Object Counting}$ mechanism for the trained-text intervention subset, in the same one-row $2 \times 2$ format as Figure~\ref{fig:text_mechanism} of the main paper. DAPO-tight and SFT-CoT checkpoints reach $98$--$99\%$ \textit{Object Counting} accuracy in the both-correct cell. SFT-plain checkpoints have empty both-correct and hidden-only cells (their \textit{Hidden Object Count} answers are never correct), and the surviving cells form a one-column slice.}
\label{fig:tr_e1a_mechanism}
\end{figure}

\begin{figure}[t]
\centering
\includegraphics[width=\textwidth]{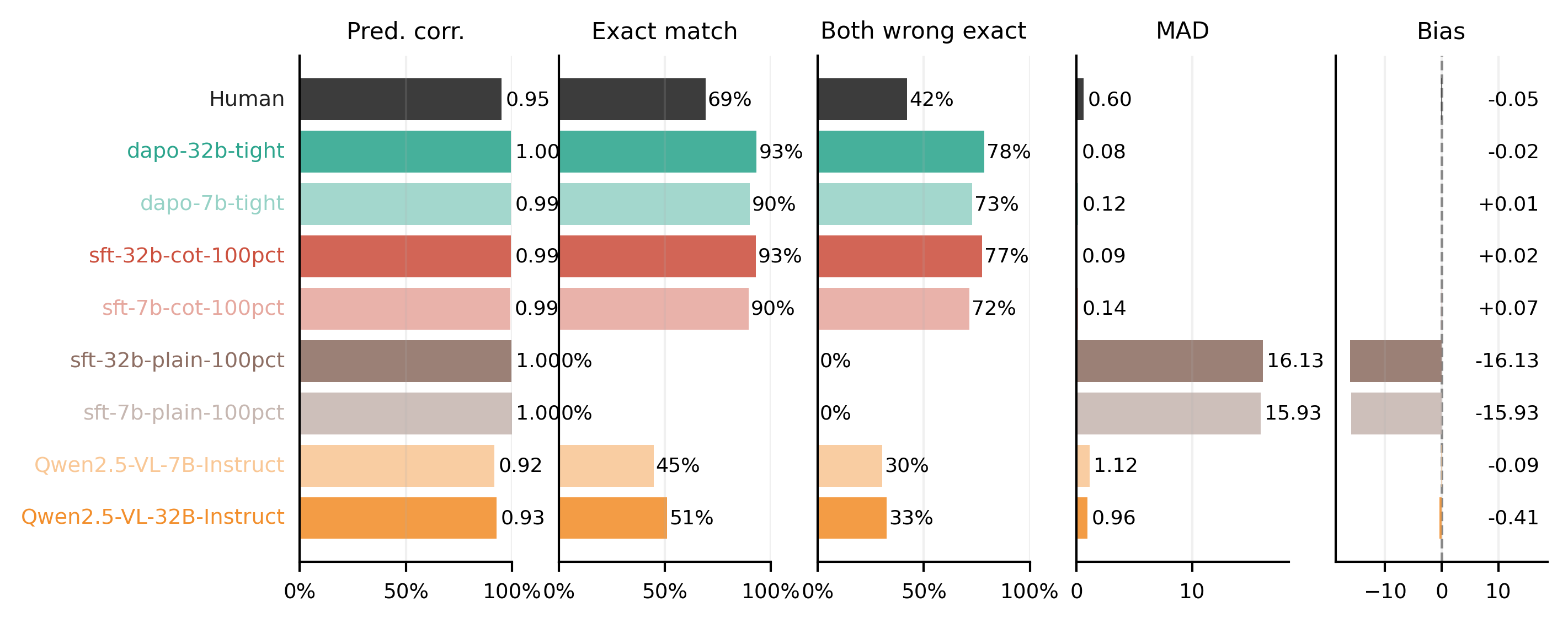}
\caption{Composition diagnostics for the trained-text mechanism analysis (companion to Figure~\ref{fig:tr_e1a_mechanism}). DAPO-tight and SFT-CoT checkpoints reach prediction correlations $\sim 1.00$, exact-match rates of $\approx 90\%$ or higher ($89.6$--$93.3\%$), and both-wrong-exact rates $\geq 71\%$ ($71.6$--$78.5\%$), all read from one response and so not directly comparable with the human baseline. SFT-plain checkpoints have zero exact-match and both-wrong-exact rates (their prediction correlation is near $1$ only because their component answers usually repeat the total) and MAD values of about $16$. Their sub-task answers come from separate queries.}
\label{fig:tr_e1aa_composition}
\end{figure}

\paragraph{Hierarchy and mechanism under training interventions.}
Figures~\ref{fig:tr_e1a_mechanism} and~\ref{fig:tr_e1aa_composition} report the Summation Mechanism analysis for the trained checkpoints in the same format used for the zero-shot text models (Figure~\ref{fig:text_mechanism} of the main paper and Figure~\ref{fig:text_e1aa_composition} above). The decomposition-aware checkpoints closely follow the Summation Mechanism. DAPO-tight 32B reaches $99\%$ \textit{Object Counting} accuracy in the both-correct cell ($n = 1704$), with the both-wrong cell at $40\%$ and the off-diagonal cells small. SFT-CoT 32B and SFT-CoT 7B show the same pattern at slightly lower magnitudes ($98\%$ both-correct in each case). The explicit composition diagnostics in Figure~\ref{fig:tr_e1aa_composition} show the same pattern: prediction correlations reach $\sim 1.00$ for all DAPO-tight and SFT-CoT checkpoints, exact-match rates reach $89.6$--$93.3\%$, and even the both-wrong-exact rate exceeds $70\%$. Because component predictions are extracted from a single chain-of-thought response, these diagnostics measure within-response consistency rather than independently elicited sub-task competence. The exact-match statistic uses the error-based definition in Appendix~\ref{app:hierarchy_dependency}. The SFT-plain checkpoints produce a different structure. Their $V \checkmark H \checkmark$ cell is empty (n/a) because their \textit{Hidden Object Count} answers are never correct, so the joint $\textit{Visible Object Count} \times \textit{Hidden Object Count}$ table degenerates into a one-column slice: V$\checkmark$H$\times$ at $89\%$ ($32$B) / $85\%$ ($7$B) and V$\times$H$\times$ at $18\%$ / $21\%$. The exact-match rates in Figure~\ref{fig:tr_e1aa_composition} for SFT-plain are zero, with mean absolute deviations of about $16$. SFT-plain sub-task answers repeat the total on $88$--$93\%$ of queries, so the reported hidden count is never correct. The DAPO-tight checkpoints improve all four bundles over the zero-shot backbone. The contrast with SFT-plain gives a training-based dissociation between final-task gain and structural recovery that complements the observational dissociation reported in Section~\ref{sec:analysis}. The decomposition serves as a training signal that jointly improves target and sub-task competence, whereas the elicited sub-task answers of final-answer imitation do not show the hidden branch or the visible-plus-hidden assembly. This response-level dissociation partly reflects how sub-tasks are elicited and does not establish the decomposition as a causal prerequisite of counting.

\subsection{Direct Dependence Tests}
\label{app:dependence}

\paragraph{Inference-time intervention.}
We give three zero-shot models the correct intermediate values and ask for the total. Table~\ref{tab:intervention} shows \textit{Object Counting} accuracy rising from $14$--$18\%$ to $77$--$81\%$. Because \textit{Object Counting} is defined as visible plus hidden objects, recovering the total from supplied values is expected, so the test shows sufficiency at most. It does not show that a model computes the total through these values when they are withheld, and it does not validate the two Count Hierarchies, which remain diagnostic hypotheses.

\begin{table}[t]
\centering
\small
\caption{Inference-time intervention: \textit{Object Counting} accuracy in percent, with and without the correct intermediate values.}
\label{tab:intervention}
\begin{tabular}{lcc}
\toprule
Model & Zero-shot & Correct decomposition supplied \\
\midrule
Qwen & 17.7 & 80.6 \\
Gemini & 14.8 & 79.8 \\
Claude & 14.1 & 76.6 \\
\bottomrule
\end{tabular}
\end{table}

\paragraph{Answer-masked training.}
As the converse test, we train Qwen2.5-VL-7B on SFT-CoT targets in which the final total is replaced by a literal placeholder, while every sub-task value stays supervised. The model reproduces the placeholder, so its emitted accuracy is $0\%$, and we score the implied total instead, the sum of the two operands in its own final line. On the validation split, implied acc@1 is $25.6\%$ (acc@10\_maj $30.19\%$, pass@10 $81.73\%$), about $88\%$ of the $29.2\%$ reached by SFT-CoT with supervised answers, against $5.3\%$ for the zero-shot backbone. On $48.5\%$ of training views, however, there is no hidden block, so the total equals the visible count that remains in the target. Every intermediate value is also still supervised, so the test shows that the total can be reconstructed from the sub-task chain and does not validate the Count Hierarchies. DAPO-tight on this checkpoint received no usable reward signal, because every rollout earned the same reward once the tags were correct, and implied acc@1 fell to $23.7\%$.

\subsection{Preliminary External Probes}
\label{app:external}

\paragraph{Real images.}
As an initial sim-to-real probe, we hand-built $50$ stacked cardboard-box structures, recorded every object including occluded ones for ground truth, and photographed them under real lighting. The Qwen2.5-VL-7B checkpoints, trained only on synthetic renders, reach $4.0\%$ zero-shot, $24.0\%$ after SFT-plain, $28.0\%$ after SFT-CoT and $38.0\%$ after DAPO-tight. The ordering matches the synthetic results, but the set is small and covers one object type, so we treat this result as preliminary.

\paragraph{LIBERO comparison.}
We compare \textit{Object Counting} accuracy with the average LIBERO~\citep{liu2023libero} success rate across five VLAs that retain text capability (Table~\ref{tab:libero}). The rank correlation is Spearman $\rho = 0.80$ with $n = 5$, which is not statistically significant (exact two-sided $p = 0.13$). The LIBERO values are each model's own reported average, and the \projectname{} scores sit in a compressed near-floor range ($1.5$--$4.2\%$), so we read the comparison as a suggestive trend only.

\begin{table}[t]
\centering
\small
\caption{\projectname{} \textit{Object Counting} accuracy and reported LIBERO average success rate for five VLAs, in percent.}
\label{tab:libero}
\begin{tabular}{lcc}
\toprule
Model & \projectname{} & LIBERO avg.\ success rate \\
\midrule
OpenVLA~\citep{kim2024openvla} & 1.5 & 76.5 \\
CoT-VLA~\citep{zhao2025cotvla} & 1.9 & 81.1 \\
SpatialVLA~\citep{qu2025spatialvla} & 2.0 & 78.1 \\
UniVLA~\citep{wang2026univla} & 3.4 & 95.5 \\
VLA-0~\citep{goyal2025vla} & 4.2 & 94.7 \\
\bottomrule
\end{tabular}
\end{table}

\section{Discussion and Future Work}
\label{app:discussion}

\subsection{Conclusion and Contributions}

\projectname{} reframes spatial-intelligence evaluation as a hierarchical decomposition problem rather than as a black-box leaderboard. The benchmark contributes three things in combination. First, a methodology: a single target task (\textit{Object Counting}) is decomposed into nine perceptual and cognitive sub-tasks organized by pre-specified dependency, with relations (the Summation Mechanism, the Internal Referential Chain, and the Visible and Hidden Count Hierarchies) that are tested empirically rather than treated as taxonomic groupings. Second, a dataset: approximately 80{,}000 procedurally generated scenes with per-task ground truth, three response formats, and a human baseline, all anchored to the developmental psychology literature on block counting~\citep{kaufman1983kaufman, drozdick2018kaufman}. Third, a training paradigm: chain-of-thought supervision over the sub-task hierarchy followed by reinforcement learning with verifiable rewards, where the decomposition serves as a training signal that jointly improves target and sub-task competence, not as a proven causal prerequisite. The diagnostic and training contributions are connected by the same hierarchy. The dissociations the diagnostic exposes (e.g., a model that produces correct totals while failing on the hidden-object branch, or that follows the Internal Referential Chain cleanly while collapsing on \textit{Visible Object Count}) are the failure modes that the chain-of-thought training target is constructed to mitigate.

\subsection{Limitations}
\label{app:limitations}

\paragraph{Cross-modality comparisons are approximate.}
The three response formats (text free-response, multiple-choice with image options, and image editing) probe the same underlying tasks but differ in how the answer is produced and selected, so absolute scores across formats cannot be compared point-for-point. The answer-selection burden in MCQ introduces a variance source absent from text and image editing, which compresses spread on the target task at five-choice and partly motivates our reporting of three-choice and four-choice variants. The remaining format differences and their consequences for cross-modality interpretation are documented in Appendix~\ref{app:cross_modality}.

\paragraph{Single-target diagnostic scope.}
The hierarchy presented in this paper is specific to \textit{Object Counting}. \textit{Mental Rotation}, our second target task, sits outside this decomposition by design (Section~\ref{sec:mental_rotation_complement}); it functions as an independent probe of viewpoint transformation rather than as a parallel hierarchy (Appendix~\ref{app:mental_rotation}). Other classical spatial tasks (paper folding, water-level prediction, scene reorganization under occlusion) would each require their own pre-target dependency graphs, which we do not develop here.

\paragraph{Training experiments on two model families.}
Our two-phase training paradigm is evaluated on Qwen2.5-VL-7B and -32B in the text modality only, with a Gemma-4-31B replication on the validation split (Appendix~\ref{app:training_results}). The 32B DAPO-tight validation-split evaluation timed out at about $95\%$ coverage, although Table~\ref{tab:training_summary} covers all benchmark views. We did not run the paradigm in MCQ or image-editing modalities. Decomposition-aware checkpoints recover hierarchy structure that SFT-plain does not, but we show this on Qwen2.5-VL only, and Gemma-4-31B supports the ordering on \textit{Object Counting} alone. We therefore do not claim cross-architecture generalization.

\paragraph{Synthetic scenes and limited diversity.}
All scenes are synthetic structures on a $4 \times 4 \times 4$ voxel grid, with limited object category, appearance and geometry diversity. The robustness analyses in Appendices~\ref{app:difficulty_modalities} and~\ref{app:object_results} probe some of these axes, and the external evidence in Appendix~\ref{app:external} is preliminary. We do not yet address mixed-object and irregular structures or varied object sizes, which we leave as future extensions.

\paragraph{Perception and reasoning are partly entangled.}
Most sub-tasks draw on perception and on structural reasoning, so a wrong answer does not by itself identify which of the two failed. The camera manipulations and the distinct-color control in Appendix~\ref{app:difficulty_modalities} separate them only in part.

\subsection{Future Work}
\label{app:future_work}

\paragraph{Extending the framework to additional spatial-intelligence tasks.}
The \projectname{} methodology is not specific to block counting. Any classical spatial task that admits a pre-target dependency graph (paper folding, water-level prediction, scene reorganization under occlusion, mental scanning, perspective taking, and the various subtests of standardized cognitive batteries) could be instrumented in the same way: identify the perceptual and cognitive prerequisites, specify the chains and joint supports that bind them, and test whether response patterns respect the proposed structure. The result would be a multi-task diagnostic suite in which target-task success and intermediate competence can be dissociated for each task family separately, and in which training signals analogous to our chain-of-thought paradigm can be derived per task. We see this as the natural scaling direction for the framework, and one for which the procedural-generation infrastructure we built (Isaac Sim with ground-truth voxel access, semantic-segmentation annotators, and per-task ground-truth derivation) is a reusable foundation.

\paragraph{Action-grounded training and vision-language-action models.}
On the evidence of our small VLA-0-versus-backbone comparison (Appendix~\ref{app:text_full}), action-grounded training on tasks unrelated to counting may transfer positively to the spatial reasoning the benchmark probes. We are currently pursuing a stronger test of this hypothesis: a joint-training experiment that integrates the \projectname{} chain-of-thought training signal directly into vision-language-action training, with the dual goal of improving block counting accuracy on \projectname{} while preserving (and ideally improving) standard VLA performance on its native action-prediction objectives. A successful outcome would be evidence in two directions at once: that action grounding benefits spatial intelligence, and that explicit hierarchical training over a spatial decomposition does not interfere with (and may complement) the action-decoding capabilities a VLA must retain. Beyond this immediate experiment, the broader question is whether the directional finding from VLA-0 generalizes across (backbone, VLA) pairs and across spatial benchmarks beyond \projectname{}; if it does, it would suggest that physical grounding through action data is a useful complement to passive visual or linguistic supervision for training spatial reasoning into MLLMs more broadly.

\paragraph{Broader tasks and external validation.}
Broadening the evidence calls for more target tasks and support structures, along with real-image sets larger than the $50$ structures of Appendix~\ref{app:external}. Mixed-object and irregular structures and varied object sizes are natural extensions of the generator.

\paragraph{Extension to video and world-model evaluation.}
A separate line of work asks whether video generation and world-models simulate physical reality or produce plausible-looking pixels without grounded physical understanding~\citep{li2025worldmodelbench, meng2024towards, duan2025worldscore, qin2024worldsimbench}. The \projectname{} hierarchy is in principle extensible to this setting: the same pre-target dependency structure could be evaluated on video sequences in which the structure is built up or knocked down across time, with each sub-task evaluated on intermediate frames. We see this as a natural way to bridge the static-MLLM and video-generation evaluation literatures, both of which currently report end-task scores without diagnosing where the underlying spatial reasoning breaks down.


\newpage
\section*{NeurIPS Paper Checklist}

\begin{enumerate}

\item {\bf Claims}
    \item[] Question: Do the main claims made in the abstract and introduction accurately reflect the paper's contributions and scope?
    \item[] Answer: \answerYes{}
    \item[] Justification: The abstract and introduction describe three contributions: (1) a hierarchical diagnostic framework that decomposes 3D object counting into nine perceptual and cognitive sub-tasks, (2) a procedurally generated dataset of approximately 80,000 scenes of stacked 3D structures with per-task ground truth and a human baseline, and (3) a two-condition training paradigm (SFT-CoT and DAPO-tight) that improves both target-task accuracy and intermediate competence. Each claim is supported in the body: the framework in Section~\ref{sec:analysis}, the dataset in Section~\ref{sec:dataset_generation}, and the training results in Section~\ref{sec:training}.

\item {\bf Limitations}
    \item[] Question: Does the paper discuss the limitations of the work performed by the authors?
    \item[] Answer: \answerYes{}
    \item[] Justification: Limitations are discussed in the main-paper Discussion (Section~\ref{sec:discussion}) and expanded in Appendix~\ref{app:limitations}. They include cross-modality comparisons being approximate because response format changes the answer-selection demands, the single-target diagnostic scope (the hierarchy is specific to \textit{Object Counting} and does not extend to \textit{Mental Rotation}), and training experiments conducted on two model families in the text modality only. The appendix also covers the synthetic $4 \times 4 \times 4$ voxel grid with limited object category, appearance and geometry diversity, the partial entanglement of perception and reasoning, and the preliminary status of the external evidence.

\item {\bf Theory assumptions and proofs}
    \item[] Question: For each theoretical result, does the paper provide the full set of assumptions and a complete (and correct) proof?
    \item[] Answer: \answerNA{}
    \item[] Justification: The paper does not include theoretical results.

    \item {\bf Experimental result reproducibility}
    \item[] Question: Does the paper fully disclose all the information needed to reproduce the main experimental results of the paper to the extent that it affects the main claims and/or conclusions of the paper (regardless of whether the code and data are provided or not)?
    \item[] Answer: \answerYes{}
    \item[] Justification: All training and evaluation details necessary for reproduction are provided. The dataset generation procedure is described in Section~\ref{sec:dataset_generation} and Appendix~\ref{app:dataset_details}; the prompt structure, concept definitions, and per-task queries are reproduced verbatim in Appendix~\ref{app:prompts}; the training procedure (SFT-CoT and DAPO-tight) is described in Section~\ref{sec:training} with the full chain-of-thought trace structure, reward function, hyperparameters, and evaluation recipe in Appendix~\ref{app:training}; the human study protocol is described in Appendix~\ref{app:human_study}.

\item {\bf Open access to data and code}
    \item[] Question: Does the paper provide open access to the data and code, with sufficient instructions to faithfully reproduce the main experimental results, as described in supplemental material?
    \item[] Answer: \answerYes{}
    \item[] Justification: The released \projectname{} benchmark view images and integer ground-truth annotations are available at \url{https://huggingface.co/datasets/patrickqrim/spatial-iq}. The data-generation, evaluation, analysis and training-recipe code, together with raw model outputs, is available at \url{https://github.com/NVIDIA/Spatial-IQ}.

\item {\bf Experimental setting/details}
    \item[] Question: Does the paper specify all the training and test details (e.g., data splits, hyperparameters, how they were chosen, type of optimizer, etc.) necessary to understand the results?
    \item[] Answer: \answerYes{}
    \item[] Justification: Section~\ref{sec:validation} describes the model set, evaluation set size, and the three response modalities. Section~\ref{sec:training} describes the three training conditions and the zero-shot baseline, the train/benchmark split (enforced at the structure level), and the model scales (Qwen2.5-VL-7B and -32B). Appendix~\ref{app:training} reports the chain-of-thought trace structure, the integer-distance reward function, the GRPO/DAPO hyperparameters, the KL anchor on the SFT-CoT initialization, and the validation-split evaluation recipe ($k{=}10$, temperature $1.0$, top-$p$ $1.0$, max $1024$ new tokens).

\item {\bf Experiment statistical significance}
    \item[] Question: Does the paper report error bars suitably and correctly defined or other appropriate information about the statistical significance of the experiments?
    \item[] Answer: \answerYes{}
    \item[] Justification: Statistical significance is reported throughout. Paired exact McNemar tests with Benjamini-Hochberg FDR correction cover between-model comparisons on \textit{Object Counting} (Section~\ref{sec:validation_spread}, Table~\ref{tab:text_task_accuracy_full}). View-level Wilson 95\% confidence intervals cover per-model and per-object accuracies (Tables~\ref{tab:text_task_accuracy_full} and~\ref{tab:text_difficulty_bins}, Figure~\ref{fig:training_object_generalization}). Permutation tests with FDR correction cover within-scene lifts on hierarchy edges, and view-clustered bootstrap confidence intervals cover the two-way fixed-effects estimates (Appendix~\ref{app:hierarchy_dependency}). One-sided binomial tests against the chance baseline cover wrong-answer preferences in MCQ (Figure~\ref{fig:mcq_wrong_answers}), and BH-FDR-corrected significance markers cover per-responder camera-effect deltas (Figure~\ref{fig:c13_camera_within_structure}).

\item {\bf Experiments compute resources}
    \item[] Question: For each experiment, does the paper provide sufficient information on the computer resources (type of compute workers, memory, time of execution) needed to reproduce the experiments?
    \item[] Answer: \answerYes{}
    \item[] Justification: All training runs were conducted on an NVIDIA GB200 cluster (4 GPUs per node, 184 GB per GPU, aarch64) using the NeMo-RL v0.5.0 container (a NeMo-RL main checkout for Gemma-4-31B), with 8 nodes (32 GPUs) per job for the Qwen2.5-VL runs and up to 32 nodes per job for the Gemma-4-31B replication. SFT-CoT and SFT-plain training of Qwen2.5-VL-7B-Instruct and Qwen2.5-VL-32B-Instruct each completed in well under a day per run. DAPO-tight reinforcement learning ran for 500 steps and took approximately 2.6 hours at 7B and 4.2 hours at 32B. Open-weights model evaluation was served locally via vLLM (tensor-parallel size 1 for 7B, 4 for 32B); closed frontier models were accessed via their respective APIs and incurred no local GPU cost. Including preliminary and failed runs not reported in the paper (earlier 3B attempts, KL/LR sweeps, and a $1 \times 64$ prompts/generations configuration that aborted at step 0 due to dynamic-sampling collapse), the full project consumed approximately 5{,}000 GPU-hours.
    
\item {\bf Code of ethics}
    \item[] Question: Does the research conducted in the paper conform, in every respect, with the NeurIPS Code of Ethics \url{https://neurips.cc/public/EthicsGuidelines}?
    \item[] Answer: \answerYes{}
    \item[] Justification: The research conforms to the NeurIPS Code of Ethics. Human studies were conducted with NVIDIA's in-house annotation team under standard internal compensation channels (Appendix~\ref{app:human_study}), with no external recruitment platform and no collection of personally identifying information. The dataset is procedurally generated and contains no human subjects, identifiable images, or scraped content.

\item {\bf Broader impacts}
    \item[] Question: Does the paper discuss both potential positive societal impacts and negative societal impacts of the work performed?
    \item[] Answer: \answerNA{}
    \item[] Justification: \projectname{} is a diagnostic benchmark and training paradigm aimed at improving the spatial-reasoning competence of multimodal models, which has positive implications for embodied AI applications such as warehouse robotics, assistive devices, and physical task automation that depend on accurate counting and support reasoning. The introduction notes these positive implications. However, we see no direct path to negative societal impact: the benchmark uses synthetic scenes of stacked geometric objects and the training paradigm targets a narrow spatial-reasoning skill that has no clear direct malicious application. This is why we answer NA.
    
\item {\bf Safeguards}
    \item[] Question: Does the paper describe safeguards that have been put in place for responsible release of data or models that have a high risk for misuse (e.g., pretrained language models, image generators, or scraped datasets)?
    \item[] Answer: \answerNA{}
    \item[] Justification: Not applicable, as the released artifacts (synthetic scenes of stacked objects, ground-truth annotations, and chain-of-thought training traces) pose no foreseeable misuse risk.

\item {\bf Licenses for existing assets}
    \item[] Question: Are the creators or original owners of assets (e.g., code, data, models), used in the paper, properly credited and are the license and terms of use explicitly mentioned and properly respected?
    \item[] Answer: \answerYes{}
    \item[] Justification: We have properly attributed the work of others in our paper and have followed licensing and usage terms.

\item {\bf New assets}
    \item[] Question: Are new assets introduced in the paper well documented and is the documentation provided alongside the assets?
    \item[] Answer: \answerYes{}
    \item[] Justification: The new assets introduced are the \projectname{} benchmark views with per-task integer ground truth (constructed from a pool of approximately 80,000 procedurally generated scenes of stacked 3D structures), the human-baseline responses, and the raw model outputs. Dataset construction is documented in Section~\ref{sec:dataset_generation} and Appendix~\ref{app:dataset_details} (sampling distribution, camera and environment variation, depth-buffer occlusion test, ground-truth derivation, and partitioning). Prompts and concept definitions are reproduced verbatim in Appendix~\ref{app:prompts}, and the training procedure for the evaluated checkpoints is documented in Appendix~\ref{app:training}. Documentation accompanying the release is available at \url{https://github.com/NVIDIA/Spatial-IQ}.

\item {\bf Crowdsourcing and research with human subjects}
    \item[] Question: For crowdsourcing experiments and research with human subjects, does the paper include the full text of instructions given to participants and screenshots, if applicable, as well as details about compensation (if any)? 
    \item[] Answer: \answerYes{}
    \item[] Justification: Human responses were collected from NVIDIA's in-house annotation team as described in Appendix~\ref{app:human_study}. The protocol includes annotator chunk size (100 questions per task), warm-up calibration trials, the verbatim instructions and concept definitions shown to annotators, and screenshots of the three annotator interfaces (Figures~\ref{fig:human_text_interface}, \ref{fig:human_mcq_interface}, and~\ref{fig:human_image_interface}). Annotators are NVIDIA employees compensated through standard internal channels.

\item {\bf Institutional review board (IRB) approvals or equivalent for research with human subjects}
    \item[] Question: Does the paper describe potential risks incurred by study participants, whether such risks were disclosed to the subjects, and whether Institutional Review Board (IRB) approvals (or an equivalent approval/review based on the requirements of your country or institution) were obtained?
    \item[] Answer: \answerNA{}
    \item[] Justification: Not applicable.

\item {\bf Declaration of LLM usage}
    \item[] Question: Does the paper describe the usage of LLMs if it is an important, original, or non-standard component of the core methods in this research? Note that if the LLM is used only for writing, editing, or formatting purposes and does not impact the core methodology, scientific rigorousness, or originality of the research, declaration is not required.
    \item[] Answer: \answerNA{}
    \item[] Justification: LLMs are not a non-standard core methodological component of this research. The trained Qwen2.5-VL checkpoints described in Section~\ref{sec:training} are an evaluated artifact of the paper but apply standard supervised-fine-tuning and RLVR techniques to an open-weights backbone, not a novel LLM methodology.

\end{enumerate}

\end{document}